 \documentclass[final,5p,times,twocolumn]{elsarticle}

\usepackage{amssymb} 

\usepackage{graphicx}
\usepackage{adjustbox}
\usepackage{subfigure}
\usepackage{amsmath}
\usepackage{lineno,hyperref}
\usepackage{xcolor}
\usepackage{caption}

\journal{Expert Systems with Applications}

\begin{document}

\begin{frontmatter}

\title{A review of   ensemble  learning   and data augmentation  models  for   class imbalanced problems: combination, implementation and evaluation}


\author[Second]{Azal Ahmad Khan \corref{cor1}}
\ead{k.azal@iitg.ac.in}

\author[Second]{Omkar Chaudhari}
\ead{c.omkar@iitg.ac.in} 

\author[First,third]{Rohitash Chandra\corref{cor1}}
\ead{rohitash.chandra@unsw.edu.au}  


\address[First]{Transitional Artificial Intelligence Research Group, School of Mathematics and Statistics, University of New South Wales,  Sydney,  Australia}

\address[third]{Pingala Institute of  Artificial Intelligence,   Sydney,  Australia}
\address[Second]{ Department of Chemistry,  Indian Institute of Technology Guwahati, Assam, India}

\begin{abstract}
 
 Class imbalance (CI) in classification problems arises when the number of observations belonging to one class is lower than the other. Ensemble learning combines multiple models to obtain a robust model and has been prominently used with data augmentation methods to address class imbalance problems. In the last decade, a number of strategies have been added to enhance ensemble learning and data augmentation methods, along with new methods such as \textit{generative adversarial networks} (GANs). A combination of these has been applied in many studies, and the evaluation of different combinations would enable a better understanding and guidance for different application domains. 
 In this paper, we present a computational study to evaluate data augmentation and ensemble learning methods used to address prominent benchmark CI problems.  We present a general framework that evaluates 9 data augmentation and 9 ensemble learning methods for CI problems. Our objective is to identify the most effective combination for improving classification performance on imbalanced datasets. The results indicate that combinations of data augmentation methods with ensemble learning can significantly improve classification performance on imbalanced datasets. We
find that traditional data augmentation methods such as the \textit{synthetic minority oversampling technique} (SMOTE) and \textit{random oversampling} (ROS) are
not only better in performance for selected CI problems, but also computationally less
expensive than GANs. Our study is vital for the development of novel models for handling imbalanced datasets.

\end{abstract}


\end{frontmatter}

\section{Introduction} 
 Class imbalance (CI) is a challenging problem for machine learning that arises with a disproportionate ratio of instances. CI problem is a typical problem in classification tasks \cite{bi2018empirical} in a wide range of applications. In the case of binary classification, the larger number of instances makes the majority class, while a much lower number of instances makes the minority class.  Imbalanced classification datasets are predominant in scenarios where anomaly detection is crucial, such as fraud bank transactions \cite{wei2013effective}, rare disease \cite{bria2020addressing} \cite{qin2020gan}, natural disasters \cite{haixiang2017learning}, and software defect detection \cite{rodriguez2014preliminary}.
 
 Machine learning models used for classification are typically designed with the assumption of an equal number of instances for each class.  In CI datasets, machine learning models tend to be more biased towards the majority class, causing improper classification of the minority class and leading to poor classification performance. This causes high true positive rates(TPR) but a low true negative rate (TNR) when most instances are positive \cite{wei2013role}. We note that typical metrics do not apply to CI problems, and specialised metrics have been developed, such as the precision and recall curve \cite{buckland1994relationship}, and the F1 score \cite{goutte2005probabilistic}. In the case of CI datasets, although overall classification accuracy would be very high -- up to 98\%, conventional machine learning models would misclassify a significant portion of the data samples. In the case of medical diagnosis, the model would be unreliable since it cannot identify patients effectively and incorrect classification can lead to further medical problems and even death \cite{johnson2019survey}. Therefore, the conventional classification accuracy is not an appropriate evaluation metric and sensitivity and specificity are often used for evaluation \cite{liu2022solving}. Japkowicz et al. \cite{japkowicz2000class}, and Japkowicz et al.  \cite{japkowicz2002class} are amongst the first studies that highlighted the CI problem for the machine learning community (2000 and 2002). Several techniques have been proposed at the data and model level to deal with CI datasets, such as undersampling majority class \cite{liu2008exploratory,lin2017clustering,yen2006under,bunkhumpornpat2011mute}, oversampling minority class \cite{perez2015oversampling,barua2011novel}, cost-sensitive learning \cite{ling2008cost}, one-class classification \cite{tax2002one,hempstalk2008one,ruff2018deep}, and data augmentation  \cite{zhu2017synthetic} that focuses on creating synthetic data to support minority classes \cite{makki2019experimental,haixiang2017learning,johnson2019deep}.

Ensemble learning methods combine multiple models to obtain a comprehensive and robust model \cite{xiao2019svm}.  Ensemble methods have produced better results than other algorithms for datasets with CI problems. Examples of ensemble methods include bagging, boosting, and stacking with prominent implementations such as AdaBoost \cite{solomatine2004adaboost} and random forests \cite{breiman2001random}.  In the case of Twitter spam detection, ensemble methods have shown better accuracy results and other CI metrics such as the F1 score \cite{liu2017addressing}. Ensemble methods have been prominent on CI datasets such as predicting cyanobacteria blooms \cite{shin2021effects}, drug-target interaction prediction \cite{ezzat2016drug}, diagnosing bearing defects \cite{farajzadeh2016efficient}, and empowering one-vs-one decomposition \cite{zhang2016empowering}.  A survey on applications of ensemble learning methods for CI problems  \cite{mienye2022survey}  suggested that random forest and XGBoost have been mostly used in the literature and both methods offer reliable performances. In fraud detection, AdaBoost has been reported to be the most popular, whereas stacking has been mainly preferred for sentiment analysis \cite{hajek2023speech}.

  Data augmentation \cite{shorten2019survey,iwana2021empirical,bayer2022survey} is a data-level approach for enhancing machine learning models which has also been used for handling the CI problem by reducing class imbalance. Data augmentation is a field of active research, and several methods have been introduced lately that focus on multi-class imbalanced problems \cite{prachuabsupakij2012clustering}.
 Examples of data augmentation for CI problems include fault classification \cite{9115249,abdelgayed2017fault,yang2019real,jan2021distributed}, and protein classification \cite{zhao2008protein}. 

 In this paper, we review and evaluate data augmentation and ensemble techniques to address prominent benchmark CI problems. We propose a general framework that evaluates 23 binary class datasets with different imbalance ratios on 9 data augmentation and 9 ensemble learning methods. Following this, we share insights into the results by comparing these techniques over different metrics. \textcolor{black}{We also demonstrate the suitability of our framework for evaluating multi-class imbalance problems.}
 
 We note that a related paper that provided a computational review of related methods was published in 2011 \cite{galar2011review} which did not cover GAN-based data augmentation. \textcolor{black}{Moreover, a review by Haixiang et al. \cite{haixiang2017learning}  of papers published over the past decade highlighted the broad applicability of rare event detection and imbalanced learning across diverse research domains. They conducted a comprehensive analysis of these papers, examining modelling methods, including data preprocessing, classification algorithms, and model evaluation. Notably, we identify a crucial gap, which is the absence of performance analysis due to the lack of computational evaluation, which we aim to address in our work}. Furthermore, in our paper, we not only provide a comprehensive review but also a computational review by evaluating the key methods and presenting open-source code and data to the community. 
 
 The rest of the paper is organised as follows. In Section 2, we present a literature review and related work on class imbalance problems and applications of ensemble learning and data augmentation in class imbalance problems.  Section 3 presents a review of ensemble learning and data augmentation for CI problems. Section 4  presents the methodology and Section 5 presents the computational review with an experimental analysis. Finally, in Sections 6 and 7, we provide a discussion and conclude the paper. 

\subsection{Methodology}

We review the latest papers that use data augmentation and ensemble methods to counter class imbalance problems.   The main keywords used for searching documents on Google Scholar \footnote{https://scholar.google.com.au/} and Scopus \footnote{https://scopus.com} include “class imbalance,” “data augmentation,” and “ensemble learning.”

\begin{figure}
\centering
\subfigure[Keyword search: ``Class Imbalance."]{
    \includegraphics[height=5cm]{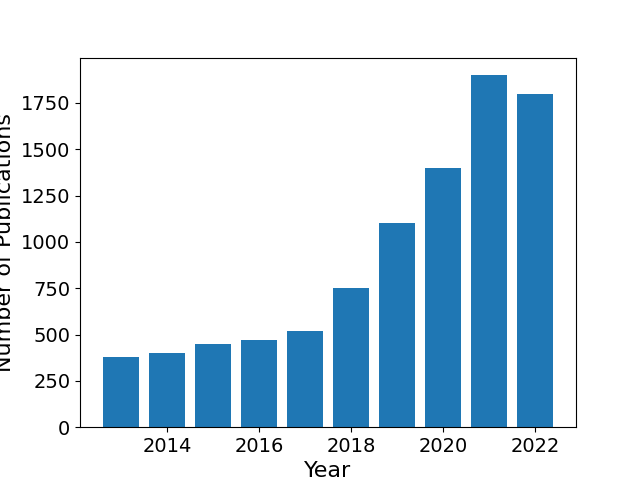}%
}
\subfigure[Keyword search:  ``Ensemble Learning."]{
    \includegraphics[height=5cm]{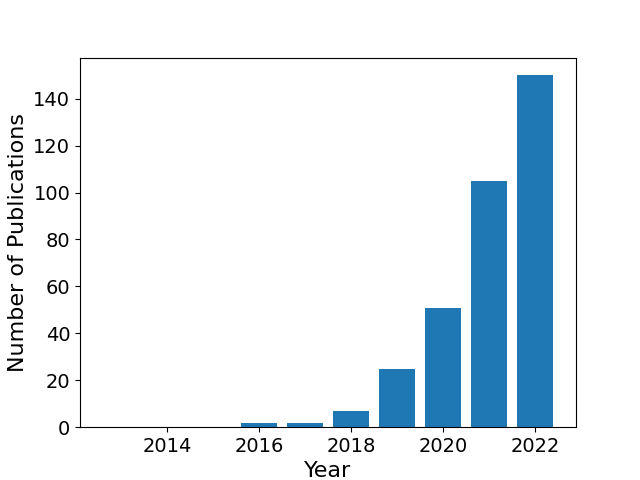}%
}
\subfigure[Keyword search:  ``Class Imbalance" and ``Ensemble Learning."]{
    \includegraphics[height=5cm]{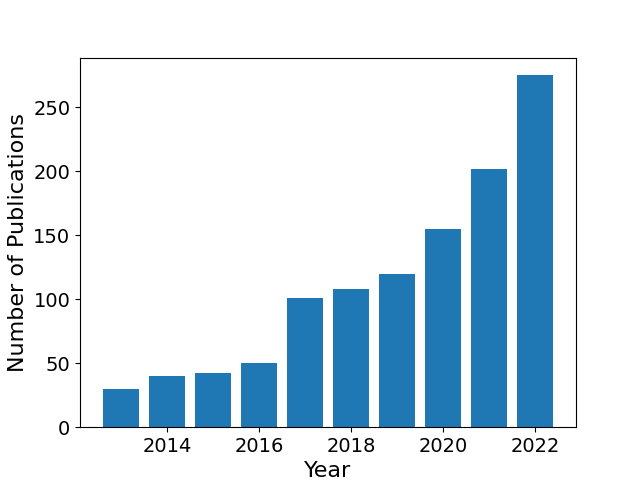}%
}
\subfigure[Keyword search:  ``Class Imbalance" and ``Data Augmentation" and ``Ensemble Learning."]{
    \includegraphics[height=5cm]{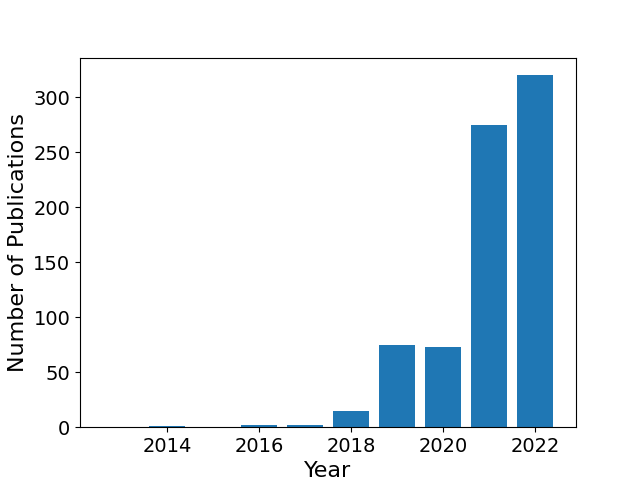}%
}
\caption{Number of publications with different combinations of keyword search using the Scopus database.}
\label{fig:keywords}
\end{figure}

We plot the frequency of publications on this topic using different combinations of keyword searches in the Scopus database as shown in Figure \ref{fig:keywords}. The figure shows the number of publications in recent years using the keywords “class imbalance” (Panel a), “class imbalance” and “data augmentation” (Panel b), and    “class imbalance,” “data augmentation,” and “ensemble learning” (Panel c). As shown in these plots, the number of publications that focus on the applications of ensemble learning and data augmentation techniques has continuously increased in the last decade.

\section{Ensemble Learning}

Ensemble learning is a meta-learning machine learning method that seeks better predictive performance by combining the predictions from multiple models. Ensemble learning strategically combines classifiers or expert models to address problems such as regression and classification \cite{polikar2012ensemble,dietterich2002ensemble}. There are three main classes of ensemble learning \textit{bagging}, \textit{boosting}, and \textit{stacking}  \cite{zenko2001comparison, dong2020survey}. 

In essence, ensemble learning methods construct multiple, diverse predictive models from adapted versions of the training data (most often reweighed or re-sampled), and they combine the predictions of these models in some way, often by simple averaging or voting (possibly weighted). These classes address multiple problems such as intrusion detection \cite{syarif2012application}, landslide assessment \cite{dou2020improved}, and short-term prediction \cite{ribeiro2020ensemble}. We discuss these approaches and some popular ensemble algorithms based on these approaches.

\subsection{Bagging}

 Bagging combines \textit{bootstrapping} and \textit{aggregation} to form an ensemble model \cite{zhou2012ensemble}. Bootstrapping refers to a sampling technique to create diverse samples. Aggregation refers to the average or the majority of the predictions that are taken to compute a more accurate estimate. Bagging involves fitting many decision trees on different samples of the same dataset and combining their predictions with either averaging or voting. Bagging considers homogeneous weak learners, trains them independently of each other (in parallel) and combines their predictions. There exists a \textit{bias-variance} trade-off when fitting any model, a high bias may result in the model over-generalising the data and missing the relevant features between samples while training. High variance while training may end up with the model trying to fit random noise in the data. Bagging creates bags that feature random subsets of samples, training classifiers on them and combining them to form an ensemble, which helps reduce the variance.

In the last two decades, a number of strategies have been proposed to enhance conventional bagging with a focus on CI problems. Hido et al. \cite{Hido2009Bagging} presented roughly balanced bagging, a new under-sampling technique using a negative binomial distribution for bagging on imbalanced data.  Lango et al. \cite{lango2018multi} extended bagging to handle a higher number of attributes and multiple minority classes. Blaszczynski et al. \cite{Blaszczynski2013}   introduced \textit{local-and-over-all balanced bagging} for CI problems where the probability of sampling an instance from the training data was dependent on the class distribution over the neighbourhood. Similarly,  Blaszczynski et al. \cite{BLASZCZYNSKI2015529} presented \textit{neighbourhood balanced bagging}, where the probability of sampling a data instance was modified according to class distributions in their neighbourhood. Khoshgoftaar et al. \cite{Khoshgoftaar2011} compared boosting techniques (SMOTEBoost and RUSBoost) against bagging techniques (Exactly Balanced Bagging and Roughly Balanced Bagging) and found that the bagging techniques performed better for imbalanced data with noise.

\subsubsection{Random Forests}
 
Decision trees \cite{quinlan1986induction,utgoff1989incremental} implement a tree-like model of decisions and their possible consequences based on measures such as entropy and information gain, and also enable regularisation methods such as pruning \cite{suthaharan2016decision}. Decision trees, when compared to neural networks, are considered white-box models since they are intuitive, i.e. the decision-making process is transparent and easy to understand, and they are among the most widely used learning algorithms. Hence, decision trees have made a huge impact in the area of medicine, and biotechnology \cite{podgorelec2002decision,kingsford2008decision} since transparency of the decision-making process is important. \textit{Iterative dichotomiser} (ID3)\cite{vasudevan2014iterative} and \textit{classification and regression trees} (CART)\cite{loh2011classification, timofeev2004classification} are the popular implementation of decision trees. They can be used for learning classification models (with discrete value predictions) and regression models (with continuous value predictions) \cite{loh2014fifty}. The major limitation of decision trees is their bias in over-training and inability to generalise due to the bias-variance problem \cite{kotsiantis2013decision}.

The random forest is a supervised ensemble learning model widely used in classification and regression problems \cite{ho1995random}. The random forest features an ensemble of decision trees implemented using a bagged ensemble learning framework \cite{breiman2001random,citation-0}. It builds decision trees on different samples and takes their majority vote for classification and average prediction in case of regression \cite{ho1995random}. In comparison with conventional bagging, random forests can also split the data feature-wise, i.e., not all features need to be associated with an instance in a bag. Random forests restrict decision trees as their base model in the ensemble, whereas bagging methods can enable any type of base model in the ensemble; however, random forests are the most prominent implementation of bagging. A random forest ensemble features  $n$ random records (bags) from a dataset containing $k$ records, and a separate decision tree model is constructed from the data in each bag, which generates an output after training. The final decision of the random forest ensemble is based on \textit{majority voting} (classification problems) or \textit{averaging} (regression problems), respectively \cite{breiman2001random}.  Hence, random forests tend to provide greater diversity in the decision tree and generally result in higher bias, and lower variance, i.e., better generalisation on test data when compared to decision trees \cite{biau2016random}. Random forests also have the ability to provide transparency in the decision-making process through the extraction of rules \cite{benard2021interpretable} and also provide feature ranking, i.e.,  information about how much the features in the data contributed to the decision-making process. In this way, the users can discard features and reduce the dataset \cite{alam2019random,sanchez2018feature}. Another advantage of random forests is that they can be easily implemented via parallel computing for big data problems and hence faster implementations, such as \textit{extra-trees} exist \cite{geurts2006extremely}.

Random forests have been prominent in a wide range of domains, that not only include tabular data but also have been applied to image-based data for regression, classification and segmentation problems. Hence,  random forests have been widely applied to remote sensing that deals with very large-scale image-based data \cite{belgiu2016random}. Random forests have also been prominent in areas of medicine and medical diagnosis, such as neuroimaging for Alzheimer's disease \cite{sarica2017random}. For instance, Khalilia et al. \cite{khalilia2011predicting} used random forests to predict the disease risk of individuals based on their medical diagnosis history, which outperformed models such as support vector machines, bagging, and boosting. Furthermore, random forests have been widely used for network intrusion detection systems \cite{resende2018survey}.

Random forests have been widely used to address CI problems \cite{more2017review} and some innovations are as follows. Siers et al. \cite{SIERS201562} presented a cost-sensitive decision forest along with a cost-sensitive voting method for software defect prediction.  Wu et al. \cite{WU2014105} presented a new tree induction method where the idea is to stratify features into two groups, one containing positive features for the minority class, and the other containing negative features for the majority class. The selection of features from each group is based on the term weights, ensuring that each feature subspace contains enough information for both majority and minority classes. 
Zhu et al. \cite{Zhu2018ClassWR} presented a method to train a collection of classifiers assigning different weights to each class instead of a single weight. Badar et al. \cite{bader2018biased} presented a method where instead of oversampling on the data level, the number of classifiers that represent the minority class in the ensemble is over-sampled.

\subsection{Boosting}

Boosting adds the models in the ensemble sequentially that correct the predictions made by prior models and outputs a weighted average of the predictions. The term boosting refers to a family of models that combine a series of weak learners (models) to develop a strong learner \cite{zhou2012ensemble}. Each new weak learner tries to correct its predecessor in order to improve the overall accuracy, and popular boosting methods include  AdaBoost (Adaptive Boosting) \cite{freund1996experiments}, Gradient Boost \cite{friedman2001greedy}, and XGBoost \cite{chen2016xgboost}.

 Furthermore, several strategies have been used to improve canonical boosting. Rayhan et al. \cite{rayhan2017cusboost} combined cluster-based undersampling with boosting. Le et al. \cite{le2018cluster} presented CBoost, a clustering-based boosting algorithm for bankruptcy prediction. Zhang et al. \cite{zhang2019wotboost} proposed a weighted oversampling technique, which uses a weighted distribution to adjust its oversampling strategy at each boosting round. Furthermore, other  alterations  of boosting algorithms  include likelihood-based boosting \cite{tutz2006generalized}, model-based boosting \cite{hothorn2010model}, GAMBoost \cite{hofner2014gamboostlss}, CoxBoost \cite{de2016boosting}, and CatBoost \cite{dorogush2018catboost}.

\subsubsection{AdaBoost} 

AdaBoost is a  boosting technique that was initially designed to improve binary classification performance \cite{freund1996experiments}. AdaBoost combines several weak classifiers based on decision trees to form a single strong classifier \cite{freund1996experiments}. The weak decision trees are known as stumps, which are typically only one level in depth.   AdaBoost uses an iterative approach to learn from stumps and combines them to form an ensemble. Equation \ref{eq:adaboost1} presents the Adaboost ensemble model assuming a dataset with $n$ points (instances).

\begin{equation}
x_i \in \mathbb{R}^{n}, y_i \in \{-1, 1\}
\label{eq:adaboost1}
\end{equation}
 where $x$ is a set of data points and $y$ is the target variable (either -1 or 1) to represent a binary classification problem. We begin by calculating weighted samples for each data point and assign a weight to each training example to determine its importance in the training dataset.  The set of training data instances will have more effect on the training set and stump when the assigned weights are high. Similarly, it will have minimal effect on the training dataset and stump in the case when the weights are low.

\begin{equation}
w = 1/N \in [0,1]  
\label{eq:adaboost2}
\end{equation}

The sum of the weighted samples is  1; hence, the value of each individual weight will lie between 0 and 1. We then calculate the actual influence of this model (classifier) on the classification of the data points using Equation  \ref{eq:adaboost3}.

\begin{equation}
\alpha_t = 0.5 \ln(\frac{(1-TotalError)}{TotalError})  
\label{eq:adaboost3}
\end{equation}

where $\alpha$ determines how much influence the stump will have in the final classification, where $t\epsilon T$ and $T$ are the number of iterations. \textcolor{black}{The TotalError is the weighted sum of errors.} We update the weights ($w$) which we had initially taken as $1/N$ for every data instance ($i$) using Equation \ref{eq:adaboost4}.

\begin{equation}
    w_i = w_{i-1} \times e^{\pm\alpha}
    \label{eq:adaboost4}
\end{equation}

The $\alpha$ is positive when the predicted and the actual output agree, i.e., the sample is classified correctly. In this case, we decrease the sample weight since the ensemble performing well. The $\alpha$  is negative when the predicted output does not agree with the actual class, i.e., the sample is misclassified. In this case, we need to increase the sample weight so that the same misclassification does not repeat in the next stump. This is how the stumps are dependent on their predecessors.

In the comparison of AdaBoost with Random Forests, we note that Random Forests can feature fully grown trees,  while AdaBoost features an ensemble of stumps.  It is reasonable to question the validity of a stump since it just divides the dataset into 2 parts.    AdaBoost combines weak learners to form a strong ensemble. In the Random Forest ensemble, each tree is independent, and the order of the trees does not matter. Once the Random Forest ensemble is created, the class that most trees in the forest vote is chosen as a winner (in case of classification). On the other hand, not all stumps in the AdaBoost have equal weightage.


 Theoretically, AdaBoost does not overfit; Wang et al. \cite{wang2019feature} highlighted the resistance of AdaBoost towards overfitting which is one of its major advantages. The fact that AdaBoost can consistently improve the performance of the classifier by combining multiple weak classifiers is a desirable property for many applications, and there are several studies that discuss the consistency of AdaBoost. Jiang et al. \cite{jiang2004process} demonstrated that under general regularity conditions, a consistent prediction is generated in the process of training the AdaBoost. Bartlett et al. \cite{bartlett2006adaboost} proposed a stopping strategy for  AdaBoost to achieve a universal consistency in performance. There are further variations of AdaBoost to address specific limitations. The multi-class AdaBoost employs forward stage-wise additive modelling that minimises a novel exponential loss for multi-class classification \cite{hastie2009multi}. The combination of AdaBoost with support vector machines (SVM)  achieves better generalisation performance than SVM alone for imbalanced classification problems \cite{li2008adaboost}.

In terms of limitations,   AdaBoost is sensitive to noisy data and outliers since it assigns equal weight to all the training examples. The misclassifications caused by noise and outliers can lead to poor decision boundaries, and hence, there are updated versions of AdaBoost to counter noisy data \cite{ratsch1998regularizing, oza2004aveboost2, gao2010edited, domingo2000madaboost}. AdaBoost has been reported to be computationally slower than Extreme Gradient Boost (XGBoost) \cite{natras2022ensemble} for a set of problems due to implementation and parallel computing capabilities.

 AdaBoost is relatively simple to implement and can be applied to various classification problems. AdaBoost has a wide range of applications in several domains such as intrusion detection \citet{hu2008adaboost}, distributed and scalable online learning \cite{fan1999application}, bankruptcy forecasting \cite{alfaro2008bankruptcy},  health sciences \cite{hatwell2020ada}, demand-driven acquisition prediction \cite{walker2019application}, and sports \cite{markoski2015application}.

Adaboost has also been prominent in addressing CI problems. Taherkhani et al. \cite{taherkhani2020adaboost} presented AdaBoost-CNN for multi-class imbalanced datasets using transfer learning. Kumar et al.\cite{kumar2019tlusboost} presented \textit{Tomek link undersampling-based boosting} (TLUSBoost), which combines \textit{Tomek link and redundancy-based undersampling} (TLRUS) \cite{DEVI20173} and AdaBoost for boosting. Yuan et al. \cite{yuan2012adaboost} proposed a sampling and reweighting strategy to tune AdaBoost towards a certain performance measure. Li et al. \cite{li2019adaboosta}   redefined the error calculation strategy for achieving better prediction accuracy.

\subsubsection{Gradient Boosting}
Gradient Boosting is a machine learning technique that can be used for both classification and regression problems. The \textit{Gradient Boosting regressor} uses the mean-squared error loss, while the \textit{Gradient Boosting classifier} uses the log-likelihood loss \cite{friedman2001greedy}, and also known as \textit{gradient boosting machines}  (GBM) \cite{natekin2013gradient}. The objective is to minimise the loss function by adding consecutive models that are trained using the error residuals from the previous models  \cite{natekin2013gradient}. 

\textcolor{black}{In an ensemble, $M$ is the total number of stages, whereas $m$ is the current stage of Gradient Boosting. We denote the final model as $F\_M(x)$  and $F\_m(x)$ is the model obtained after adding base learners. The boosting procedure begins from high bias and gradually reaches low bias $F\_0 = \gamma$. Next, each model is added from $m=1$ to $m=M$ one at a time, after one other. We obtain the model $F_{m-1}(x)$   by adding $m-1$ weighted base learners and compute pseudo-residuals for each $i$th training example. We compute the loss  $L$ given in Equation \ref{eq:refloss}.}

\begin{equation}
L = \frac{1}{n}\Sigma_{i=1}^n(y_i - F(x_i))^2
\label{eq:refloss}
\end{equation}
$y_i$ is the observed value.

\begin{equation}
r_{im} = - {\frac{\partial L(y, F_{m-1} (x)}
                 {\partial F_{m-1}(x)}}|_{x=x_i, y=y_i} \forall{i=1, 2, ... ,n}
                 \label{eq:gradboost1}
\end{equation}
               
In Equation 
                 \ref{eq:gradboost1}, we compute each residual calculation $r_im$ for $i$th training example to the current base learner $m$  on the weighted sum of base learners from $1$ to $m-1$, and the initial constant function. Next, we generate a new dataset from the original dataset and train (fit) the base learner $h_{m}(x)$ as shown in Equation \ref{eq111}.

\begin{equation}
D = \{(X_i, \gamma_{im}): i=1, 2, ... ,n\}
\label{eq111}
\end{equation}

\begin{equation}
\gamma_{im} = argmin \Sigma_{i=1}^n L(y_i, F_{m-1}(x_i)+\gamma h_m(x_i))
\label{eq:gradboost5}
\end{equation}

Therefore, we calculate $F_m(x)$  using Equation \ref{eq:gradboost3}:
\begin{equation}
F_m(x) = F_{m-1}(x)+\gamma h_m(x)
\label{eq:gradboost3}
\end{equation}

We fit the function $h_m(x)$  to the rate of change of loss $L$ with respect to $F_m-1(x)$. The function $h_m(x)$  approximates the behaviour of the derivative of the loss with respect to $F_m-1(x)$. The function $h_m(x)$ represents the direction in which the loss function decreases with respect to $F_m-1(x)$.

In Equation \ref{eq:gradboost6}, we  formulate an optimisation problem to find $\gamma_{optimum}$:

\begin{equation}
\begin{split}
    \gamma_{optimum} & = argmin\sum_{i=1}^{n}L(y_i, F_m(x_i)) \\
    & = argmin\sum_{i=1}^{n}L(y_i,  F_{m-1}(x)+\gamma h_m(x))    
\end{split}
\label{eq:gradboost6}
\end{equation}

\textcolor{black}{We run the procedure for each base model $m = 1 to M$ and after the $M$ iterations, we obtain the final model $F_M(x)$ in Equation \ref{eq:gradboost7}.}
\begin{equation}
F_m(x) = F_{m-1}(x)+\gamma_{optimum} h_m(x)
\label{eq:gradboost7}
\end{equation}

 Gradient Boosting is known for its high accuracy and ability to handle large datasets. According to a study by Chen et al. \cite{chen2016xgboost}, Gradient Boosting was one of the best machine learning models for various benchmark application datasets. Gradient Boosting can handle categorical variables directly without requiring them to be encoded as numerical values.     Chen et al. \cite{chen2016xgboost} reported that Gradient Boosting was effective in handling non-linear relationships and was less prone to overfitting than other boosting algorithms.   Gradient Boosting does not restrict itself to simple models such as decision trees and neural networks. It can also be used with deep learning models such as Convolutional Neural Networks \cite{walach2016learning} and other machine learning models\cite{badirli2020gradient}.

A limitation of Gradient Boosting is that it is slow to train on large datasets \cite{chen2016xgboost}. Since Gradient Boosting features a series of base models in an ensemble, it cannot be easily implemented with parallel computing. Furthermore,  Gradient Boost requires careful tuning of the hyperparameters, such as the number of base models and the learning rate. According to a study by Benntejac et al. \cite{bentejac2021comparative}, Gradient Boosting requires careful tuning of the parameters to achieve good performance. Gradient Boost is considered to be a black-box model, meaning it is difficult to interpret what is happening inside the model \cite{friedman2001greedy}.   Finally, Gradient Boosting learns the new model very quickly, which leads to over-fitting. This can be addressed by slowing the learning using a weighting factor for the corrections by new predictors (base models), known as the learning rate and shrinkage \cite{zeiler2012adadelta}.

 Some examples of Gradient Boosting applications are disease risk assessment \cite{ma2022retrieval}, credit risk assessment \cite{chang2018application}, mobility prediction \cite{semanjski2015smart}, anti-money laundering \cite{vassallo2021application}, optimisation of transduction \cite{oono2020optimization}, strength prediction \cite{shahani2021application}, travel time prediction \cite{zhang2015gradient}, modelling energy consumption \cite{touzani2018gradient}, geography \cite{georganos2018very, moisen2006predicting}, medical diagnosis \cite{liu2022early, liu2020early, shobana2021prediction}, image recognition \cite{zhang2013real}. It is worth mentioning that Gradient Boost has been widely used in \textit{Kaggle} \footnote{https://www.kaggle.com/} and other machine learning competitions. Examples of applications of Gradient Boosting for imbalanced data are spam filtering \cite{he2007asymmetric}, stroke diagnosis  \cite{lyashevska2021class}, modelling energy consumption \cite{touzani2018gradient}, fraud detection \cite{devi2017fraud}, anomaly detection \cite{tama2019depth}, and natural language processing \cite{natekin2013gradient}.

\subsubsection{ XGBoost}
 
XGBoost is a   Gradient Boosting method that uses advanced regularisation methods such as \textit{lasso} (L1) and \textit{ridge} (L2) regularisation to improve model generalisation capabilities \cite{chen2016xgboost}. XGBoost delivers better computational performance when compared to Gradient Boosting \cite{chen2015xgboost} since its training can be parallelised across clusters. The XGBoost loss function and regularisation at $t$th iteration are given by  Equation \ref{eq:xgloss}.

\begin{equation}   
L^{(t)} = \sum_{i=1}^n l(y_i,\hat{y_i}^{(t-1)}+f_t(x_i))+ \Omega(f_t)
\label{eq:xgloss}
\end{equation}

\begin{equation}
\Omega(f) + \gamma T +\frac{1}{2} \lambda \Sigma_{j=1}^T w_j^2
\label{eq:xgloss11}
\end{equation}
 
where

\begin{equation}
\hat{y_i}=\Sigma_{k=1}^K f_k(x_i), f_i \epsilon \mathcal{F}
\end{equation}

$K$ is the number of trees, and $f$ is the functional space of $F$ that represents the set of possible classification and regression trees.    XGBoost uses Taylor approximation to transform the original objective function to a function in the Euclidean domain in order to use traditional optimisation techniques. At iteration $t$, we need to train a model that achieves the maximum possible reduction of loss using  Equation \ref{eq:xgb}.

\begin{equation} 
\tilde{L}^{(t)}(q) = -\frac{1}{2}\sum_{j=1}^T \frac{(\sum_{i \in I_j}g_i)^2}{\sum_{i \in I_j}h_i+\lambda}+\gamma T
\label{eq:xgb}
\end{equation}

\begin{equation} 
 g_i = \partial_{\hat{y_i}^{(t-1)}} l (y_i, \hat{y}_i^{(t-1)}) 
\label{eq:xg1}
\end{equation} 
 
\begin{equation}  
h_i = \partial_{\hat{y_i}^{(t-1)}}^2 l (y_i, \hat{y}_i^{(t-1)}) 
\label{eq:xg2}
\end{equation}

XGBoost uses the \textit{exact greedy algorithm} that has complexity $O(n*m)$, where $n$ is the number of training samples and $m$ is the number of features. In the case of binary classification, XGBoost employs a log-loss objective function. XGBoost is also considered to be a strong model in \textit{Kaggle} competitions \cite{bojer2021kaggle} due to high accuracy and scalability.   XGBoost is a highly efficient and scalable implementation of Gradient Boosting, making it well-suited for large-scale datasets and high-dimensional problems. XGBoost has built-in support for parallel processing and distributed computing, allowing it to take full advantage of modern hardware. It has a variety of regularisation options, including L1 and L2 regularisation, which can help prevent overfitting.  

Similar to other ensemble learning methods, XGBoost has some disadvantages, as it can be sensitive to the choice of parameters and requires fine-tuning for optimal performance. XGBoost is a complex ensemble model with specialised trees similar to decision trees; however, it is an interpretable model that provides feature importance. XGBoost can be used to identify the most important features in a dataset, which can be useful for feature selection and understanding the underlying relationships in the data \cite{banga2021performance}. XGBoost  has been applied in various fields, such as energy consumption prediction \cite{wang2017electricity}, stock market prediction \cite{naik2021novel}, and sales forecasting \cite{shilong2021machine}.  XGBoost is commonly used for binary and multi-class classification problems and has been applied in various domains, such as rock classification \cite{zhang2017machine}, real-time accident detection \cite{parsa2020toward}, prediction of hourly air pollution (PM 2.5 concentration) \cite{pan2018application}, customer churn prediction \cite{tang2020customer, zhuang2018research}, credit risk analysis \cite{liu2022two, wang2022research, li2022hybrid}, and medical diagnosis \cite{ogunleye2019xgboost, li2020research}. 

 XGBoost handles class imbalance problems by using techniques such as weighting the loss function \cite{wang2020imbalance} and other built-in hyperparameters. In the medical field, XGBoost has been used for class imbalance problems, such as identifying rare diseases \cite{li2019xrare}, detecting rare adverse drug reactions\cite{nichols2019machine, letinier2021artificial} and multiple imbalanced datasets \cite{wang2020imbalance, hancock2020performance,zhang2022research,le2022xgboost, mishra2021dtcdwt,mushava2022novel,le2022xgboost}. In computer vision, it has been used for object detection, where the object of interest is present only in some images \cite{ranjan2017l2}.

\subsubsection{LightGBM}

 LightGBM  \cite{ke2017lightgbm} is a  GBM that uses tree-based learning and is called “Light” because of its computing power, as it requires less memory and can handle large amounts of data. LightGBM can easily overfit small data due to its sensitivity. It can be used for larger datasets, but no fixed threshold helps decide the usage of LightGBM.  LightGBM constructs a tree-based model in a forward stagewise manner, where at each stage, a new tree is added to the ensemble in order to reduce the residual error of the current ensemble. This is done by fitting the new tree to the negative gradient of the loss function using gradient-based optimisation.

 One of the key features of LightGBM is its use of a histogram-based algorithm for constructing decision trees, which is able to handle large numbers of features and samples efficiently. This algorithm is called \textit{gradient-based one-side sampling} (GOSS) \cite{ke2017lightgbm}  which works by randomly selecting a subset of the data for each tree split rather than using all the data. This allows the algorithm to be more efficient and to avoid over-fitting, especially when the number of features is very large. Another important feature of LightGBM is its use of a leaf-wise tree growth strategy, which is able to find the optimal split point in a leaf by using the histogram-based algorithm. This is different from traditional tree-based algorithms, which typically use a level-wise tree growth strategy. Similar to other tree-based algorithms, LightGBM can be prone to overfitting, especially when the number of trees is large. LightGBM can be seen as a more complex model when compared to other machine learning models due to the various hyperparameters that need to be tuned.

 LightGBM has been applied to a wide range of applications in various fields, involving predictive modelling tasks such as classification, regression, and ranking. It has been successfully used in various applications, such as natural language processing \cite{quinto2020next, qin2021natural, lin2020sentiment}, computer vision \cite{sun2022multi, zeng2019lightgbm}, and speech recognition \cite{xiwen2021speaker, yu2021speech, kannapiran2023voice}.  LightGBM has also been applied to application domains such as finance \cite{machado2019lightgbm, wang2022corporate, sun2020novel}, healthcare \cite{ghourabi2022security, shin2020emergency, chen2019prediction}, and marketing \cite{liang2019product, li2018monthly}.  

 In terms of imbalanced datasets, LightGBM features the use of different objective functions and the ability to assign different weights to different classes. LightGBM's ability to handle imbalanced datasets makes it an attractive option for a wide range of applications such as fraud detection \cite{huang2020optimized, ge2020credit, minastireanu2019light, hu2019short, hancock2021gradient}, anomaly detection \cite{islam2020network, yanabe2020anomaly, ekpo2022lightgbm}, and medical diagnosis \cite{liao2021study, zhang2022coronary}; where the imbalance is a common problem.

\subsection{Stacking and Voting}
 
\emph{Stacking} \cite{wolpert1992stacked} is an ensemble learning method that features a collection of models (base learners) on the same data and uses a meta-learner model to learn how to best combine the predictions \cite{rokach2010pattern}. 
The predictions from base learners (level-0 models) are used as input for meta-learners (level-1 model). The base learners typically give predictions as probabilities, which are used as input to meta-learners.
In principle, any learning scheme can be applied for meta learners; however, it is preferable to use “relatively global, smooth” models \cite{wolpert1992stacked}, since base learners do most of the work, this avoids the risk of overfitting.

\emph{Voting} is an ensemble learning method that involves summing the predictions in the case of classification and averaging the predictions in the case of regression. Typically, much better performance is achieved by stacking as compared to voting \cite{dvzeroski2002stacking}. There are two types of voting in classification problems - hard voting \cite{gandhi2015hybrid} and soft voting \cite{sherazi2021soft}. Hard voting means choosing the prediction with the highest number of votes, while soft voting means combining the probabilities of each prediction in each model and choosing the prediction with the highest overall probability. Models built using voting should not be considered a \textit{one-size-fits-all} approach in machine learning. The voting method has its limitations since there are cases where a single model outperforms a group of models, and since voting requires multiple models, they are naturally computationally intensive.

 Caruana et al. \cite{caruana2006empirical} compared the performance of different ensemble techniques, including voting and stacking, on various datasets and showed that ensemble methods perform better than individual models. Gaye et al. \cite{gaye2021tweet} demonstrated the effectiveness of stacking ensemble techniques for sentiment analysis. Belouch et al. \cite{belouch2017comparison} compared the performance of different ensemble techniques for network intrusion detection and found that stacking was better than bagging and boosting.  Al-Azani et al. \cite{al2017using} compared multiple voting and stacking models with and without data augmentation techniques on imbalanced datasets and found that stacking outperformed voting. Despite the prevalence of imbalanced datasets, there remains a lack of use of voting classifier applications.

\section{Data Augmentation for CI problems}

If the dataset for training the machine learning model is sufficiently balanced, the model performs better and more accurately. Data augmentation has the potential to improve the model performance of such models \cite{van2001art}. Data augmentation provides a set of techniques to synthetically increase the amount of data by generating new data points from existing data to address imbalance in classes. A simple strategy to artificially generate data would require minor alterations such as flip, transformation, or rotations in the case of image data. Data augmentation methods have been prominent for image data in deep learning \cite{shorten2019survey} and graph-based data \cite{ding2022data}. It has also been used for natural language processing problems \cite{feng2021survey,li2022data} such as text classification \cite{bayer2022survey}. Data augmentation methods have also been used for time series \cite{wen2020time} and tabular datasets for a wide range of applications.

\subsection{SMOTE-based methods} 

The focus of this paper is the CI problem and the combination of data augmentation and ensemble learning for tabular datasets. The most prominent approach in using data augmentation for CI datasets is generating examples in the minority class using \textit{synthetic minority oversampling technique} (SMOTE)
  \cite{chawla2002smote} that uses the \textit{k-nearest neighbour} (KNN) model \cite{cover1967nearest}.
SMOTE generates synthetic examples in a less application-specific manner, by operating in “feature space” rather than “data space.” We can increase the minority class by taking each minority class sample and introducing synthetic examples along the line segments joining any or all of the nearest neighbours of the minority class. We can choose neighbours for KNN depending on the amount of oversampling required. Although SMOTE was developed for classification problems, it can also be extended to regression problems \cite{torgo2013smote}. SMOTE has made an impact in various domains over the decade with a diverse set of applications \cite{fernandez2018smote}.

Furthermore, a number of alterations of SMOTE, such as geometric SMOTE \cite{douzas2019geometric} and weighted SMOTE \cite{prusty2017weighted} exist. In our study, we implement some of the prominent SMOTE-based methods discussed hereafter for implementing a computational review.

\subsubsection{SMOTE-ENN}

This hybrid technique combines SMOTE  with an undersampling method known as \textit{edited nearest neighbour}(ENN) \cite{dasarathy2000nearest}. ENN first finds the k-nearest neighbour of each observation and then checks if the majority class from the observation’s k-nearest neighbour is the same as the observation’s class. If both the classes are different, then the observation and its k-nearest neighbour are deleted. SMOTE-ENN  harnesses the strengths of the respective methods and has been successful in several domains. Xu et al. \cite{XU202010} used SMOTE-ENN for medical data using a hybrid sampling algorithm that combines the misclassification-oriented  SMOTE and ENN based on Random Forests.

\subsubsection{K-means SMOTE} 

K-means SMOTE \cite{douzas2018improving} consists of three steps: clustering, filtering, and oversampling. In the clustering step, we can cluster the input space using k-means clustering. The filtering step selects clusters for oversampling, retaining those with a
high proportion of minority class samples.  We can then distribute the number of synthetic samples assigning more
samples to clusters where minority samples are sparsely distributed. Finally, in the oversampling step, we can apply SMOTE   in each selected cluster to achieve the target ratio of minority and majority instances.

 K-means SMOTE has been used in several applications that include combination with ENN to handle noisy imbalanced data \cite{puri2022improved}, imbalanced medical data \cite{xu2021cluster}, and malicious domain detection \cite{wang2020malicious}. Land cover maps often suffer from class imbalance due to the nature of imbalance in remotely sensed data. Fonseca et al. \cite{Fonseca2021} demonstrated the efficacy of K-means SMOTE on 7 well-known imbalanced land cover datasets, and found that in almost 90\% of classification tasks, K-means SMOTE performed better than SMOTE and other related methods.

\subsubsection{SMOTE-SVM} 

 The \textit{support vector machine} (SVM) \cite{cortes1995support} is a prominent machine learning model used for classification originally and has also been extended to regression problems and applied to a wide range of domains \cite{noble2006support,salcedo2014support,mountrakis2011support}. More specifically, SVMs have been prominently used in a  number of challenging applications such as handwriting recognition \cite{bahlmann2002online}, and cancer genomics\cite{huang2018applications}. SVM finds a hyperplane to separate different classes in a $N$-dimensional space. Since SVM uses a few support vectors instead of all the training samples, the computational complexity is reduced significantly, along with the problems of dimensionality.  A disadvantage of SVM is that although it performs well on balanced datasets, its performance drops sharply on imbalanced datasets \cite{akbani2004applying}. This led to the combination of SVM with  SMOTE to address CI problems\cite{wang2017novel}, and since then, SMOTE-SVM has been prominent in a number of applications such as spatial datasets  \cite{vitianingsihapplication}.

\subsubsection{Borderline SMOTE}

Borderline SMOTE \cite{han2005borderline} attempts to learn the borderline between classes as exactly as possible, where the training examples on the borderline or near it are more likely to be misclassified.
The method employs SMOTE to calculate the k-nearest neighbours for all minority samples and selects random samples according to the oversampling rate. It then proceeds to generate new synthetic samples along the borderline between two classes, to strengthen the borderline minority examples \cite{han2005borderline}.
 
In terms of applications, Li et al. \cite{li2021application} used borderline-SMOTE for susceptibility assessment of debris flow, and found that using borderline SMOTE resulted in high accuracy of the susceptibility map. Smiti et al. \cite{smiti2020bankruptcy} presented an approach combining borderline SMOTE and stacked autoencoder based on the softmax classifier for bankruptcy prediction. Ning et al. \cite{ning2021novel} used borderline SMOTE with \textit{Tomek Links}  for the detection of glutarylation sites on protein residues.  Tomek Links \cite{tomek1976} is a modification from the \textit{condensed nearest neighbour} undersampling technique.

\subsection{Other Data Augmentation Methods}

\subsubsection{Adaptive Synthetic Sampling} 

  Adaptive Synthetic Sampling (ADASYN)\cite{4633969} adaptively generates minority data samples according to their distributions. In ADASYN, more synthetic data is generated for minority class samples that are harder to learn when compared to the minority samples that are easier to learn. This means that the bias introduced by class imbalance is reduced, and the decision boundary of the classifier has shifted towards the more difficult training examples. ADASYN has been used to address the CI problems in areas such as recreational water quality modelling \cite{xu2020predictive}, warning systems for harmful algae blooms \cite{kim2021improving} and design of wireless intrusion detection systems \cite{hu2020novel}.

\subsubsection{Random sampling methods}

\textit{Random over-sampling} (ROS) \cite{moreo2016distributional} and \textit{random under-sampling} (RUS) \cite{mohammed2020machine,hasanin2018effects} are naive approaches to combat imbalanced dataset in order to balance the class distribution. These are defined as naive techniques, as they simply duplicate the training samples. Random oversampling duplicates the samples from the minority class, while random undersampling deletes training samples from the majority class. However, these techniques often run into problems of overfitting when used with conventional machine learning models \cite{mohammed2020machine}. In terms of applications,   ROS has been used for malware detection \cite{pang2019signature}  and finite element methods \cite{calo2016randomized}. RUS has been used for web attacks \cite{zuech2021detecting}, feature selection in bioinformatics \cite{hasanin2019investigating}, and multi-label classification \cite{tahir2012inverse}.

\subsection{Generative Adversarial Networks}
 
Data augmentation has been prominent in image classification problems, in applications where a sufficient dataset of high-quality images is often missing. A commonly followed approach is to flip, rotate or randomly change the hue, saturation, brightness, and contrast of an image. This procedure is quite simple but has limited impact on generalizability compared to using GANs \cite{shorten2019survey} as it does not introduce new synthetic training samples into the dataset. Sauber-Cole et al. \cite{Sauber-Cole2022} provided an extensive survey of GANs for class imbalance, especially for tabular data.


Generative adversarial networks (GANs) \cite{goodfellow2020generative,creswell2018generative} have revolutionised the area of image generation, as they can generate realistic images and videos \cite{lu2018image}. GANs consist of 2 major parts that include the generator and the discriminator. The generator learns to generate realistic data from a batch of samples. It produces samples with identical distribution to that of the original dataset, which act as negative examples for the discriminator. The discriminator's role is to learn to distinguish synthetic data from real data. The advantage of GANs lies in their ability to generate diverse and quality synthetic data, which can be augmented to the original dataset for superior performance.

Although GANs have been prominent for image-related tasks, they can also be used for tabular data. Sharma et al. \cite{sharma2022smotified} developed a combination of GANs with SMOTE for addressing CI problems in tabular datasets.\textcolor{black}{Ding et al. \cite{ding2022imbalanced} presented a KNN and GAN-based hybrid approach for intrusion detection}. GANs have been applied to a wide range of problems which include synthetic brain medical image generation \cite{han2018gan}, text-to-image generation \cite{liao2022text}, and personal image generation \cite{ma2021must}. Wang et al. \cite{wang2021generative} presented a survey for GANs in computer vision problems.

\textcolor{black}{GANs have been applied for generating quality faulty state data in machines, as collecting faulty signals is difficult and expensive to collect~\cite{ZHANG2020107377}. Wang et al. \cite{Wang2022DualAtt} presented a novel dual-attention GAN, which selectively enhanced feature representation that outperformed existing methods for fault diagnosis. Wang et al. \cite{Wang_Liu_Zio_2022} and Li et al. \cite{Li2021GanRotate} presented methods for detecting fault diagnosis in rotating machines and fault diagnosis in high-speed train components, respectively. Mao et al. \cite{Mao2019GanStudy} presented a comparative study on imbalanced fault diagnosis in ball bearings based on GANs.}

 \textcolor{black}{Conditional GAN (CT-GAN)  provides synthetic data generation with high fidelity for single tabular data \cite{xu2019modeling}. CT-GANs were designed to address the problem of generating realistic
synthetic data for tabular data that usually contains a mix of discrete
 and continuous columns, where continuous columns can have multiple modes whereas
discrete columns feature imbalance. CT-GANs have been demonstrated to model distributions more than Bayesian networks, where mode-specific normalisation
has the ability to convert continuous values into a bounded vector representation
suitable for neural networks. The conditional generator can help
generate data with a specific discrete value that can be used for data augmentation for class imbalance. In our framework, we use CT-GANs as a novel data augmentation method for class imbalance problems. }

\subsection{Data augmentation for CI problems: recent progress}

We next review some of the recent strategies that used data augmentation methods to address class imbalance problems.  Emu et al. \cite{Emu2022IEEE} introduced a unique oversampling strategy for obtaining a balanced dataset, where the minority class was increased in such a way that the misclassification rate of the majority class remained low. Cai et al. \cite{CAI202250} proposed a novel clustering algorithm called  \textit{single and multi-source based adaptive grid partition and decision graph} to mine clusters in an imbalanced data set simultaneously, even if there are large density differences infused data set. Ofek et al. \cite{OFEK201788} developed a fast clustering-based undersampling technique based on the sample set partitioning approach, splitting the original training dataset into pairwise disjoint subsets and training classifiers on them.

Lin et al. \cite{LIN201717} proposed two strategies to undersample the data using clustering. The first strategy employed k-means clustering where \textit{k} was equal to the number of samples of the minority class, and the centroids of the clusters replaced the original majority class samples. In the second strategy, instead of replacing the samples with the centroids of the clusters, the method replaced the samples with the nearest neighbour of the centroids. Several combinations of the clustering-based undersampling approach with different classifiers were tested on small-scale and large-scale datasets for breast cancer and protein homology prediction, and it was observed that the second strategy outperformed 5 state-of-the-art approaches. Chamseddinr et al. \cite{chamseddine2022handling} used SMOTE with deep learning models for a multi-class classification task that distinguished between COVID-19, normal, and viral pneumonia cases. In order to address the class imbalance issue, the authors first investigated the use of a weighted categorical loss mechanism and then used SMOTE on 3 COVID-19 datasets.
Hu et al. \cite{HuMSMOTE2009} presented a modified version of SMOTE, wherein the minority samples were classified into 3 categories, i.e. security samples, border samples and latent noise samples. The algorithm used different strategies for selecting neighbours of each sample; selecting a random data point from the $k$ nearest neighbours of the security sample, the nearest neighbour from the border sample, but nothing for the latent noise sample. The algorithm was tested on UCI imbalanced datasets and it outperformed the traditional SMOTE algorithm.  Nanni et al. \cite{NANNI201548} used a modified version of SMOTE along with a \textit{reduced reward-punishment} technique \cite{NANNI20112395} for undersampling to develop a novel ensemble classifier.

Mediavilla et al. \cite{mediavilla2022imbalance} proposed a 2-step Bayesian methodology to address the CI problem, due to its formulation allowing the inclusion of the individual example costs in the classification and taking into account the class probabilities. This is an example-dependent cost classification that provides an advantage in less favourable situations, such as cases with a strong imbalance and highly nonlinear classification borders. Fu et al. \cite{fu2023automatic} addressed the CI problem and insufficient training samples in grading \textit{diabetic macular edema} which is one of the main causes of permanent blindness. They added class weight to the loss function and data augmentation techniques to expand the data.  

Sun et al. \cite{SUN20151623} proposed a novel ensemble method that converts an imbalanced dataset into several balanced datasets. The majority class samples were first divided into bins with a number of instances in each bin equal to that of the instances in the minority class, creating several balanced datasets. Each dataset was used to create a binary classifier, which was then combined into an ensemble classifier. Sun et al. proposed a method to handle binary-class imbalance problems, outperforming conventional methods such as SMOTEBoost, EasyEnsemble, and RUSBoost. Rahman et al. \cite{abd2022waveguide}  used data augmentation for resampling quantum cascade laser images, along with ensemble learning and deep learning models such as  DenseNet, MobileNet, and ResNet-50. Alsulami et al. \cite{alsulami2022toward} presented an ensemble-based decision-making phase to address the limitations of a single classifier by relying on the predictions of multi-classifiers. Kamalov et al. \cite{kamalov2022kde} developed an ensemble model that trained each tree (model) in the ensemble using uniquely generated synthetically balanced data. 

Hajek et al. \cite{hajek2022fraud} developed a semi-supervised ensemble model integrating multiple unsupervised outlier detection algorithms. They found that while XGBoost achieved the best results, the highest cost saving was achieved by combining random under-sampling with XGBoost. Gao et al. \cite{gao2020identification} presented an ensemble learning method that combined different algorithms including borderline SMOTE,  ADASYN, SMOTE-Tomek, and SMOTE-ENN with the XGBoost for gene identification. They reported that the ensemble classifiers outperformed single classifiers, and XGBoost combined with SMOTE-ENN performed the best. Dai et al.  \cite{dai2022class} developed  \textit{three-line hybrid positive instance augmentation} that combined the features of majority-class and minority-class to construct unlabelled instances, which was effective in boosting the diversity of generative pseudo-positive instances.  

\subsubsection{One-Class Classification}

\textcolor{black}{One-class classification (OCC)  has been an effective way to counter CI problems \cite{KHAN2010OCC,seliya2021literature}. OCC performs well when the negative class is either absent, poorly sampled or not well-defined. These models are not built to distinguish between classes, but to only identify the majority class~\cite{khan_madden_2014}. } Runchi et al. \cite{runchi2023ensemble} designed a novel ensemble model that utilised data preprocessing, and then applied a data (instance) balancing algorithm to generate several training subsets with different imbalance ratios to develop sub-models, respectively. Furthermore,  the weight of predicted results for different classes of each sub-model was dynamically updated according to the performance of each sub-model in the validation stage. The results indicated that the ensemble model had the ability to recognise default samples, with the best generalisation ability on most of the given datasets.   Gao et al. \cite{GAO2020101935} presented a deep learning-based OCC to handle medical imaging data that demonstrated promising results in datasets from the cellular level to the tissue level, as well as different image modalities.

\subsubsection{Online Learning}

\textcolor{black}{We find it essential to cover \textit{online learning} \cite{ben1997online} also known as  \textit{sequential learning} in relation to CI problems. In machine learning, online learning is a paradigm where data is used to train the model as it becomes available in a sequential order \cite{HOI2021249}.     Conventional machine learning trains models by using the entire training data set (batch learning) at once. In real-world applications such as credit card fraud detection \cite{wang2013online}, spam filters \cite{li2018comparative}, and recommendation systems \cite{bobadilla2013recommender},    data is available in continuous streams where models need to be trained and updated iteratively \cite{HOI2021249}.  Data streams for online learning often face problems of skewed data distributions, with class imbalance ratios changing constantly. This degrades the performance of machine learning algorithms, as the lack of prior knowledge about data classes makes it difficult to determine which samples belong to the minority or majority class. Online learning has been shown to be computationally more efficient than traditional batch learners \cite{Liang2006,Shilton2005}.  Wang et al. \cite{Wang2018} presented a systematic study and framework of existing techniques for CI in online learning with concept drift. Wang et al. \cite{wang2016multiclass}   extended CI in online learning for multi-class problems, and developed ensemble learning techniques. Concept drift is another issue that arises for online learning models, where changes in the underlying distribution of the incoming data require models to track such changes and adapt \cite{Widmer1996, tsymbal2004problem}. Mirza et al. \cite{MIRZA2015316, MIRZA201679} presented novel methods to address CI with concept drift in online sequential learning.
}


\section{Methodology: combination of ensemble and data augmentation methods}

In this study, our goal is to evaluate combinations of prominent machine learning and data augmentation methods for CI problems. Hence, we present our methodology that 
 evaluates our selected combinations for benchmark datasets for a computational review.  In this section, we discuss the techniques and datasets used for the review.

\subsection{Evaluation Framework}

We select nine prominent data augmentation techniques that include SMOTE, SMOTE-ENN, Borderline SMOTE, SVM-SMOTE, ADASYN, RUS, ROS, and CT-GAN. We also evaluate the selected ensemble learning methods without data augmentation so that we can evaluate the role of data augmentation in imbalanced class datasets. 

\begin{figure}      
\centering
\subfigure[Stacking-I]{
    \includegraphics[height=4.5cm]{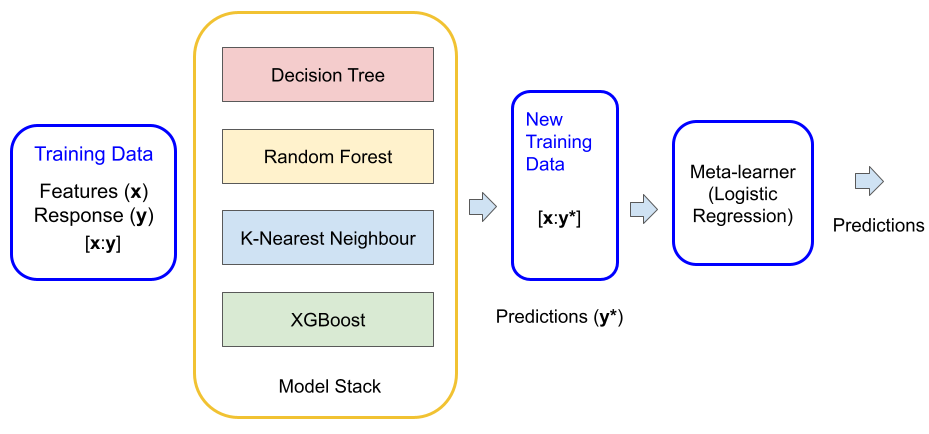}%
}
\subfigure[Stacking-II]{
    \includegraphics[height=3cm]{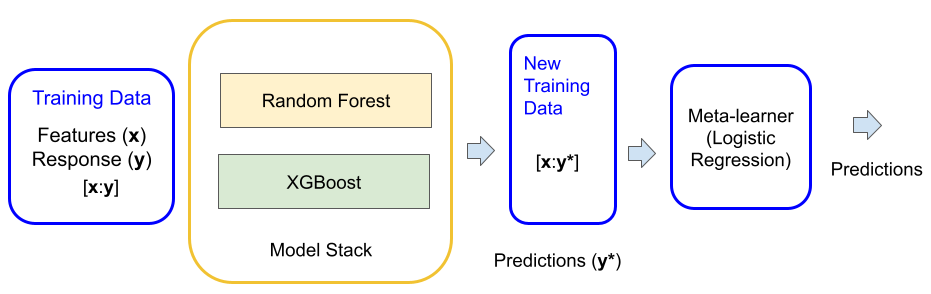}%
    }
\caption{Ensemble learning architecture for Stacking-I and Stacking II.}
\label{fig:SC}
\end{figure}

\begin{figure}
\centering
\subfigure[Voting-I]{
    \includegraphics[height=4cm]{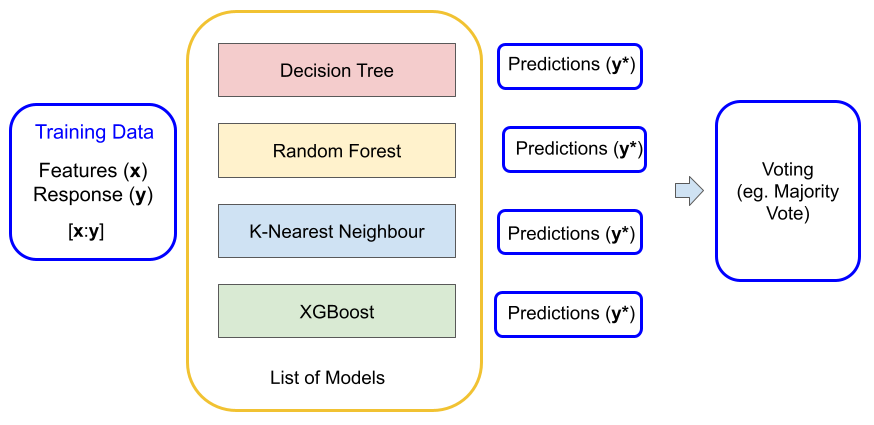}%
}
\subfigure[Voting-II]{
    \includegraphics[height=3cm]{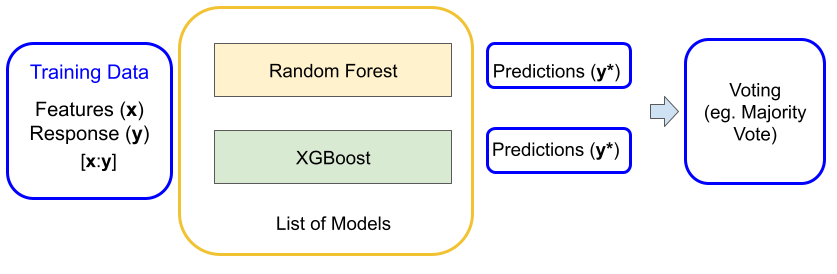}%
    }
\caption{Ensemble learning architecture for Voting-I and Voting II.}
\label{fig:VC}
\end{figure}

We use ten different ensemble learning methods that are based on bagging, boosting, stacking, and voting. The ensemble learning methods include LightGBM, AdaBoost, XGBoost, Gradient Boosting,  Random Forest, Voting-Soft, Voting-Hard, Stacking-I, and Stacking-II.  
Figure \ref{fig:SC} (Panel a) shows Stacking-1 that features Decision Trees, Random Forests, $K$-Nearest Neighbours, and XGBoost in the Model Stack, and Logistic Regression as a meta-learner which makes the final prediction by using predictions from base learners as features.   Stacking-II in Figure \ref{fig:SC}) (Panel b) features Random Forests and XGBoost as base learners in the Model Stack and Logistic Regression as the meta-learner. These base learners with logistic regression as a meta-learner are standard approaches used in the literature \cite{abro2020stacking, liu2022predictive, yoon2023multi, jayapermana2022implementation}.  Furthermore, Voting-I employs  Decision Trees, Random Forest, $K$-Nearest Neighbours and XGBoost in the Model Stack as shown in Figure \ref{fig:VC}.  Voting-II shown in Figure \ref{fig:VC} employs  Random Forest and XGBoost in the Model Stack.


\begin{figure*}
\centering
\includegraphics[width=18cm]{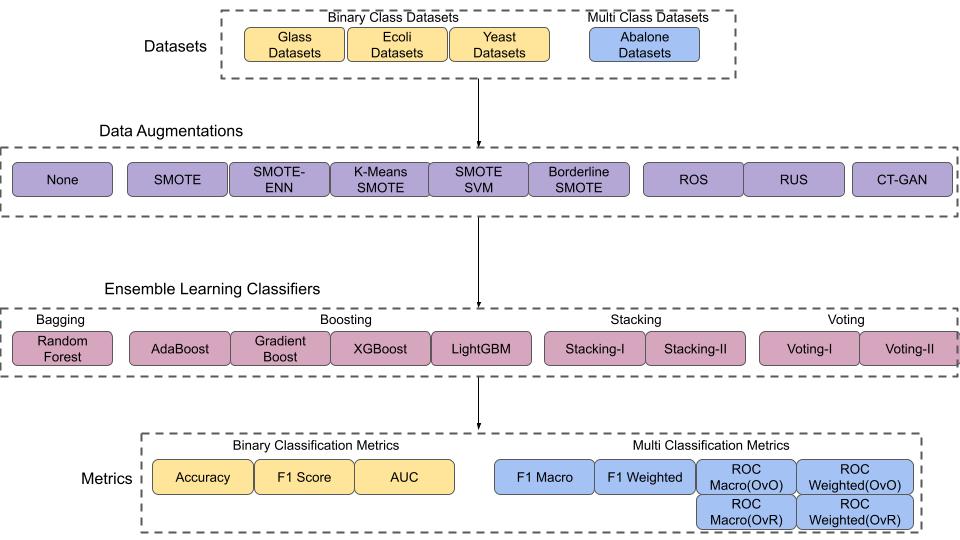}%
\caption{The evaluation framework  showing the family of datasets, data augmentation methods, ensemble learning methods,  and evaluation metrics for binary and multi-class classification tasks.}
\label{framework}
\end{figure*}

\textcolor{black}{In Figure \ref{framework}, we present the evaluation framework, which features the family of datasets, data augmentation methods, ensemble learning methods, and   evaluation metrics for binary and multi-class classification tasks.}

\subsection{Evaluation Metrics}

In this study, we focus on binary CI datasets that feature a positive (minority) class and a negative (majority) class. We define the imbalance ratio (IR) as the number of negative class data instances divided by the number of positive classes in order to categorise the different datasets.


We use the \textit{classification accuracy}, \textit{F1 score}, and \textit{area under the curve} (AUC) of the \textit{receiver operating characteristic} (ROC) curve \cite{
bradley1997use,faraggi2002estimation,flach2015precision}   to evaluate the performance of the respective models. The classification accuracy fails to capture model performance in CI problems, especially in the case of the minority classes \cite{
luque2019impact,hossin2015review}, since it does not account for the difference in the size of the classes. Hence, the classification accuracy may lead to erroneous conclusions for imbalanced datasets; i.e. for a dataset with an imbalance ratio of 9.5, an accuracy of 95\% could mean that the model has misclassified all the members of the minority class.

The  AUC measures the ability of a classifier to distinguish between classes and is used as a summary score of the ROC curve.  We define the classification accuracy $\alpha $ of a model using the number of true positives (TP), true negatives (TN), false negatives (FN),  and false positives (FP) as shown in Equation \ref{eq:class}.
\begin{equation}
    \centering
    \alpha = \frac{TP + TN}{TP + FN + FP + TN}
\label{eq:class}
\end{equation}

Although the AUC-ROC curve is an evaluation metric for binary classification problems, it can also be extended to multi-class classification. The AUC-ROC curve plots the True Positive Rate (TPR)   against the False Positive Rate (FPR) at various threshold values. 

\begin{equation}
    \centering
    TPR = \frac{TP}{TP + FN}
\label{eq:tpr}
\end{equation}

\begin{equation}
    \centering
    FPR = \frac{FP}{FP + TN}
    \label{eq:fpr}
\end{equation}

\begin{equation}
    \centering
    AUC = \frac{1 + TPR - FPR}{2}
 \label{eq:fpr}
\end{equation}

The higher the AUC, the better the model distinguishes between classes. A model with an AUC of 1 would be a perfect classifier. Similarly, a model with an AUC of 0 would classify all negatives as positives and all positives as negatives.

Furthermore, other measures need to be taken into account that have been prominent for CI problems. The \textit{precision} ($\hat{p}$) is a measure of the proportion of positive predictions that were correctly classified, while the \textit{recall} ($\hat{r}$)  is the proportion of the actual positive data that were classified correctly, as given below:

\begin{equation}
    \centering
    \hat{p} = \frac{TP}{TP + FP}
 \label{eq:pre}
\end{equation}

\begin{equation}
    \centering
\hat{r} = \frac{TP}{TP + FN}
 \label{eq:recall}
\end{equation}

The precision and recall do not consider true negatives; hence, they are unaffected by class imbalance and pose as good metrics to evaluate the performance of a model. The F1 score is the harmonic mean of precision and recall and combines them into a single metric as given below:

\begin{equation}
    \centering
 F1 = 2\frac{\hat{p} * \hat{r}}{\hat{p} + \hat{r}}
 \label{eq:f1}
\end{equation} 

The F1 score assesses both types of errors; errors caused by false positives and errors caused by false negatives \cite{espindola2005extending}.

\subsubsection{Metrics for multi-class classification}

\textcolor{black}{As noted earlier, the prominent metrics that provide valuable insights into the model performance of CI problems are the F1 and AUC scores. Since we are dealing with multiple classes, the AUC score can be computed for both One-Versus-One (OvO) and One-Versus-Rest (OvR) strategies \cite{rocha2013multiclass}. In OvO, we compute the  AUC  for each pair of the respective class and take the mean, which provides a pairwise assessment of classification quality. In contrast, OvR we compute the AUC  for each class against the rest, effectively treating each class as a binary classification problem. The resulting scores offer insights into how well the model can differentiate each class from all others, aiding in the performance evaluation for multi-class models. We use these metrics for the evaluation of multi-class problems in our study.}

\textcolor{black}{In the case of multi-class problems \cite{van2008multi}, there are variations in the F1 score, which include the Macro-F1, Micro-F1, and Weighted-F1 \cite{grandini2020metrics}.   The Macro-F1 score computes the average F1 score across all classes, treating each class equally. This metric is useful when each class holds equal importance in the application. On the other hand, the Micro-F1 score calculates the aggregate F1 score by considering all instances as a single (collective) classification problem, which is valuable in cases when the class imbalance ratio is significant. Lastly, the Weighted-F1 score computes the average F1 score, but weights each class's contribution according to its frequency in the dataset, making it effective for imbalanced class distributions.}

\subsubsection{Classification Threshold Value}

\textcolor{black}{Most machine learning models produce a probability associated with each sample, which must be mapped to a crisp class label. Typically, the default value is 0.5 above which a sample is marked as belonging to one class and lower values are marked to another class. This threshold is called the classification value. The TPR and FPR are used to calculate the AUC score, but there arises an issue since  TPR and FPR both depend on the classification threshold value.  In our study,  we use the default classification threshold in the \textit{scikit-learn library}, which is 0.5. We note that such thresholds have important implications for different applications, and caution must be taken especially in the case of medicine, as discussed by Ruopp et al. \cite{ruopp2008youden}.}

\section{Computational Study}

\subsection{Experimental Setting}

 We evaluate every combination (ensemble learning model plus the data augmentation method) on the selected benchmark CI datasets.   
In our results, we provide the mean, the best, and the standard deviation of thirty independent experiment runs for model training. \textcolor{black}{We execute the data augmentation in each run, prior to the classification step of the framework (Figure \ref{framework}). We define the train-test ratio to 60:40, where in each experiment, we shuffle the data randomly based scikit-learn library \cite{scikit-learn}. We note that in the case of deep learning models such as GANs; in different experimental runs, we used random initial weights and biases.
} We use the framework in Figure \ref{framework} and utilise the default hyper-parameters for the respective models using the scikit-learn library.  In Table \ref{hyperPara}, we present the hyperparameters used by the respective models in our experiments. In the case of the CT-GAN, we use three hidden layers with 50, 25, and 10 nodes. We use the ReLU activation function in hidden layers with mean squared error (MSE) loss, and the Adam optimiser \cite{kingma2014adam} with the maximum training time of 1000 epochs.

\begin{table*}[htbp]
\small
\centering
\caption{Hyperparameters from the scikit-learn library \cite{scikit-learn} used in the computational study.}
\label{hyperPara}
\begin{tabular}{ll}
\hline
Dataset & Instances \\
\hline
\multicolumn{2}{c}{Ensemble Learning Models}\\
\hline
LghtGBM   & max\_depth = -1, min\_data\_in\_leaf = 20, bagging\_fraction = 1 \\
XGBoost Classifier & booster = gbtree, gamma=0, max\_depth=6, min\_child\_weight=1, max\_leaves=0,\\
    &   num\_parallel\_tree = 1, grow\_policy = depthwise\\
AdaBoost   & n\_estimators=50, learning\_rate=1.0 \\ 
Gradient Boosting & learning\_rate=0.1, n\_estimators=100, max\_depth=3, min\_samples\_split=2, min\_samples\_leaf=1 \\ 
Random Forest  & n\_estimators=100, criterion='gini', max\_depth=None, max\_leaf\_nodes=None, n\_jobs=None \\
\hline
\multicolumn{2}{c}{Data Augmentation Techniques}\\
\hline
SMOTE & sampling\_strategy='auto', k\_neighbors=5, n\_jobs=None \\
SMOTE-ENN & sampling\_strategy='auto', smote=None, enn=None, n\_jobs=None \\
Borderline SMOTE & sampling\_strategy='auto', k\_neighbors=5, n\_jobs=None, m\_neighbors=10, kind='borderline-1 \\
SMOTE-SVM & sampling\_strategy='auto', k\_neighbors=5, n\_jobs=None, m\_neighbors=10, svm\_estimator=None, out\_step=0.5 \\
K-means-SMOTE & sampling\_strategy='auto', k\_neighbors=2, n\_jobs=None, kmeans\_estimator=None,\\
 & cluster\_balance\_threshold='auto', density\_exponent='auto'\\
ADASYN & sampling\_strategy='auto', n\_neighbors=5, n\_jobs=None \\
CT-GAN &  max\_depth: 2, max\_bin: 100, learning\_rate: 0.02, n\_estimators: 500\\

\hline
\end{tabular}
\end{table*}

\subsection{Datasets}

 In this study, we modified the Glass, Yeast, and Ecoli datasets taken from the UCI data repository \cite{Dua:2019}. We modified the datasets to develop binary classification datasets with different imbalance ratios, as shown in Table \ref{binaryDatasets}. We rank the datasets in decreasing order of imbalance ratios, where the lower imbalance ratio implies more class imbalance in the dataset.  We have adopted the naming conventions from the previous evaluation study by Galar et al.  \cite{galar2011review}, as evident in the dataset nomenclature such as  Ecoli-0137vs26. This nomenclature signifies a binary transformation, designating classes 0, 1, 3, and 7 as 'positive' and classes 2 and 6 as 'negative.' 

\begin{table*}[htbp]
\small
\centering
\caption{Description of the binary class datasets used in the computational study}
\label{binaryDatasets}
\begin{tabular}{llllllll}
\hline
Dataset & Instances & Attributes & Minority Class & Majority Class & \%Minority & \%Majority & Imbalance Ratio \\
\hline
    Yeast-6 & 1484 & 8 & EXC & remaining classes & 2.36 & 97.64 & 41.38 \\
    Ecoli-0137vs26 & 282 & 7 & omL,imS & cp,im,imU,om & 2.48 & 97.52 & 39.32 \\
    Yeast-5 & 1484 & 8 & EXC,ERL & remaining classes & 2.70 & 97.3 & 36.03 \\
    Yeast-1289vs7 & 947 & 8 & VAC & NUC,CYT,ERL,POX & 3.17 & 96.83 & 30.55 \\
    Yeast-4 & 1484 & 8 & ME2 & remaining classes & 3.44 & 96.56 & 28.07 \\
    Yeast-2vs8 & 483 & 8 & POX & CYT & 4.14 & 95.86 & 23.15 \\
    Glass-5 & 214 & 9 &  6 & remaining classes & 4.2 & 95.8 & 22.8 \\
    Yeast-1458vs7 & 693 & 8 & VAC & NUC,ME3,ME2,POX & 4.33 & 95.67 & 22.09 \\
    Glass-016vs5 & 184 & 9 & 6 & 1,2,7 & 4.89 & 95.11 & 19.45 \\
    Ecoli-4 & 336 & 7 & om & remaining classes & 5.95 & 94.05 & 15.8 \\
    Glass-6 & 214 & 9 &  5 & remaining classes & 6.07 & 93.93 & 15.5 \\
    Yeast-1vs7 & 459	& 8 & VAC & NUC & 6.54 & 93.46 & 14.29 \\
    Glass-5vs12 & 159 & 9 & 5 & 1,2 & 8.18 & 91.82 & 11.26 \\
    Ecoli-3 & 336 & 7 & cp & remaining classes & 10.42 & 89.58 & 8.6 \\
    Yeast-3 & 1484 & 8 & ME3 & remaining classes & 10.98 & 89.02 & 8.11 \\
    Glass-2 & 214 & 9 &  7 & remaining classes & 13.55 & 86.45 & 6.38 \\
    Ecoli-2 & 336 & 7 & pp & remaining classes & 15.48 & 84.52 & 5.46 \\
    Ecoli-1 & 336 & 7 & im & remaining classes & 22.92 & 77.08 & 3.36 \\
    Glass-123vs567 & 214 & 9 & 567 & remaining classes & 23.83 & 76.17 & 3.20 \\
    Yeast-1 & 1484 & 8 & NUC & remaining classes & 28.9 & 71.1 & 2.46 \\
    Glass-0 & 214 & 9 & 1 & remaining classes & 32.71 & 67.29 & 2.09 \\
    Ecoli-0vs1 & 220 & 7 & im & cp & 35 & 65 & 1.86 \\
    Glass-1 & 214 & 9 &  2 & remaining classes & 35.51 & 64.49 & 1.82 \\

\hline
\end{tabular}
\end{table*}

\begin{table*}[htbp]
\small
\centering
\caption{Description of the multi-class datasets used in the experimental study}
\label{multiDatasets}
\begin{tabular}{llllllll}
\hline
Dataset & Instances & Attributes & Class 1 & Class 2 & Class 3 & Class 4 & Description \\
\hline
Abalone-v1 & 4177 & 8 & 189 & 2577 & 1186 & 225 & Abalone-1vs2vs3vs4  \\  
Abalone-v2 & 4177 & 8  & 2730 & 1186 & 261 & - & Abalone-12vs3vs4\\ %
Abalone-v3 & 4177 & 8  & 2730 & 1411 & 36  & - & Abalone-12vs34  \\ 
Abalone-v4 & 4177 & 8  & 1600 & 2541 & 36 & - & Abalone-134vs2 \\ 
\hline
\end{tabular}
\end{table*}




\begin{table*}[htbp]
\footnotesize
\centering
\caption{Test results for glass datasets}
\label{tab1}
\begin{tabular}{lllll}
\hline
Data Augmentation & Ensemble Model & Accuracy (Best, Std) & F1 (Best, Std) & AUC (Best, Std)\\
\hline
\multicolumn{5}{c}{Glass-0}\\
\hline
    SMOTE-ENN & Random Forest  & 100.000(100.000, 0.000) & 100.000(100.000, 0.000) & 100.000(100.000, 0.000) \\
    RUS &	Random Forests &	99.225(100.000, 1.072) &	99.435(100.000, 0.782) &	98.810(100.000, 1.647) \\
    SMOTE-SVM &	XGBoost &	98.721(100.000, 1.196) &	99.068(100.000, 0.866) &	98.036(100.000, 1.837) \\ 
    No Augmentation &	Voting-Hard &	97.713(98.837, 0.212) &	98.333(99.145, 0.153) &	96.488(98.214, 0.326) \\ 
    SMOTE &	Gradient Boost &	97.674(97.674, 0.000) &	98.305(98.305, 0.000)  &	96.429(96.429, 0.000) \\
    Borderline-SMOTE &	Random Forests &	97.674(97.674, 0.000) &	98.305(98.305, 0.000) &	96.429(96.429, 0.000) \\
    ADASYN &	Gradient Boost	 & 97.674(97.674, 0.000) &	98.305(98.305, 0.000) &	96.429(96.429, 0.000) \\
    CT-GAN &	Voting-Hard &	97.248(97.674, 0.715) &	97.931(98.246, 0.528) &	97.652(98.276, 1.024) \\
    SMOTE-ENN &	Stacking-II &	96.512(96.512, 0.000) &	97.479(97.479, 0.000) &	94.643(94.643, 0.000) \\
    ROS &	XGBoost &	96.512(96.512, 0.000) &	97.479(97.479, 0.000) &	94.643(94.643, 0.000) \\
    CT-GAN &	AdaBoost &	93.023(93.023, 0.000) &	94.737(94.737, 0.000) &	92.980(92.980, 0.000) \\
\hline
\multicolumn{5}{c}{Glass-6}\\
\hline
    ROS &	AdaBoost  &	99.981(100.000, 0.150) &	79.970(99.363, 27.641) &	99.881(100.000, 0.922) \\
    SMOTE-ENN &	Voting-Hard &	99.385(100.000, 1.155) &	99.670(100.000, 0.619) &	96.247(100.000, 7.094) \\ 
    SMOTE &	Random Forest &	98.947(100.000, 1.270) &	99.435(100.000, 0.679) &	93.532(100.000, 7.799) \\
    Borderline-SMOTE &	Random Forests &	98.921(100.000, 1.289) &	99.421(100.000, 0.690) &	93.373(100.000, 7.920) \\ 
    Borderline-SMOTE &	Gradient Boosting  &	98.895(100.000, 1.364) &	99.408(100.000, 0.729) &	93.214(100.000, 8.378) \\
    Borderline-SMOTE &	AdaBoost  &	98.372(100.000, 1.688) &	99.130(100.000, 0.902) &	90.000(100.000, 10.372) \\
    SMOTE &	LightGBM  &	97.016(100.000, 0.900) &	98.404(100.000, 0.476) &	81.667(100.000, 5.528) \\
    CTG &	XGBoost &	92.636(93.023, 0.551) &	95.597(95.597, 0.000) &	69.952(70.163, 0.300) \\
    RUS &	Stacking-II &	74.725(98.837, 24.096) &	77.264(99.363, 30.218) &	84.051(99.367, 13.641) \\
    RUS &	LightGBM  &	8.140(8.140, 0.000) &	0.000(0.000, 0.000) &	50.000(50.000, 0.000) \\
\hline
\multicolumn{5}{c}{Glass-1}\\
\hline
    Borderline-SMOTE &	LightGBM &	99.070(100.000, 0.831) &	99.269(100.000, 0.651) &	98.750(100.000, 1.116) \\ 
    SMOTE-ENN &	XGBoost &	96.512(96.512, 0.000) &	97.143(97.143, 0.000) &	97.222(97.222, 0.000) \\
    ROS &	AdaBoost &	96.512(96.512, 0.000) &	97.196(97.196, 0.000) &	96.586(96.586, 0.000) \\
    ROS &	Random Forests &	96.279(97.674, 0.885) &	97.053(98.182, 0.694) &	95.849(98.148, 1.142) \\
    CT-GAN &	Voting-Soft &	95.659(96.512, 0.962) &	96.513(97.196, 0.761) &	95.673(97.222, 1.167) \\ 
    Borderline-SMOTE &	XGBoost	 & 95.349(95.349, 0.000) &	96.226(96.226, 0.000) &	95.660(95.660, 0.000) \\
    ROS &	Voting-Soft	 & 95.349(95.349, 0.000) &	96.226(96.226, 0.000) &	95.660(95.660, 0.000) \\ 
    RUS &	LightGBM &	95.000(100.000, 3.375) &	95.956(100.000, 2.800) &	94.873(100.000, 3.195) \\ 
      No Augmentation &	Voting-Hard &	61.318(63.953, 1.010) &	55.467(59.740, 1.676) &	69.198(71.296, 0.804) \\
\hline

\multicolumn{5}{c}{Glass-5vs12}\\
\hline
    SMOTE &	LightGBM &	100.000(100.000, 0.000) &	100.000(100.000, 0.000) &	100.000(100.000, 0.000) \\
    SMOTE-ENN &	Gradient Boost   &	100.000(100.000, 0.000) &	100.000(100.000, 0.000) &	100.000(100.000, 0.000) \\
    SVM-SMOTE &	Voting-Hard &	100.000(100.000, 0.000) &	100.000(100.000, 0.000) &	100.000(100.000, 0.000) \\
    ROS &	AdaBoost   &	100.000(100.000, 0.000) &	100.000(100.000, 0.000) &	100.000(100.000, 0.000) \\
    ROS &	Random Forests &	99.957(100.000, 0.257) &	100.000(100.000, 0.000) &	99.769(100.000, 1.373) \\
    ROS &	Voting-Hard &	99.941(100.000, 0.297) &	99.747(100.000, 1.498) &	99.688(100.000, 1.587) \\
    Borderline-SMOTE  &	Voting-Soft &	99.769(100.000, 0.556) &	98.658(100.000, 3.232) &	98.770(100.000, 2.963) \\ 
    No Augmentation &	Gradient Boost   &	99.609(100.000, 0.679) &	97.727(100.000, 3.953) &	97.917(100.000, 3.624) \\ 
    RUS &	Stacking-I &	96.128(100.000, 5.039) &	73.970(100.000, 33.736) &	91.638(100.000, 15.372) \\
    RUS  &	AdaBoost  &	93.542(100.000, 4.281) &	0.000(0.000, 0.000) &	73.027(100.000, 23.542) \\
    CTG &	LightGBM &	89.062(89.062, 0.000)  &	99.697(100.000, 1.635) &	56.609(56.609, 0.000) \\
\hline
\multicolumn{5}{c}{Glass-0123vs567}\\
\hline
     No Augmentation &	LightGBM &	100.000(100.000, 0.000) &	100.000(100.000, 0.000) &	100.000(100.000, 0.000) \\
     No Augmentation &	Stacking-I &	100.000(100.000, 0.000) &	100.000(100.000, 0.000)  &	100.000(100.000, 0.000) \\
    Borderline-SMOTE &	Voting-Hard &	100.000(100.000, 0.000) &	100.000(100.000, 0.000) &	100.000(100.000, 0.000) \\
    ADASYN &	Voting-Soft &	100.000(100.000, 0.000) &	100.000(100.000, 0.000)  &	100.000(100.000, 0.000) \\
    ROS &	Random Forests &	100.000(100.000, 0.000) &	100.000(100.000, 0.000) &	100.000(100.000, 0.000) \\
    CTG &	Gradient Boosting   &	96.473(98.837, 2.179) &	92.242(97.561, 5.045) &	93.570(97.619, 4.054) \\
    SVM-SMOTE &	Random Forests &	99.496(100.000, 0.984) &	98.987(100.000, 1.993)	 & 99.532(100.000, 1.195) \\ 
    SVM-SMOTE &	LightGBM&	99.845(100.000, 0.402) &	99.690(100.000, 0.804) &	99.897(100.000, 0.266) \\ 
    SMOTE-ENN &	Stacking-II &	99.981(100.000, 0.149) &	99.959(100.000, 0.313) &	99.960(100.000, 0.305) \\
\hline
\multicolumn{5}{c}{Glass-2}\\
\hline
    SMOTE &	LightGBM &	100.000(100.000, 0.000) &	100.000(100.000, 0.000) &	100.000(100.000, 0.000) \\
    SMOTE-ENN &	AdaBoost   &	100.000(100.000, 0.000) &	100.000(100.000, 0.000) &	100.000(100.000, 0.000) \\
    Borderline-SMOTE &	LightGBM &	100.000(100.000, 0.000) &	100.000(100.000, 0.000) &	100.000(100.000, 0.000) \\
    Borderline-SMOTE &	Stacking-II &	100.000(100.000, 0.000) &	100.000(100.000, 0.000) &	100.000(100.000, 0.000) \\
    ADASYN &	Stacking-1 &	100.000(100.000, 0.000) &	100.000(100.000, 0.000)	 & 100.000(100.000, 0.000) \\
    ROS &	Voting-Hard &	100.000(100.000, 0.000) &	100.000(100.000, 0.000) &	100.000(100.000, 0.000) \\
    No Augmentation &	Voting-Soft &	99.834(100.000, 0.408) &	99.905(100.000, 0.232) &	99.906(100.000, 0.231) \\
    CTG &	Random Forests &	99.606(100.000, 0.871) &	99.667(100.000, 0.580) &	99.777(100.000, 0.493) \\
    SVM-SMOTE &	XGBoost &	98.760(100.000, 0.902) &	99.291(100.000, 0.519) &	99.298(100.000, 0.511) \\
    RUS &	Stacking-II &	88.291(100.000, 25.730) &	85.758(100.000, 32.522) &	93.375(100.000, 14.558) \\
    RUS &	LightGBM &	11.628(11.628, 0.000) &	0.000(0.000, 0.000) &	50.000(50.000, 0.000) \\
\hline
\end{tabular}
\end{table*}


\begin{table*}[htbp]
\footnotesize
\centering
\caption{Test results for the Yeast datasets}
\label{tab2}
\begin{tabular}{lllll}
\hline
Data Augmentation & Ensemble Model & Accuracy (Best, Std) & F1 (Best, St d) & AUC (Best, Std)\\
\hline
\multicolumn{5}{c}{Yeast-2vs8}\\
\hline
    CTG &	Stacking-1 &	96.205(96.907, 0.889) &	97.934(98.396, 0.469) &	66.886(71.952, 8.091) \\	
    Border-SMOTE &	LightGBM	& 95.997(97.423, 0.585) &	97.901(98.660, 0.313) &	76.936(81.982, 1.001)	\\
    SVM-SMOTE &	Voting-Soft &	95.832(97.423, 1.159) &	97.818(98.660, 0.618) &	73.955(80.090, 4.350) \\	
    Border-SMOTE &	Stacking-I &	95.668(97.423, 1.141) &	97.729(98.660, 0.607) &	74.415(81.982, 4.589)	\\
    ROS &	LightGBM&	95.533(96.907, 0.391) &	67.314(93.182, 25.038) &	71.936(77.237, 2.009)	\\
    Border-SMOTE &	Gradient Boost  &	95.322(97.423, 1.352) &	97.548(98.660, 0.717) &	72.398(81.982, 5.910) \\	
    No Augmentation &	Gradient Boosting   &	95.086(96.907, 1.104) &	97.455(98.396, 0.568) &	60.426(71.952, 8.070)	 \\
    SMOTE &	XGBoost & 94.536(97.423, 2.025) &	97.112(98.652, 1.091) &	74.350(82.793, 5.786)	\\
    SMOTE-ENN &	XGBoost &	89.336(95.876, 2.955) &	94.184(97.838, 1.789) &	72.915(78.468, 2.820)	\\
    RUS &	Stacking-II &	57.134(87.629, 20.164) &	66.402(93.182, 26.249) &	64.892(80.120, 7.855) \\	
    RUS &	LightGBM&	4.639(4.639, 0.000) &	0.000(0.000, 0.000) &	50.000(50.000, 0.000)	\\	
\hline
\multicolumn{5}{c}{Yeast-1}\\
\hline
    ROS &	LightGBM &	78.137(79.125, 0.658) &	84.697(85.548, 0.504) &	73.436(74.972, 0.817) \\	
    CTG &	Gradient Boost   &	77.165(78.283, 0.935) &	84.687(85.489, 0.639) &	68.850(70.389, 1.346) \\	
    CTG &	Voting-Soft &	76.008(78.620, 2.800) &	83.764(85.810, 2.349) &	67.897(70.389, 1.947)	\\
      No Augmentation &	Voting-Soft &	75.488(78.620, 2.852) &	83.374(85.714, 2.130) &	67.589(71.782, 2.941)	\\
    Border-SMOTE  &	AdaBoost   &	74.893(79.293, 2.634) &	81.847(85.409, 2.260) &	71.720(75.401, 1.873)	\\
    ROS	 & AdaBoost   &	74.520(79.125, 3.763) &	81.381(85.548, 3.428) &	71.725(74.972, 2.034)	\\
    ADASYN &	Gradient Boost  &	74.075(78.283, 2.116) &	81.199(84.661, 1.804) &	71.007(74.386, 1.594)	\\ 
    ADASYN &	Random Forests &	73.266(78.283, 2.891) &	80.683(84.661, 2.286) &	69.740(74.386, 3.009)	\\
    RUS &	LightGBM&	71.128(74.411, 1.404) &	77.847(80.759, 1.304) &	71.520(74.180, 1.172)	\\ 
\hline
\multicolumn{5}{c}{Yeast-1458vs7}\\
\hline
    ROS &	LightGBM &	96.175(96.763, 0.291) &	12.743(24.000, 5.252) &	64.711(69.042, 1.858) \\
    No Augmentation &	XGBoost &	95.444(96.043, 0.451) &	29.698(35.294, 5.531) &	59.816(61.350, 1.919) \\ 
    CTG &	Voting-Soft &	94.938(96.043, 1.400) &	32.045(42.105, 5.038) &	59.812(66.589, 4.724) \\
    SVM-SMOTE &	Stacking-II &	93.072(96.043, 2.111) &	37.206(52.174, 6.706) &	68.898(75.602, 3.208) \\
    SMOTE &	Voting-Hard &	92.116(96.043, 2.719) &	32.184(52.174, 6.784) &	66.559(77.823, 4.156) \\
    SMOTE &	AdaBoost   &	91.313(96.043, 3.262) &	34.165(52.174, 7.343) &	69.171(76.807, 3.076) \\
    Borderline-SMOTE &	Random Forests &	90.757(95.683, 3.050) &	29.555(50.000, 8.864) &	66.380(76.618, 5.568) \\
    ADASYN &	XGBoost &	89.456(94.604, 3.548) &	29.798(46.667, 8.064) &	67.851(75.036, 2.962) \\ 
    RUS	 & Stacking-II &	67.643(95.324, 11.614) &	12.330(24.000, 5.342) &	61.778(80.871, 6.215) \\
\hline

\multicolumn{5}{c}{Yeast-1vs7}\\
\hline
    No Augmentation &	LightGBM& 	93.478(93.478, 0.000) &	96.552(96.552, 0.000) &	63.037(63.037, 0.000) \\
    No Augmentation &	Stacking-2 &	92.745(94.565, 1.170) &	96.152(97.126, 0.650) &	61.626(77.041, 5.557) \\ 
    No Augmentation &	XGBoost &	92.391(93.478, 1.181) &	95.950(96.552, 0.656) &	62.446(63.037, 0.643)  \\
    CTG &	AdaBoost   &	91.848(92.391, 0.548) & 	96.050(97.126, 0.792) &	57.594(58.817, 1.233) \\
    SVM-SMOTE &	Random Forests &	88.249(91.848, 1.782) &	93.495(95.549, 1.036) &	68.679(78.600, 3.711) \\
    Borderline-SMOTE &	Stacking-I &	86.979(91.848, 1.725) &	92.723(95.549, 1.020) &	69.264(78.008, 3.597) \\
    Borderline-SMOTE &	XGBoost &	86.552(90.761, 1.340) &	92.510(94.955, 0.771) &	66.523(74.379, 3.752) \\
    SMOTE &	XGBoost &	86.461(90.761, 1.459) &	92.430(94.955, 0.844) &	68.128(78.304, 4.330) \\
    SMOTE-ENN &	Stacking-II &	84.295(91.304, 3.006) &	90.918(95.238, 1.933) &	75.676(85.937, 4.758) \\
    RUS &	Stacking-II &	62.337(82.609, 19.130) &	69.722(89.937, 26.877) &	66.750(82.170, 7.619) \\
    RUS &	LightGBM &	8.152(8.152, 0.000) &	0.000(0.000, 0.000) &	50.000(50.000, 0.000) \\
\hline
\multicolumn{5}{c}{Yeast-3}\\
\hline
    ROS &	LightGBM &	95.359(95.960, 0.296) &	97.346(97.692, 0.169) &	90.754(92.720, 0.963)	\\
    ROS &	AdaBoost  &	94.705(95.960, 0.753) &	96.966(97.692, 0.437) &	89.714(92.720, 1.474)	\\
    SVM-SMOTE &	Gradient Boost   &	94.569(95.455, 0.423) &	96.878(97.386, 0.243) &	90.494(92.936, 1.373)	\\
    CTG &	Voting-Hard &	94.484(95.791, 1.013) &	96.883(97.612, 0.528) &	85.457(90.829, 6.480)	\\
    ROS &	Random Forests &	94.397(95.960, 0.973) &	96.789(97.692, 0.552) &	89.179(93.534, 3.012)	\\
    Borderline-SMOTE &	Stacking-I &	94.312(95.455, 0.692) &	96.726(97.406, 0.407) &	90.116(92.241, 1.295)	\\
    ADASYN &	Voting-Hard &	94.240(95.455, 0.726) &	96.676(97.396, 0.428) &	90.758(93.247, 1.356)	\\ 
    SMOTE-ENN &	Voting-Hard &	93.457(95.286, 1.008) &	96.221(97.292, 0.586) &	89.278(92.361, 2.468)	\\
    RUS &	Stacking-II &	90.085(94.108, 2.397) & 	94.058(96.585, 1.534) &	91.469(94.492, 1.809)	\\
    RUS &	AdaBoost  &	88.227(91.582, 1.776) &	92.887(95.050, 1.165) &	90.049(92.553, 1.283)	\\
\hline
\end{tabular}
\end{table*}

\begin{table*}[htbp]
\footnotesize
\centering
\caption{Test results for Yeast datasets.}
\label{tab3}
\begin{tabular}{lllll}
\hline
Data Augmentation & Ensemble Model & Accuracy (Best, Std) & F1 (Best, Std) & AUC (Best, Std)\\
\hline
\multicolumn{5}{c}{Yeast-5}\\
\hline
    ROS &	LightGBM &	98.580(98.653, 0.096) &	99.269(99.307, 0.050) &	82.134(82.413, 0.754)	\\
    ADASYN &	AdaBoost   &	98.432(98.653, 0.160) &	99.191(99.307, 0.082) &	83.666(87.239, 1.551)	\\
    SVM-SMOTE &	AdaBoost  & 98.406(98.822, 0.185) &	99.178(99.393, 0.095) &	82.607(84.913, 1.807)	\\
    ROS &	Voting-Soft &	98.373(98.822, 0.206) &	99.162(99.392, 0.106) &	81.544(87.326, 2.738)	\\
    ADASYN &	XGBoost &	98.318(98.653, 0.235) &	99.131(99.307, 0.122) &	84.438(87.239, 1.824)	\\
    No Augmentation &	Gradient Boost   &	98.273(98.316, 0.074) &	99.112(99.135, 0.038) &	77.692(79.826, 0.809)\\	
    ADASYN &	Voting-Hard &	98.241(98.653, 0.191) &	99.091(99.307, 0.100) &	84.934(87.239, 1.537)	\\
    SMOTE-ENN &	Voting-Hard &	98.072(98.822, 0.609) &	99.000(99.391, 0.319)  &	87.360(96.193, 4.726)	\\
    CTG &	XGBoost &	97.868(98.316, 0.638) &	98.908(99.133, 0.321) &	71.551(79.826, 11.769)	\\
    RUS &	Stacking-II &	93.689(98.316, 2.772) &	96.606(99.124, 1.555) &	94.490(98.868, 2.744)	\\
    RUS &	AdaBoost  &	92.141(96.801, 3.363) &	95.736(98.320, 1.915) &	93.239(98.345, 3.404)	\\
\hline
\multicolumn{5}{c}{Yeast-v6}\\
\hline
    Borderline-SMOTE	 & LightGBM &	98.490(98.653, 0.150) &	99.230(99.314, 0.077) &	76.668(76.751, 0.077)	\\
    No Augmentation &	Stacking-I &	98.185(98.653, 0.320) &	99.076(99.315, 0.165)& 	69.757(76.579, 3.658)\\	
    SVM-SMOTE &	AdaBoost   &	98.072(98.653, 0.466) &	99.015(99.314, 0.240) &	75.577(80.253, 2.014)\\	
    ROS	 & Stacking-I &	97.997(98.822, 0.415) &	98.977(99.401, 0.214) &	74.884(76.837, 2.607)	\\
    Borderline-SMOTE	& Voting-Soft &	97.949(98.653, 0.501) &	98.951(99.314, 0.259) &	75.854(80.253, 1.923)	\\
    ROS &	XGBoost &	97.866(98.485, 0.303) &	98.908(99.227, 0.157) &	76.348(76.665, 0.155)	\\
    SMOTE &	Stacking-II &	97.539(98.485, 0.708) &	98.737(99.227, 0.370) &	77.385(83.841, 2.273)	\\
    SMOTE &	Voting-Hard &	97.386(98.485, 0.707) &	98.657(99.227, 0.370) &	77.607(83.841, 2.492)	\\
    SMOTE-ENN &	Random Forests &	92.917(97.306, 1.932) &	96.273(98.618, 1.048) &	73.380(78.962, 1.694)\\	
    RUS &	Stacking-II &	87.345(94.444, 3.405) &	93.087(97.093, 2.014) &	81.085(88.581, 3.112)	\\ 
\hline
\multicolumn{5}{c}{Yeast-4}\\
\hline
      No Augmentation &	LightGBM &	97.811(97.811, 0.000) &	98.882(98.882, 0.000) &	70.998(70.998, 0.000)	\\
    CTG &	Random Forests &	97.184(98.653, 1.001) &	98.559(99.315, 0.522) &	64.616(71.429, 5.025)	\\
      No Augmentation &	Gradient Boosting   &	96.874(97.811, 0.615) &	98.396(98.882, 0.320) &	68.775(70.998, 1.826)\\	
    ADASYN &	LightGBM &	94.882(95.455, 0.341) &	97.339(97.646, 0.182) &	75.074(80.160, 2.008)	\\
    SVM-SMOTE &	Voting-Hard &	94.559(96.128, 0.798) &	97.164(97.998, 0.429) &	75.475(83.042, 3.946)	\\
    Borderline-SMOTE &	Voting-Hard &	94.041(96.128, 1.045) &	96.883(97.995, 0.567) &	75.485(83.214, 4.638)\\	
    Borderline-SMOTE	 & XGBoost &	93.788(96.128, 1.360) &	96.740(97.995, 0.739) &	78.270(83.128, 2.427)	\\ 
    SMOTE-ENN &	Stacking-II &	88.530(93.771, 3.652) &	93.762(96.729, 2.157) &	80.418(88.461, 3.688)	\\
    RUS &	LightGBM	 & 80.516(87.374, 2.730) &	88.945(93.126, 1.717) &	80.961(87.931, 3.154)	\\ 
\hline
\multicolumn{5}{c}{Yeast-1289vs7}\\
\hline
    CTG & 	Gradient Boosting   &	96.343(96.570, 0.209) &	58.333(66.667, 8.404) &	57.002(63.059, 3.128) \\
    RUS &	 LightGBM &	96.042(96.042, 0.000) &	0.000(0.000, 0.000) &	50.000(50.000, 0.000) \\ 
    ROS	 & Voting-Soft &	95.531(96.834, 0.595) &	42.537(66.667, 9.104) &	63.994(75.705, 6.013)\\
    SVM-SMOTE &	Voting-Soft &	95.204(96.834, 0.870) &	40.555(71.429, 9.491) &	67.218(75.705, 3.348)\\
    Borderline-SMOTE &	Voting-Soft& 	94.818(96.306, 0.864) &	36.773(54.545, 7.162) &	67.686(78.214, 2.957)\\
    SMOTE &	Stacking-2	 & 93.077(94.987, 1.148) &	20.564(33.333, 4.745) &	61.133(77.115, 4.815)\\
    ADASYN &	Voting-Hard &	92.321(94.459, 1.189) &	16.002(26.471, 3.657) &	59.155(76.566, 4.770)\\
    SMOTE-ENN &	LightGBM &	91.478(94.987, 1.512) & 	21.900(37.500, 5.035) &	68.078(75.879, 3.941)\\
    SMOTE-ENN &	AdaBoost   &	88.259(94.987, 4.609) &	17.308(37.500, 6.131) &	66.881(75.879, 4.231)\\
    RUS &	Stacking-II &	75.501(96.042, 9.652) &	8.182(19.355, 3.985) &	66.429(83.993, 8.269)\\
\hline

\end{tabular}
\end{table*}

\begin{table*}[htbp]
\footnotesize
\centering
\caption{Test results for the Ecoli datasets.}
\label{tab4}
\begin{tabular}{lllll}
\hline
Data Augmentation & Ensemble Model & Accuracy (Best, Std) & F1 (Best, Std) & AUC (Best, Std)\\
\hline
\multicolumn{5}{c}{Ecoli-0vs1}\\
\hline
    CTG &	Stacking-I  &	98.944(100.000, 0.974) &	99.244(100.000, 0.687) &	98.623(100.000, 1.633) \\
    ROS & 	Voting-Soft &	98.782(100.000, 1.248) &	99.123(100.000, 0.894) &	98.512(100.000, 1.619) \\
    CTG &	Random Forest &	98.636(100.000, 0.993) &	99.024(100.000, 0.696) &	98.294(100.000, 1.809) \\
    No Aug & 	Voting-Soft  &	98.577(100.000, 1.478) &	98.979(100.000, 1.053) &	98.192(100.000, 2.028) \\ 
    Borderline-SMOTE &	Gradient Boosting   &	98.210(100.000, 1.410) &	98.702(100.000, 1.023) &	98.090(100.000, 1.645) \\
    SVM-SMOTE &	XGBoost &	98.081(100.000, 1.259) &	98.609(100.000, 0.909) &	97.996(100.000, 1.566) \\ 
    RUS	 & Gradient Boosting  &	97.576(100.000, 2.088) &	98.236(100.000, 1.511) &	97.572(100.000, 2.475) \\ 
    SMOTE-ENN &	LightGBM &	92.614(100.000, 7.351) &	94.864(100.000, 4.458) &	91.335(100.000, 12.341) \\
\hline
\multicolumn{5}{c}{Ecoli-1}\\
\hline
    CTG	 & LightGBM &	91.852(91.852, 0.000) &	80.000(80.000, 0.000) &	85.238(85.238, 0.000) \\
    CTG &	AdaBoost   &	90.370(91.852, 1.494) &	76.364(80.000, 3.667) &	83.095(85.238, 2.161) \\
    ROS &	Stacking-II &	89.640(92.593, 1.553) &	76.685(83.871, 3.762) &	85.106(90.476, 2.912) \\
    No Aug &	Gradient Boosting   &	89.370(91.111, 1.676) &	75.099(80.000, 4.046) &	83.266(87.143, 2.768) \\
    Borderline-SMOTE &	AdaBoost   &	89.198(92.593, 1.806) &	76.153(83.871, 4.157) &	85.119(90.476, 2.958) \\
    SMOTE &	Gradient Boosting   &	89.043(92.593, 1.361) &	76.454(83.333, 3.017) &	85.863(90.000, 2.272) \\
    SMOTE & 	Stacking-II &	88.928(92.593, 1.320) &	76.640(83.333, 2.783) &	86.379(90.714, 2.149) \\
    SMOTE &	Voting-Soft &	88.854(92.593, 1.397) &	76.507(83.333, 2.955) &	86.315(90.714, 2.296) \\
     No Augmentation &	AdaBoost   &	88.148(89.630, 1.494) &	71.983(75.000, 3.043) &	81.071(82.619, 1.561) \\
    RUS &	Stacking-II &	87.067(91.111, 1.673) &	75.253(82.353, 2.927) &	87.547(91.905, 2.242) \\
    RUS &	AdaBoost  &	86.420(89.630, 1.734) &	73.726(79.412, 3.233) &	86.190(90.476, 2.636) \\
\hline
\multicolumn{5}{c}{Ecoli-2}\\
\hline
      No Augmentation &	LightGBM	& 97.778(97.778, 0.000)	& 91.429(91.429, 0.000) &	94.017(94.017, 0.000) \\
    ADASYN &	LightGBM &	96.346(97.037, 0.613) &	86.432(88.889, 2.200) &	92.486(93.590, 1.335) \\
    ROS &	Gradient Boosting   &	96.031(97.778, 0.905) &	85.311(91.429, 3.143) &	91.873(94.017, 1.747) \\
    CTG &	Voting-Hard &	95.827(97.778, 1.737) &	84.061(91.429, 6.039) &	89.856(94.017, 3.177) \\
      No Augmentation &	Random Forest  &	95.576(97.778, 1.649)  &	83.097(91.429, 5.882) &	89.378(94.017, 3.071) \\
    SVM-SMOTE &	Stacking-II & 	95.442(97.778, 1.779) &	83.258(91.429, 5.436) &	90.476(94.017, 2.481) \\
    ADASYN &	Gradient Boosting   &	95.204(97.037, 1.144) &	82.725(88.889, 3.774) &	91.181(93.590, 1.813) \\
    Borderline-SMOTE &	Voting-Soft &	94.949(97.778, 1.750) &	81.827(91.429, 5.446) &	90.426(94.017, 2.580) \\
    RUS &	Stacking-II &	89.469(97.037, 3.648) &	69.878(88.889, 6.988) &	89.184(95.513, 3.013) \\
    RUS &	Gradient Boosting   &	88.753(95.556, 2.952) &	68.223(85.000, 5.914) &	88.830(95.085, 2.995) \\
    SMOTE-ENN &	AdaBoost   &	87.741(94.815, 3.594) &	64.869(82.051, 7.109)	 & 85.680(92.521, 3.775) \\
\hline

\multicolumn{5}{c}{Ecoli-3}\\
\hline
    CTG &	Stacking-I &	92.027(94.074, 1.316) &	57.919(73.333, 6.914) &	74.942(87.220, 4.568) \\
    CTG &	Random Forests &	91.761(94.074, 1.437) &	58.160(73.333, 7.130) &	75.771(87.220, 4.704) \\
    ROS &	Gradient Boosting  &	90.821(94.815, 1.445) &	59.033(75.862, 5.566) &	78.773(87.633, 3.220) \\
    CTG &	LightGBM	& 90.370(90.370, 0.000) &	48.000(48.000, 0.000) &	69.362(69.362, 0.000) \\
    Borderline-SMOTE &	LightGBM &	89.802(91.852, 1.225) &	56.469(64.516, 4.897) &	78.310(82.822, 3.325) \\
    ADASYN &	Random Forests  &	88.794(95.556, 1.905) &	56.201(80.000, 6.690) &	80.081(91.204, 4.342) \\
    SVM-SMOTE &	Stacking-II &	88.558(93.333, 1.349) &	49.335(68.966, 6.076) &	73.278(88.725, 4.268) \\
    SMOTE &	Stacking-I &	88.346(92.593, 1.474) &	54.184(70.588, 5.797) &	78.749(89.551, 4.406) \\
    SMOTE-ENN &	Stacking-II &	85.812(93.333, 2.861) &	52.100(74.286, 6.902)& 	80.779(94.215, 6.413) \\
    RUS	 & Stacking-II & 	81.859(89.630, 4.109)	 & 50.567(65.000, 6.809) &	84.532(93.388, 5.554) \\
    RUS &	LightGBM	 & 78.840(89.630, 3.980) &	43.743(65.000, 8.670) &	78.616(91.057, 7.468) \\
\hline
\multicolumn{5}{c}{Ecoli-4}\\
\hline
    CTG &	LightGBM & 	100.000(100.000, 0.000)	 & 73.649(85.714, 4.721) &	100.000(100.000, 0.000) \\
    SMOTE &  Voting-Soft	 & 99.005(100.000, 0.473) &	91.215(100.000, 4.046) &	93.364(100.000, 2.544) \\
    SMOTE &	Gradient Boosting  &	98.920(100.000, 0.439) &	90.580(100.000, 3.701) &	93.374(100.000, 1.825) \\
    Borderline-SMOTE &	XGBoost &	98.757(99.259, 0.494) &	88.929(93.333, 4.566) & 	92.052(93.750, 3.001) \\
    SVM-SMOTE &	Gradient Boosting  &	98.586(99.259, 0.600) &	87.194(93.333, 5.595) &	90.659(93.750, 3.562) \\ 
    SVM-SMOTE &	LightGBM &	98.025(98.519, 0.405) &	82.464(87.500, 4.194)& 	89.190(93.356, 3.417) \\
    CTG	 & Random Forest &	97.412(100.000, 1.430) &	80.128(100.000, 12.264) &	83.105(100.000, 7.856) \\
    ROS &	LightGBM  &	97.160(97.778, 0.281) &	49.790(100.000, 22.804) &	81.898(87.106, 2.369) \\
      No Augmentation &	Voting-Hard &	96.904(99.259, 1.109)& 	68.639(93.333, 9.975) &	78.590(93.750, 6.196) \\
    RUS &	LightGBM  &	94.074(94.074, 0.000) &	0.000(0.000, 0.000) & 50.000(50.000, 0.000) \\ 
\hline

\end{tabular}
\end{table*}

\subsection{Results for binary classification}

We first present results for the respective datasets with class imbalance for binary classification problems with different versions of the datasets, as outlined in Table \ref{binaryDatasets}.  
 We present  the results of Glass datasets (Table \ref{tab1}), Yeast datasets (Tables \ref{tab2}, \ref{tab3}), and the   Ecoli datasets (Table \ref{tab4}). In these results, we report the mean (best and standard deviation)  from 30 independent experimental runs for classification accuracy, F1, and  AUC scores. Note that we fixed the train/test dataset for each independent experimental run, and randomly sampled  only the initial parameters (e.g. weights and biases of neural network-based models). Out of 100 possible combinations, we show the results of 11 combinations for each dataset that include the 100th percentile  (best), 90th percentile, and so on till the 0th percentile  (worst result). These are arranged in descending order, so the better results are towards the top of the table of a particular dataset according to the accuracy score. 

\begin{figure}
\centering
\subfigure[Data augmentation methods with LightGBM]{
    \includegraphics[height=6cm]{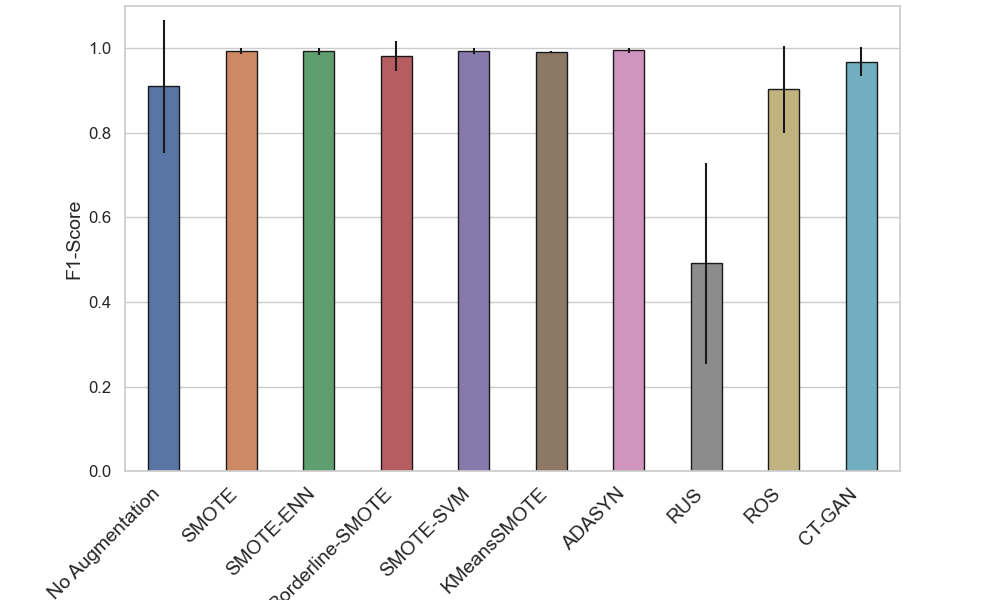}%
}
\subfigure[Ensemble learning model with no data augmentation]{
    \includegraphics[height=6cm]{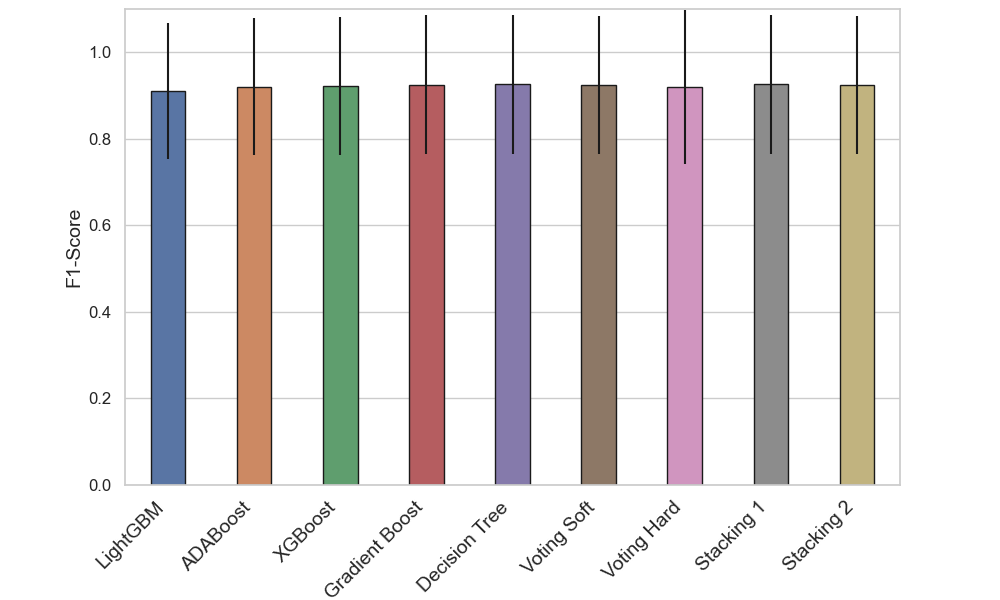}%
}
\caption{Data augmentation and ensemble learning models for Glass-derived datasets. We report the ranking based on the mean F1 score that considers the entire family of Glass-derived datasets in our study. Note that the error bars represent the standard deviation.  }
\label{fig:rankGlass}
\end{figure}

\begin{figure}
\centering
\subfigure[Data augmentation methods with LightGBM]{
    \includegraphics[height=6cm]{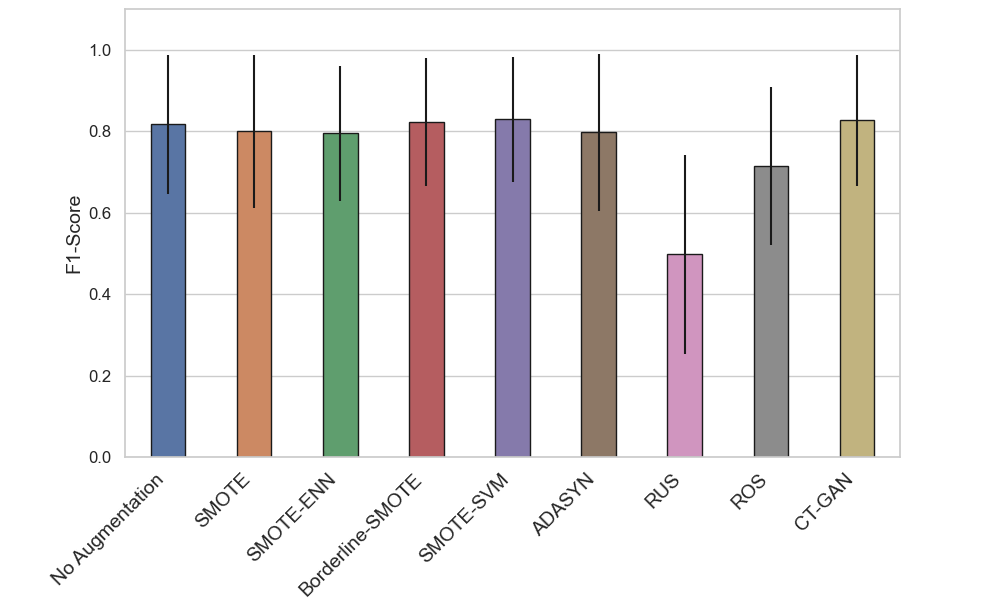}%
    
}
\subfigure[Ensemble learning model with no data augmentation]{
    \includegraphics[height=6cm]{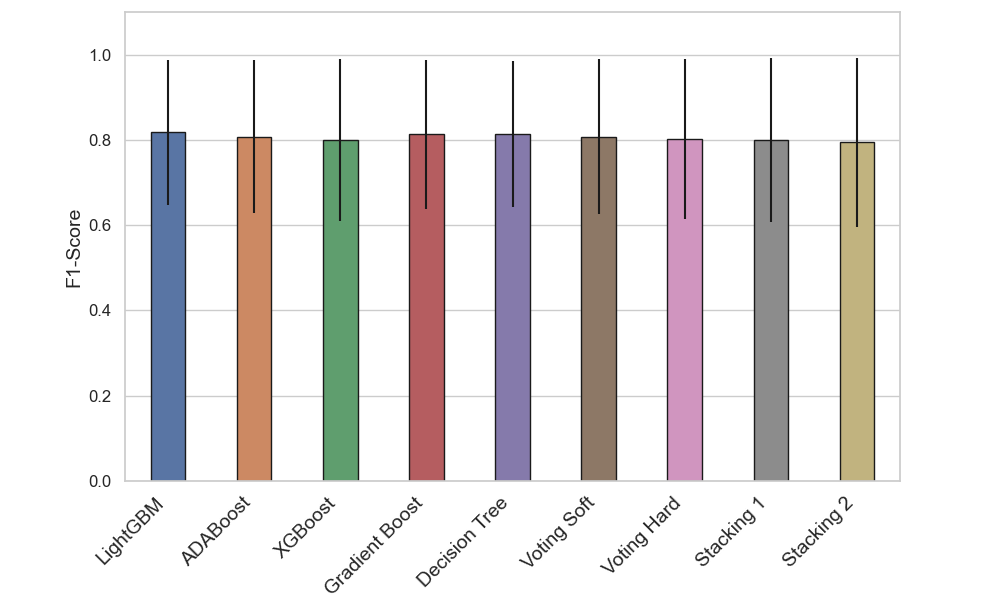}%
}
\caption{Ranking of data augmentation and ensemble learning models for Yeast-derived datasets.  We report the ranking based on the mean F1 score that considers the entire family of Yeast-derived datasets in our study.  Note that the error bars represent the standard deviation. }
\label{fig:rankYeast}
\end{figure}

\begin{figure}
\centering
\subfigure[Data augmentation methods with LightGBM]{
    \includegraphics[height=6cm]{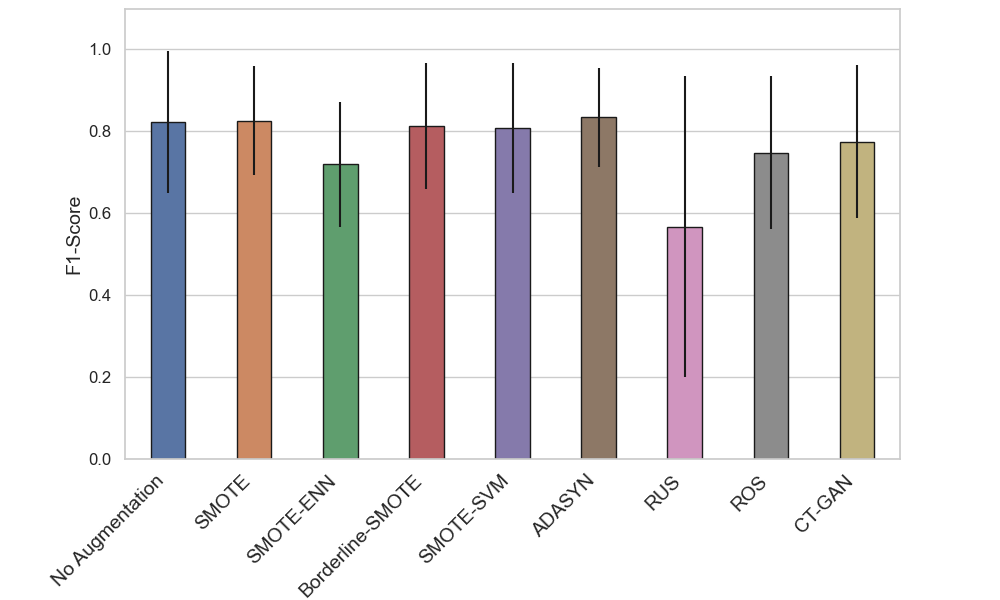}%
}
\subfigure[Ensemble learning model with no data augmentation]{
    \includegraphics[height=6cm]{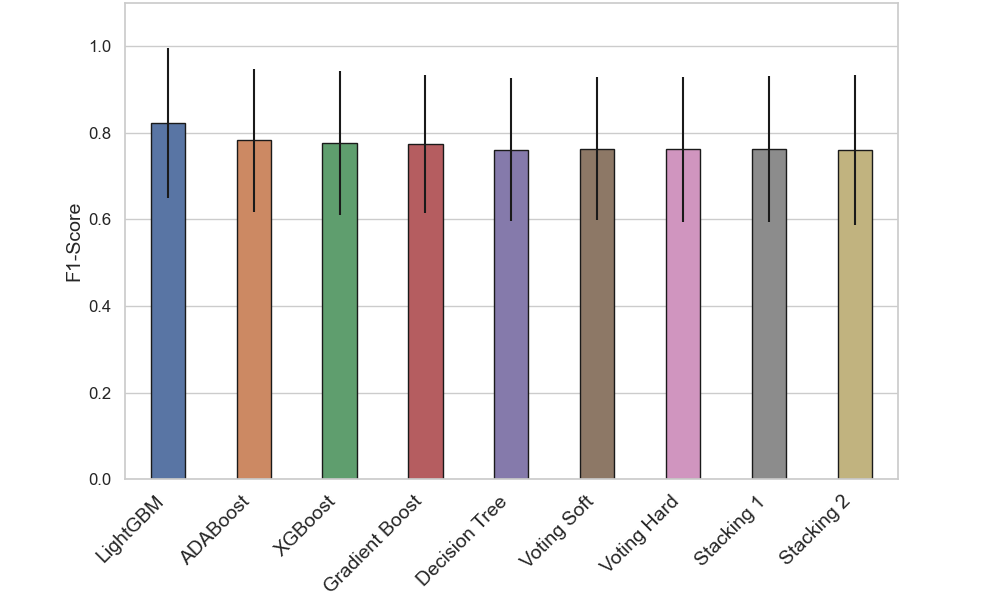}%
    
}
\caption{Data augmentation and ensemble learning models  based on the mean F1 score that considers the entire family of Ecoli-derived datasets in our study. The error bars show the standard deviation. }
\label{fig:rankEcoli}
\end{figure}

We report each combination based on the F1 score for the selected family of datasets, as shown in Figures \ref{fig:rankGlass}, \ref{fig:rankYeast}, and \ref{fig:rankEcoli}. We utilized LightGBM as the fixed ensemble model to rank various data augmentation methods.  Conversely, when ranking  the ensemble  models without any augmentation, it allowed us to systematically assess the impact of augmentation on model performance. This enabled us to objectively evaluate the performance of the methods across different datasets, in a consistent and systematic manner.

 Figure \ref{fig:rankGlass} Panel (a) presents the mean of F1 scores of Glass datasets. On the y-axis, we plot the mean F1-score, hence the higher the bar, the better the model combination. We observe that SMOTE-ENN has performed best, while  RUS performs the worst. For some datasets, \textit{no augmentation} performs better, mainly due to the smaller size of the data. Similarly, in Figure \ref{fig:rankGlass} Panel (b), we present the ranking of ensemble learning methods for the Glass datasets using the F1 score. We can interpret that LightGBM has performed best and Voting-Soft performed the worst. Figure \ref{fig:rankYeast} Panel (a) presents the   ranking of data augmentation methods on Yeast datasets. We find that SMOTE and Borderline-SMOTE perform the best, and RUS has the worst performance. Similarly, in Figure \ref{fig:rankYeast}~Panel (b), we present the  ranking of the ensemble method on Yeast datasets, where we find that the Stacking-II has the best performance and AdaBoost has the worst performance  (F1 score). Figure \ref{fig:rankEcoli} Panel (a) presents the  ranking of data augmentation methods on the Ecoli datasets where it is visible that ADASYN has performed the best, and RUS has the worst performance  (F1 score). Similarly, in Figure \ref{fig:rankEcoli} Panel (b), we present the ranking of ensemble learning methods on the Ecoli datasets, where  Voting-I has the best performance (F1 score).

\begin{figure}
\centering
\subfigure[Heatmap of F1 score of the Ecoli-3 dataset]{
    \includegraphics[height=6cm]{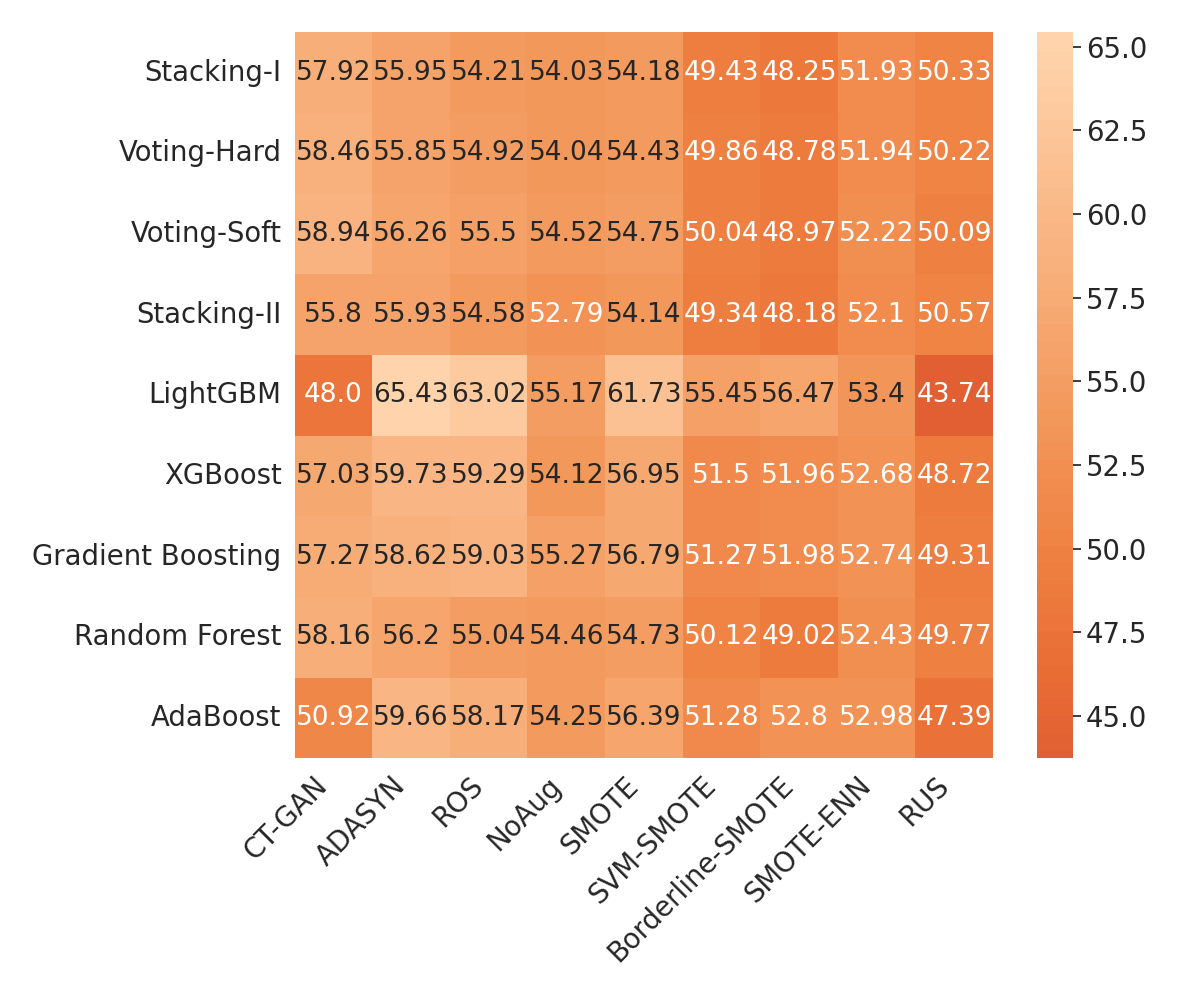}%
    \label{e3}
}
\subfigure[Heatmap of F1 score of the Ecoli-4 dataset]{
    \includegraphics[height=6cm]{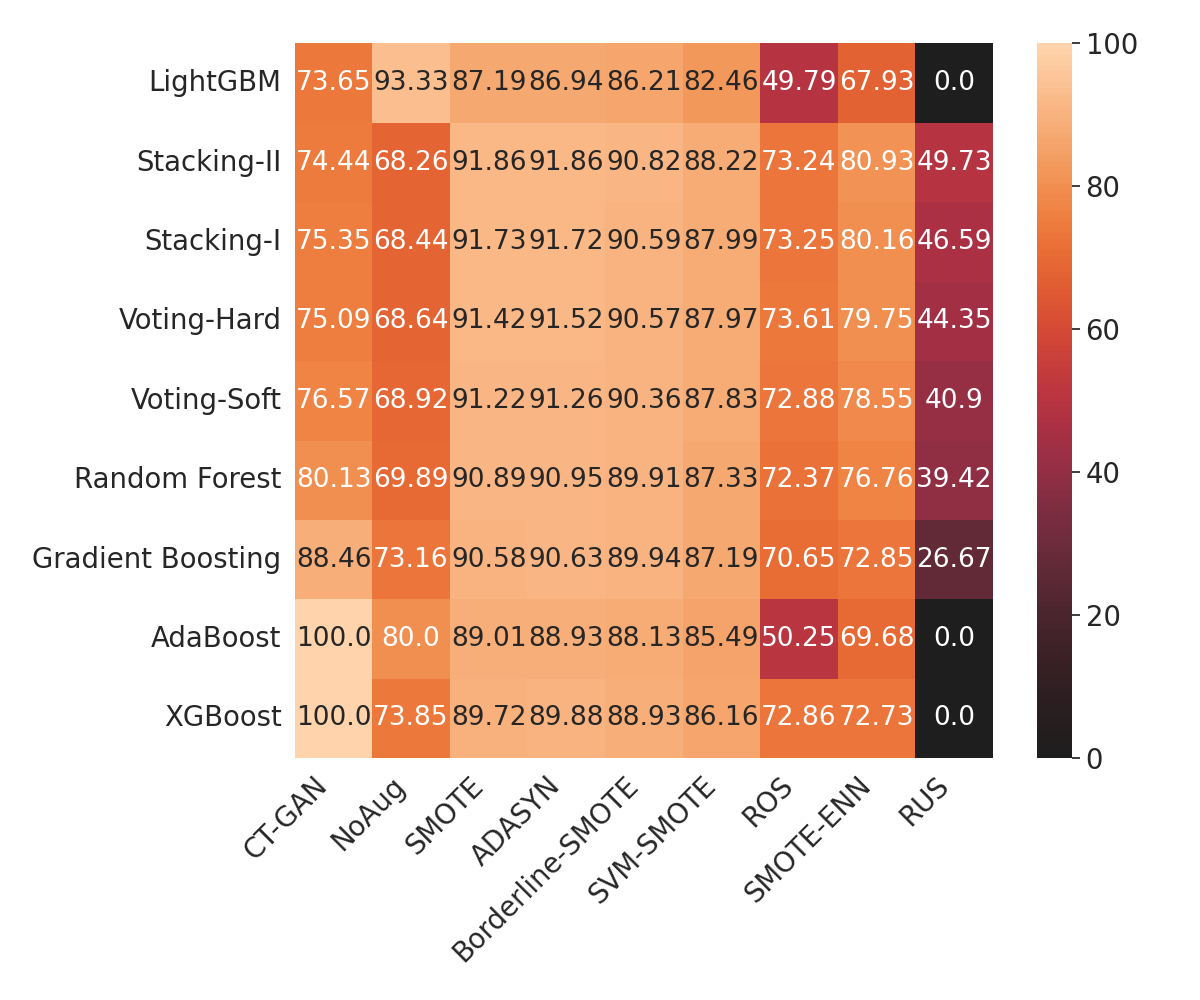}%
    \label{e4}
}

\subfigure[Heatmap of F1 score of the Ecoli-0vs1 dataset]{
    \includegraphics[height=6cm]{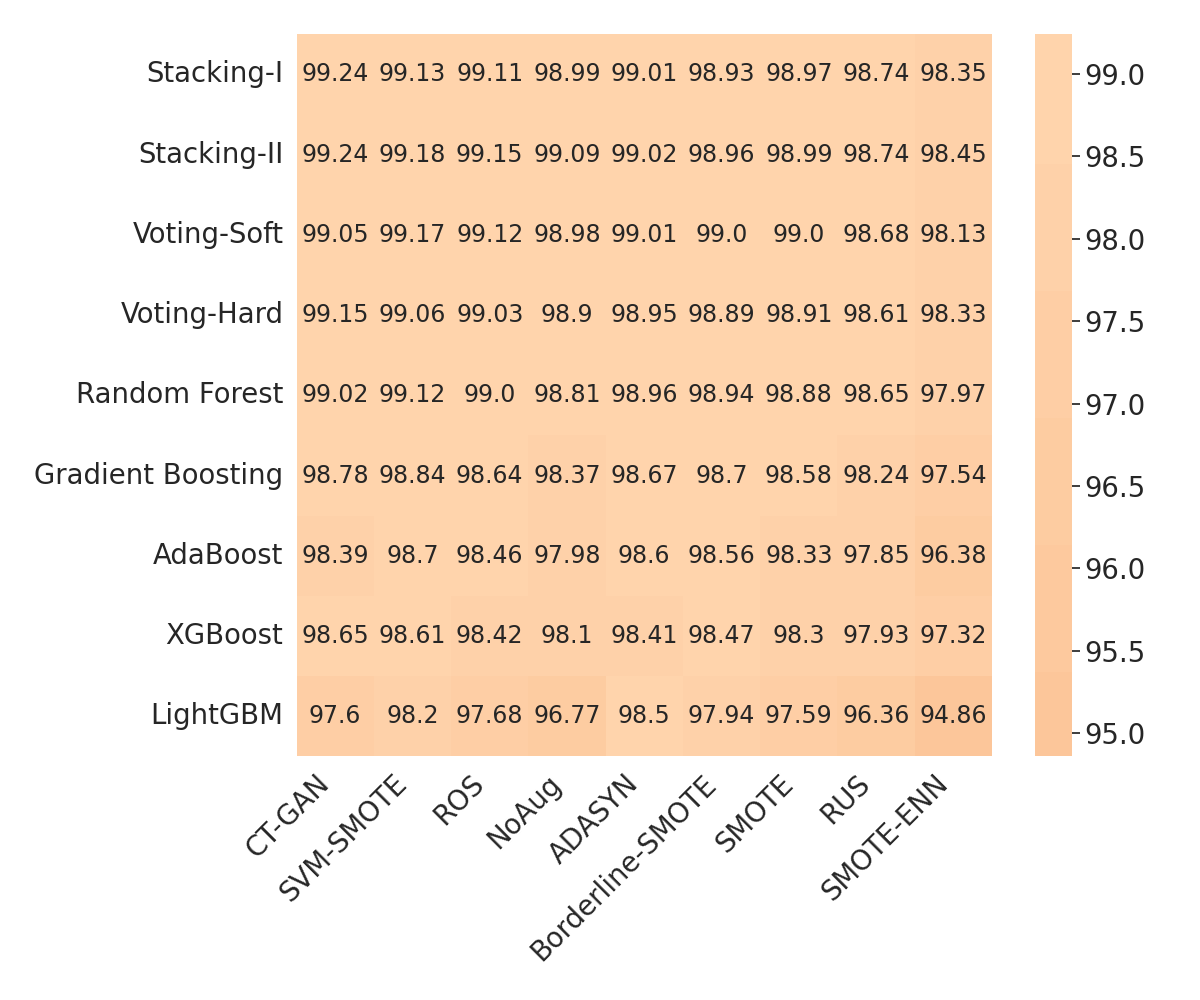}%
    \label{e0vs1}
}

 \caption{Heatmap of F1 score of the Ecoli derived datasets }
\label{EcoliHeat2}
\end{figure}

\begin{figure}
\centering
\subfigure[Heatmap of F1 score of the Glass-0 dataset]{
    \includegraphics[height=6cm]{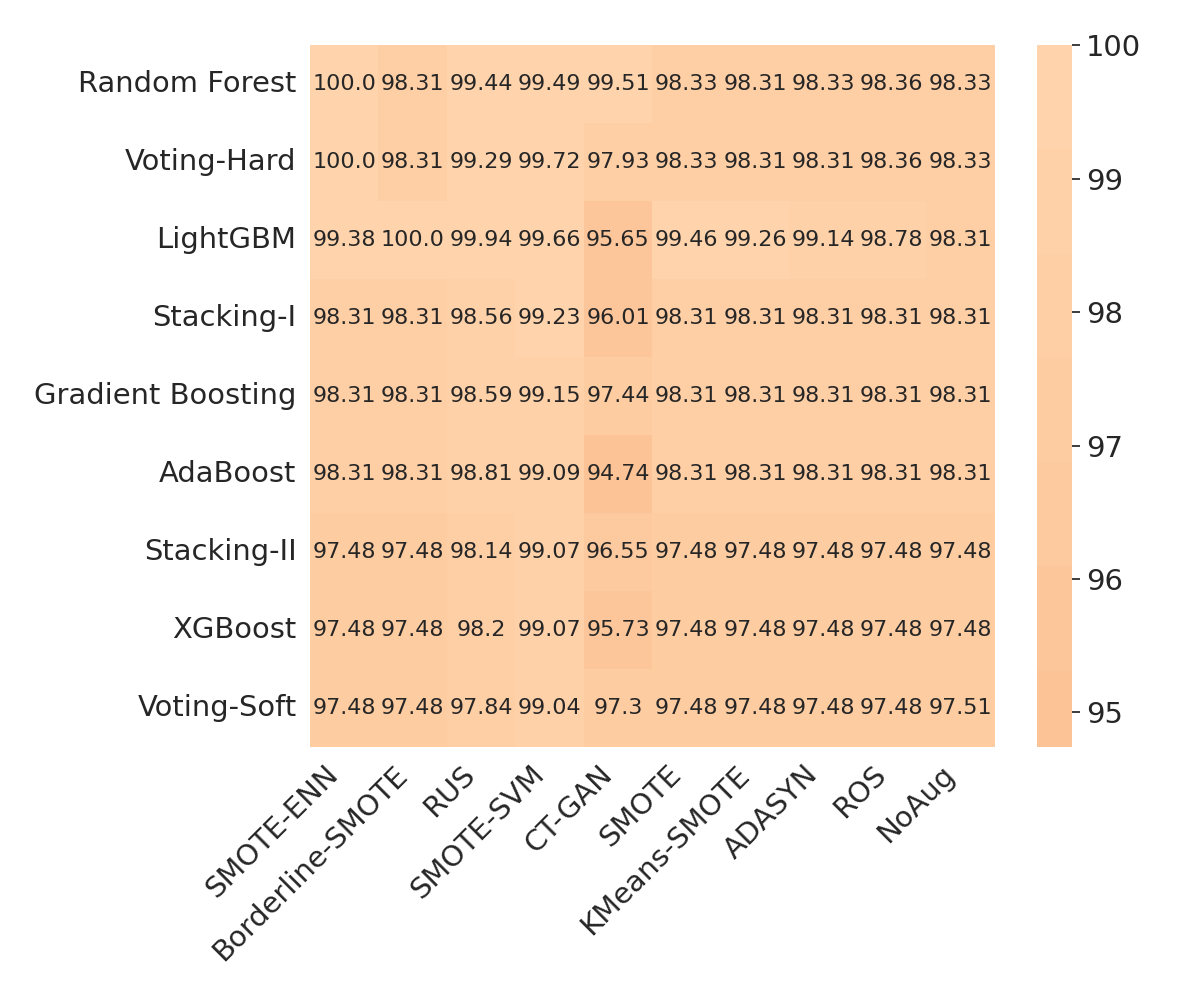}%
    \label{g0}
}
\subfigure[Heatmap of F1 score of the Glass-1 dataset]{
    \includegraphics[height=6cm]{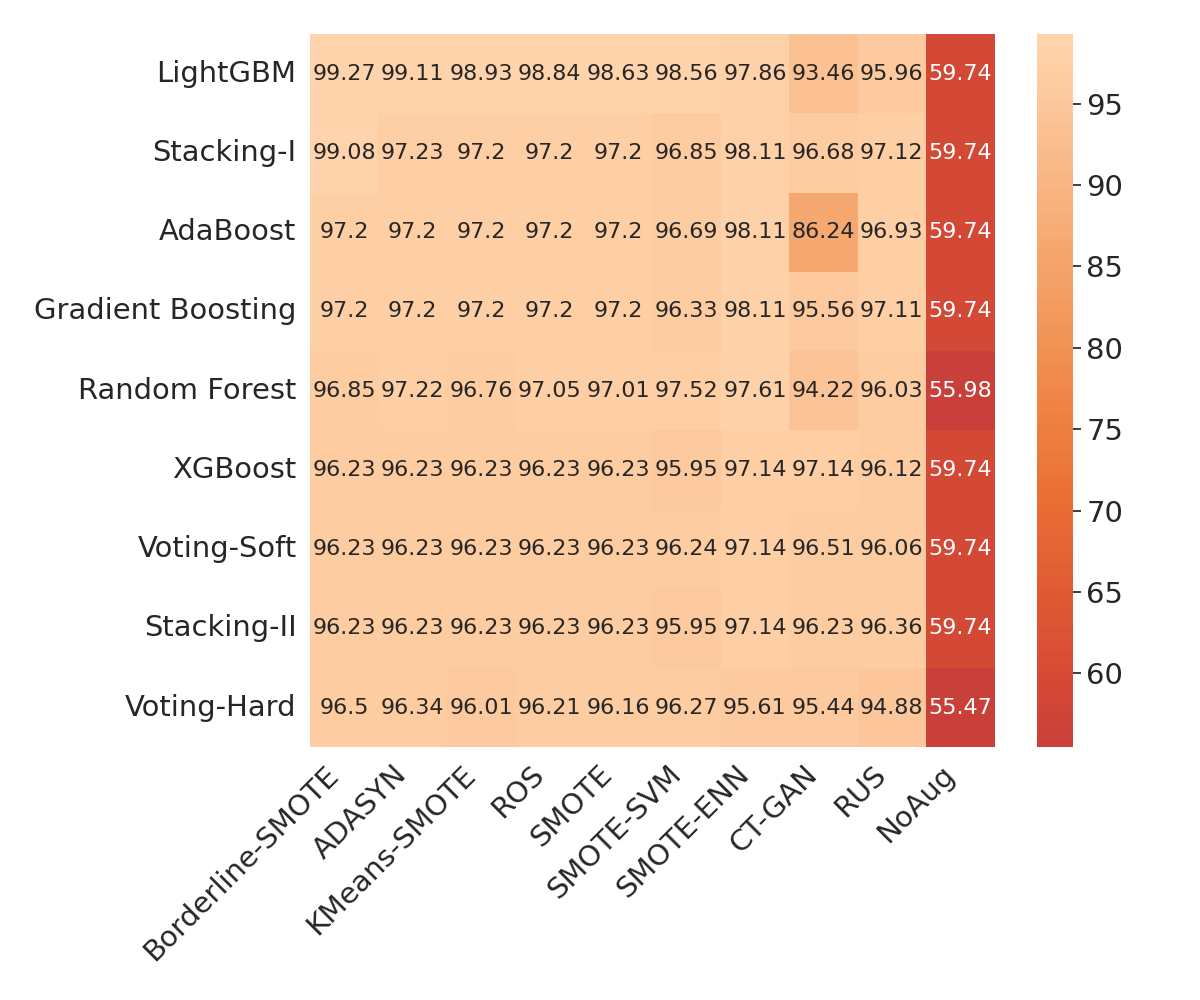}%
    \label{g1}
}
\subfigure[Heatmap of F1 score of the Glass-2 dataset]{
    \includegraphics[height=6cm]{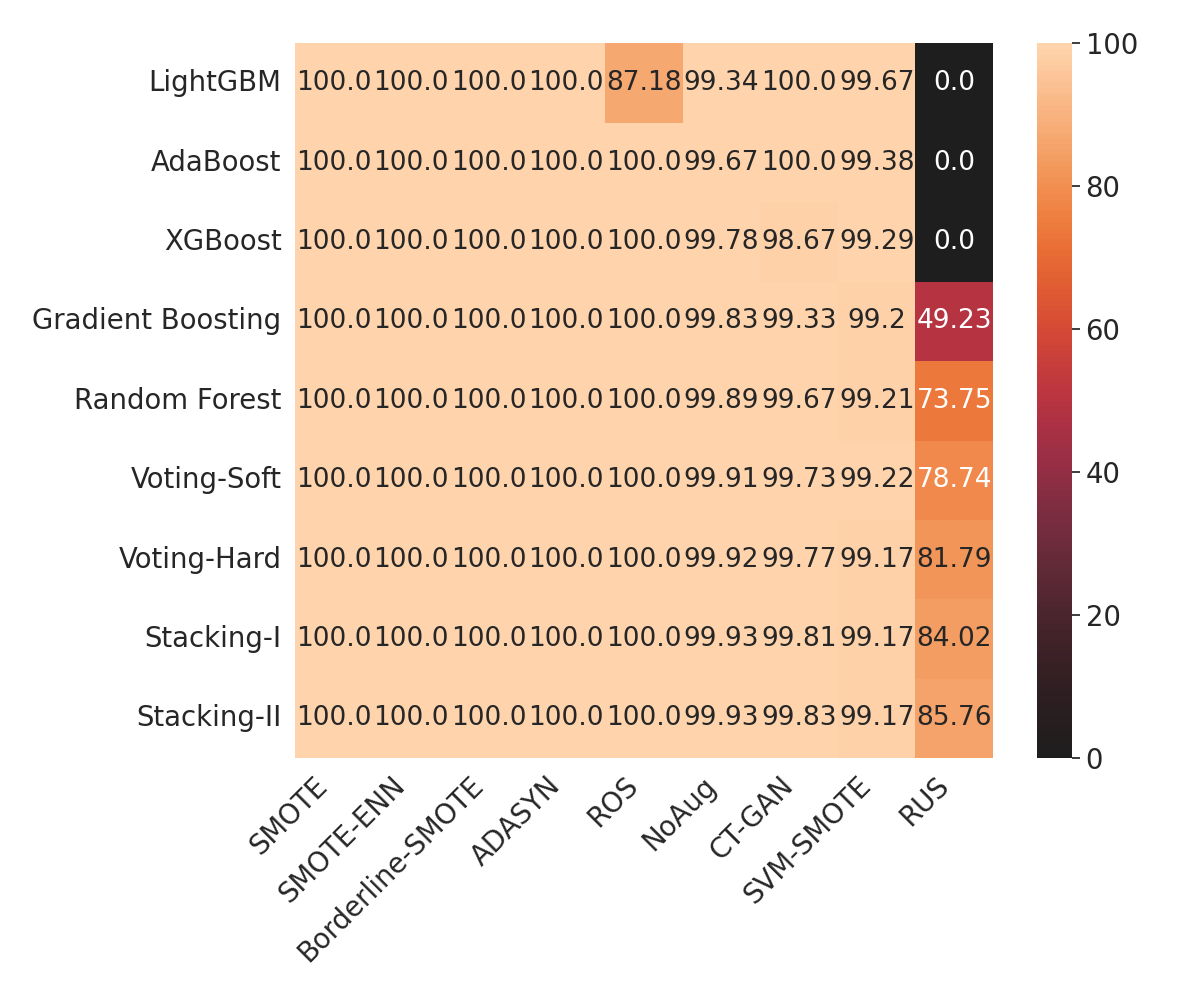}%
    \label{g2}
}

 \caption{Heatmap of F1 score of the Glass-derived datasets }
\label{GlassHeat1}
\end{figure}

\begin{figure}
\centering
\subfigure[Heatmap of F1 score of the Glass-6 dataset.]{
    \includegraphics[height=6cm]{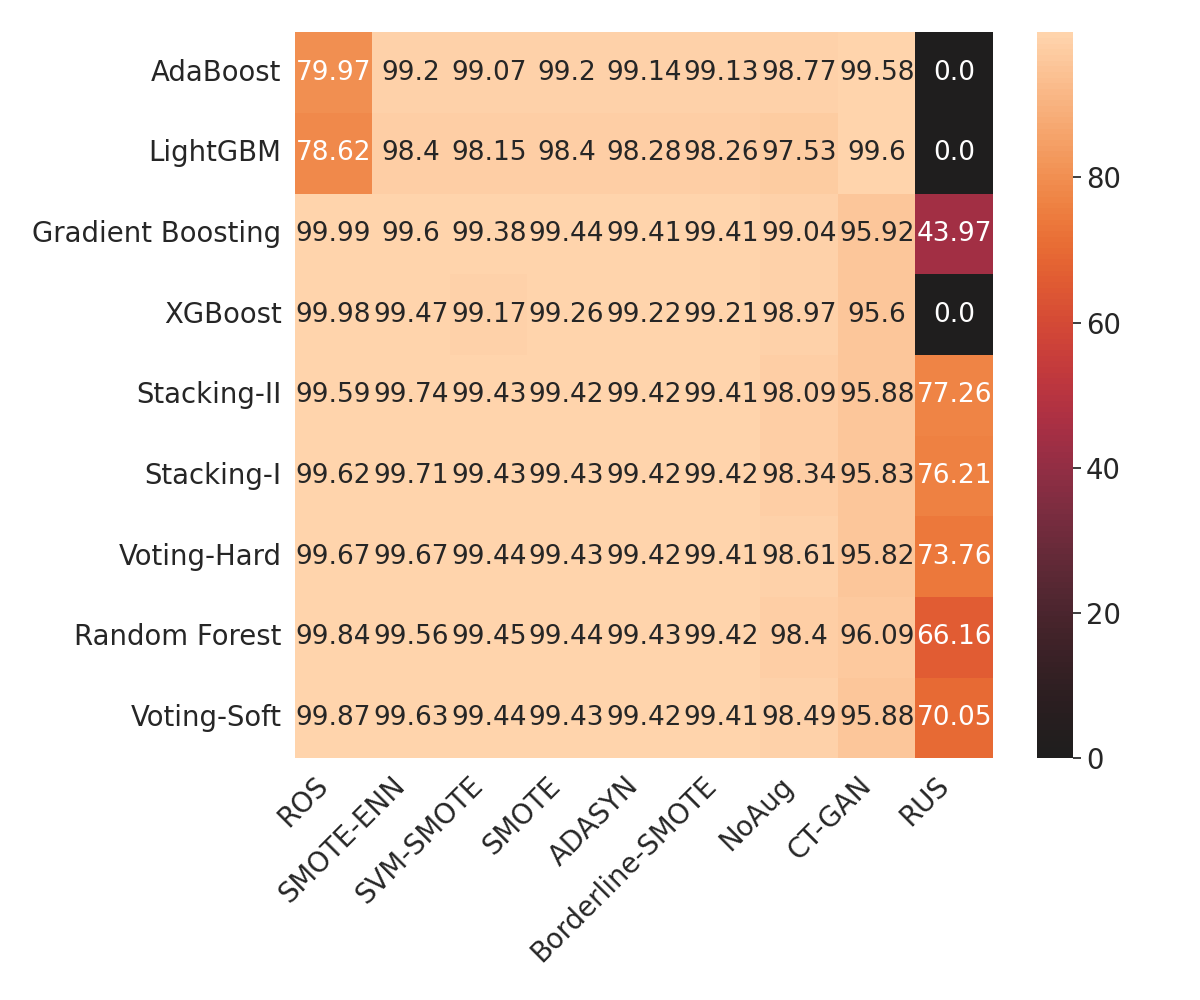}%
    \label{g6}
}
\subfigure[Heatmap of F1 score of the Glass-5vs12 dataset.]{
    \includegraphics[height=6cm]{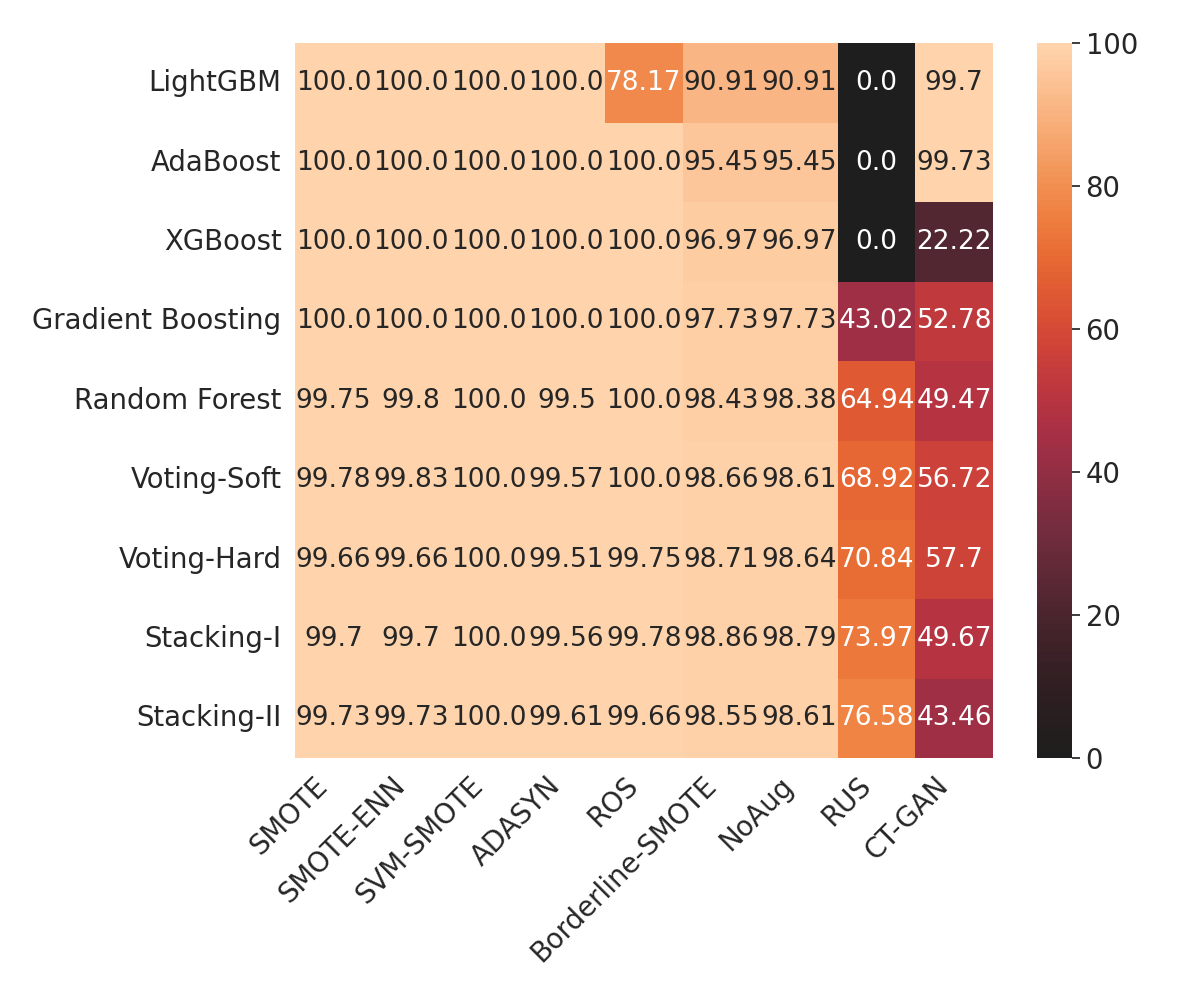}%
    \label{g5vs12}
}
\subfigure[Heatmap of F1 score of the Glass-0123vs567 dataset.]{
    \includegraphics[height=6cm]{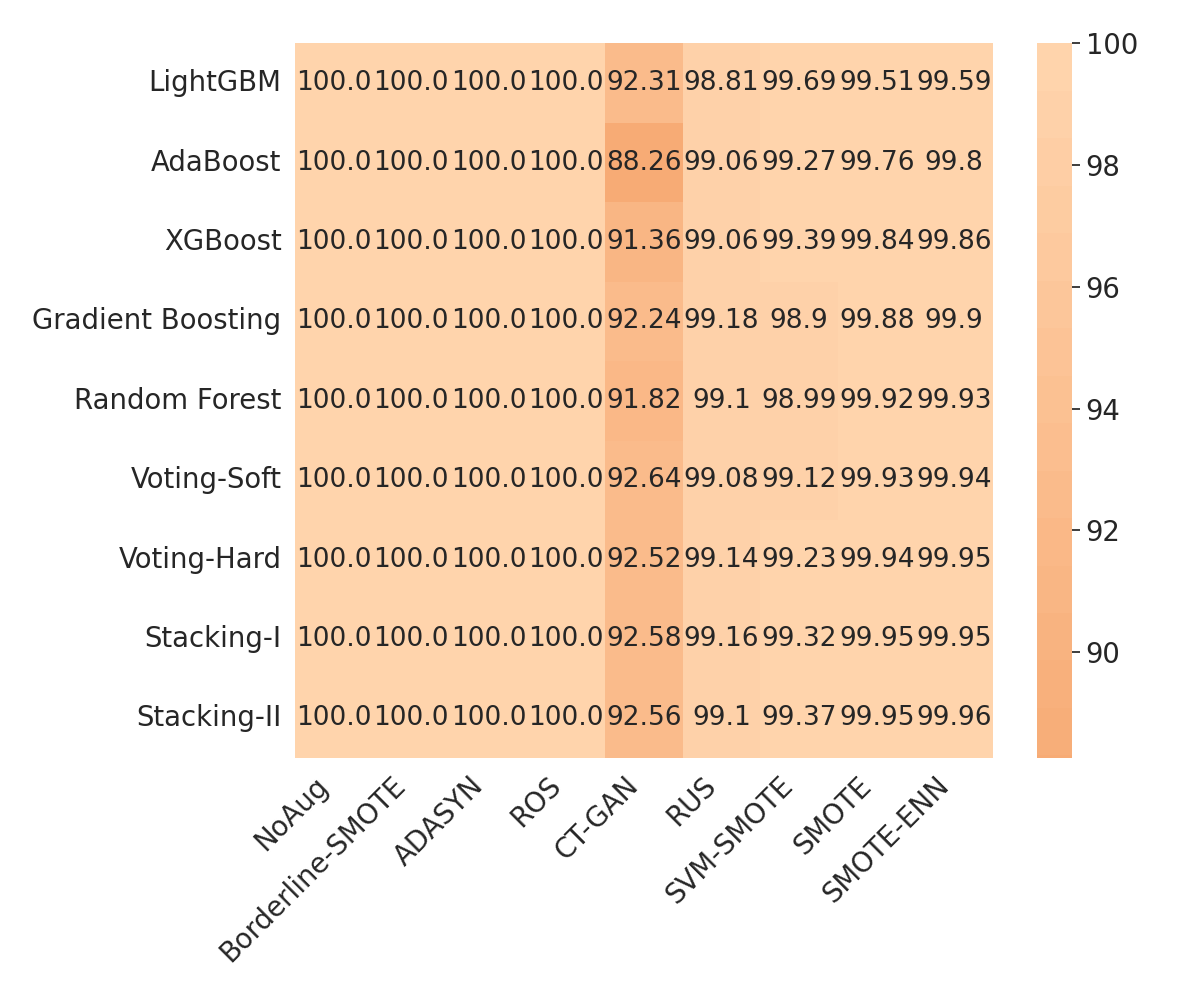}%
    \label{g0123vs567}
}
 \caption{Heatmap of F1 score of the Glass-derived datasets}
\label{GlassHeat2}
\end{figure}

\begin{figure}
\centering
\subfigure[Heatmap of F1 score of the Yeast-1 dataset]{
    \includegraphics[height=5cm]{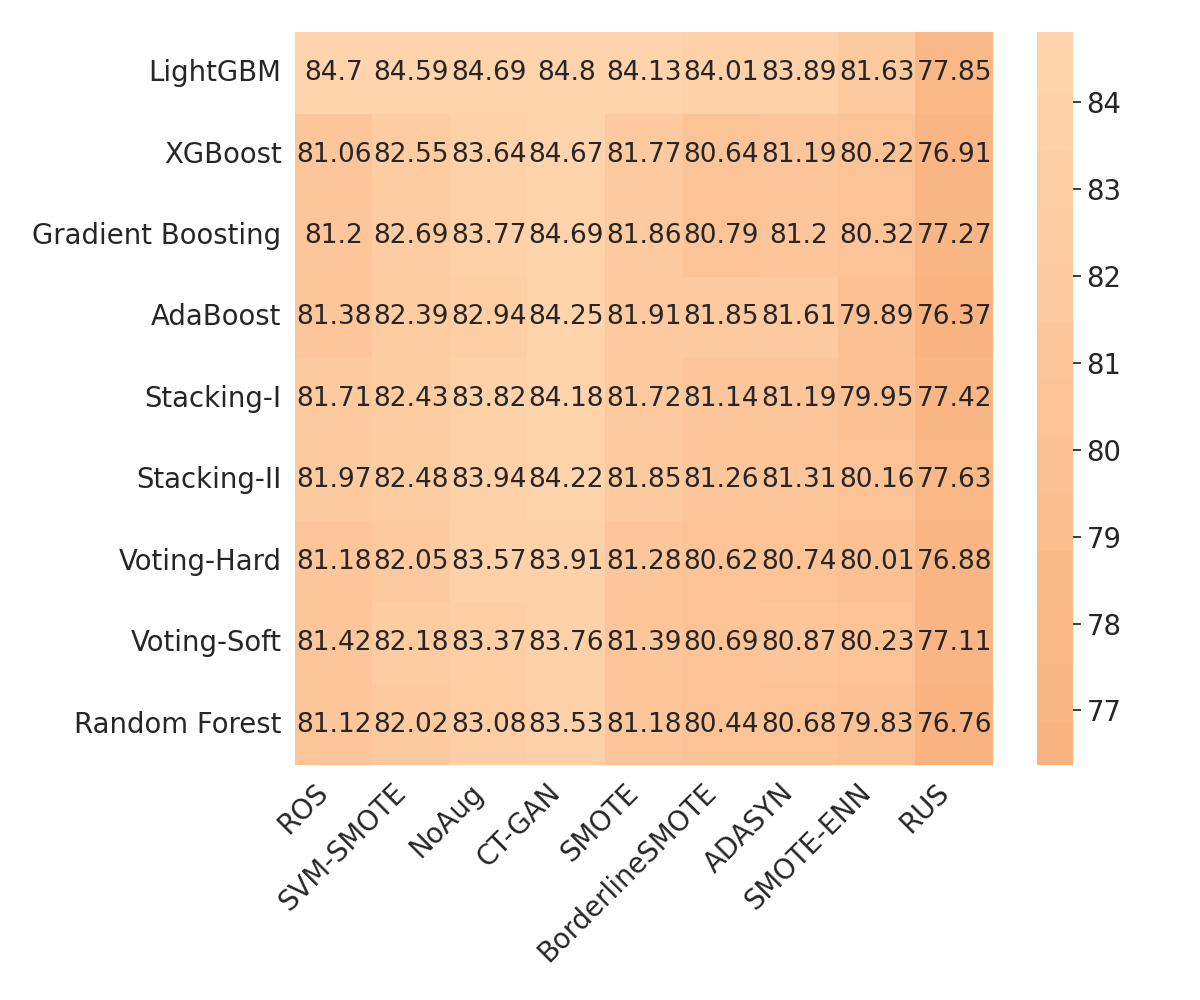}%
    \label{y1}
}
\subfigure[Heatmap of F1 score of the Yeast-1vs7 dataset]{
    \includegraphics[height=5cm]{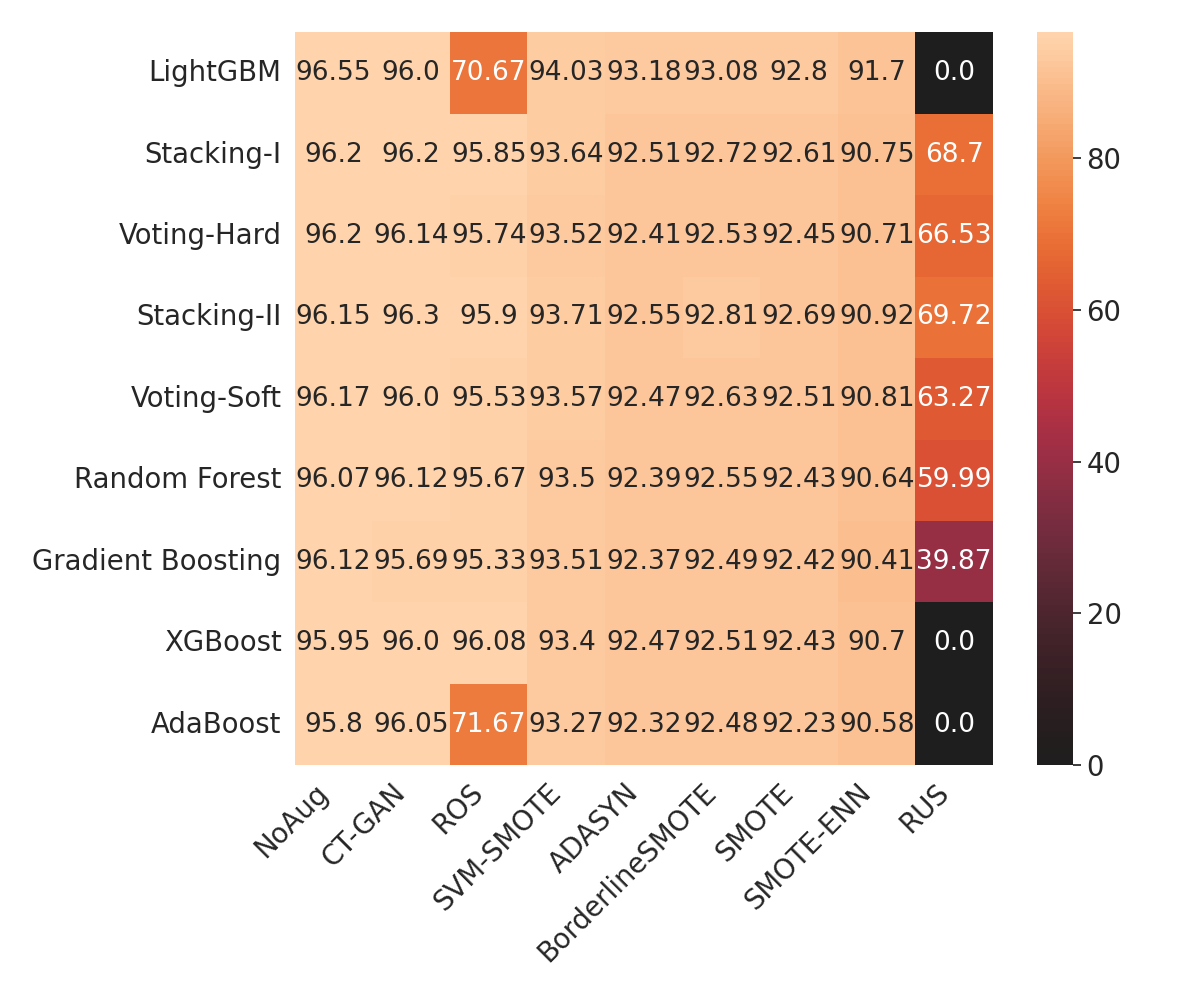}%
    \label{y1vs7}
}
\subfigure[Heatmap of F1 score of the Yeast-2vs8 dataset]{
    \includegraphics[height=5cm]{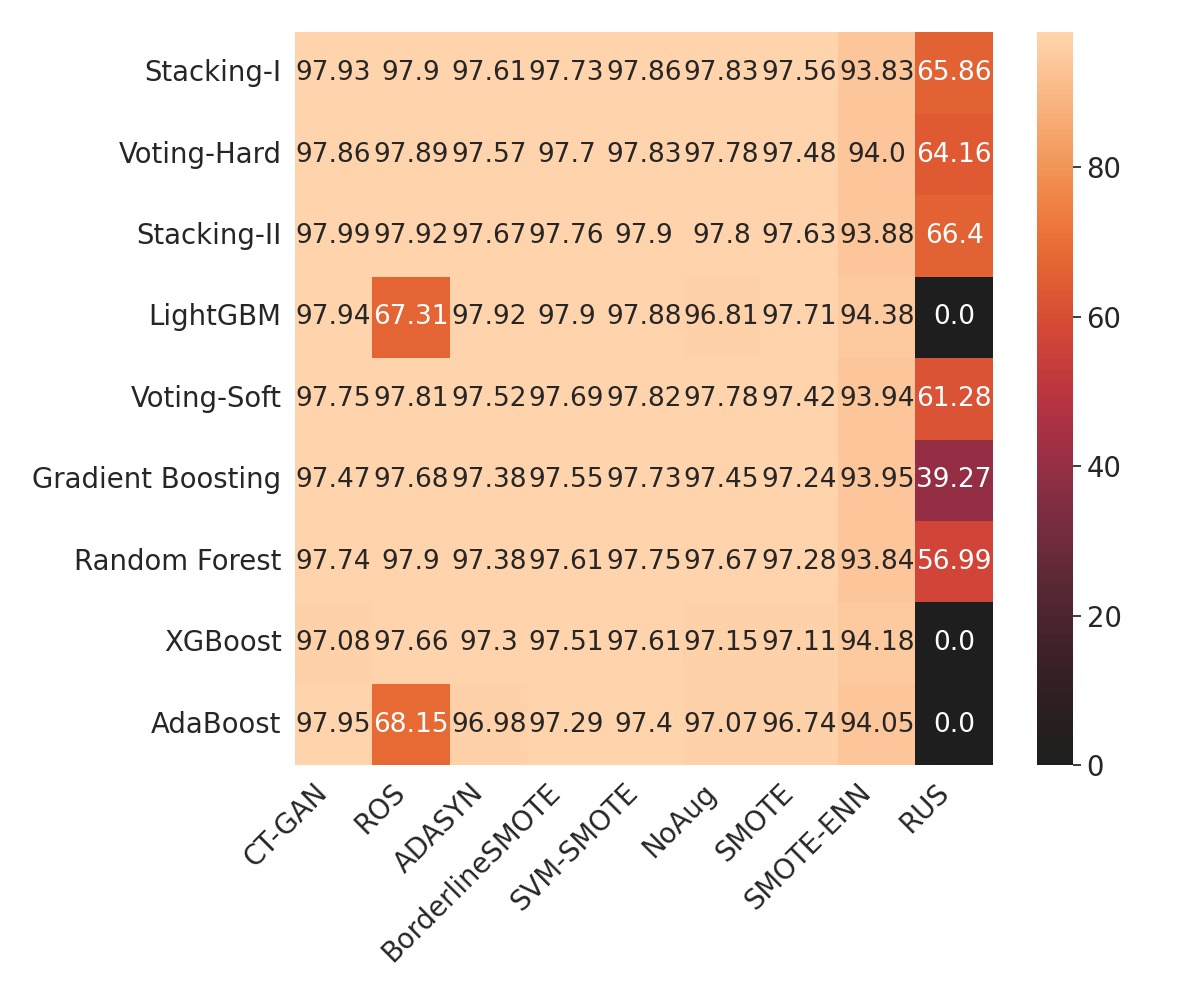}%
    \label{y2vs8}
}
\subfigure[Heatmap of F1 score of the Yeast-3 dataset]{
    \includegraphics[height=5cm]{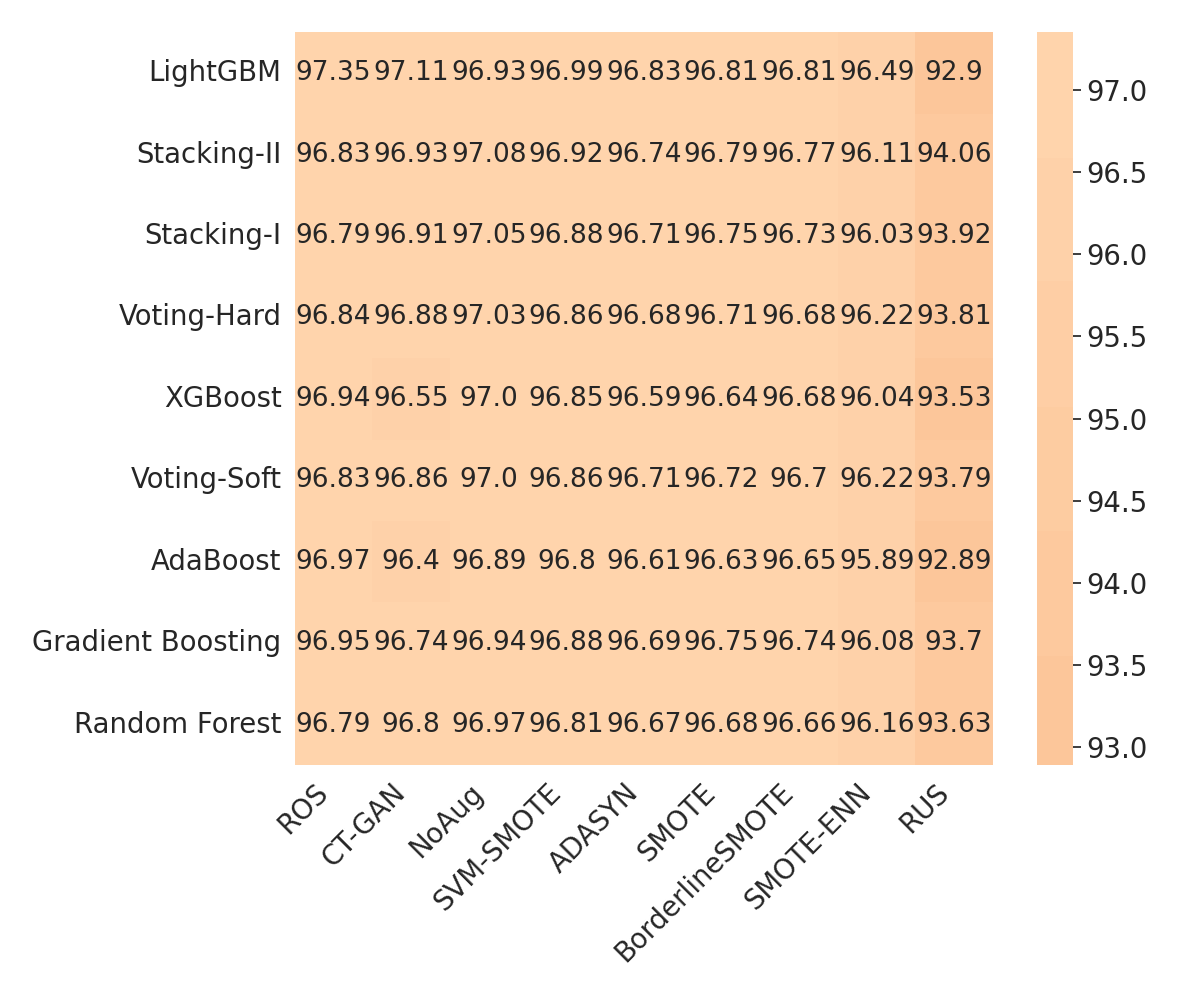}%
    \label{y3}
}
 \caption{Heatmap of F1 score of the Yeast-derived datasets }
\label{YeastHeat1}
\end{figure}

\begin{figure}
\centering
\subfigure[Heatmap of F1 score of the Yeast-4 dataset]{
    \includegraphics[height=5cm]{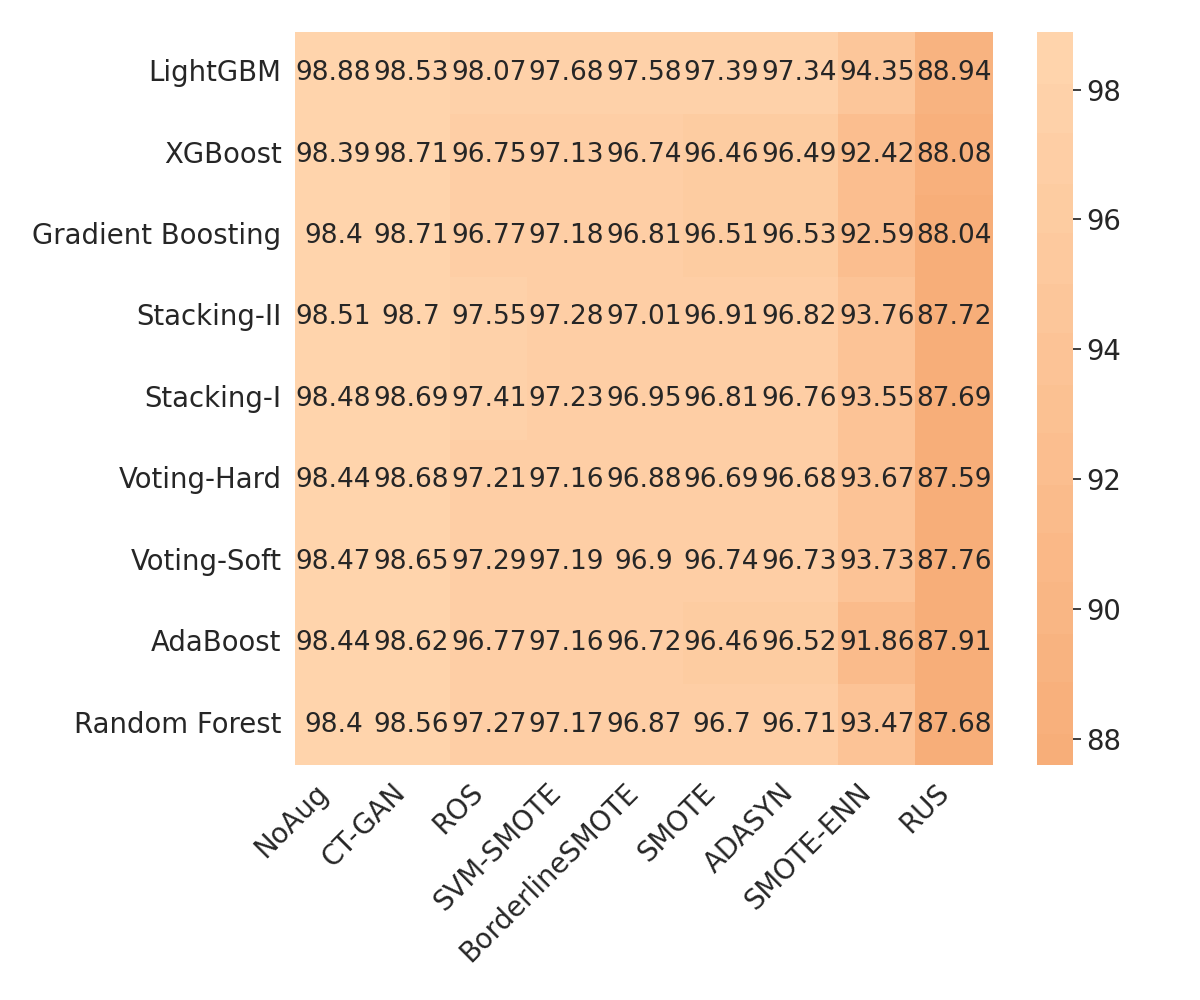}%
    \label{y4}
}
\subfigure[Heatmap of F1 score of the Yeast-5 dataset]{
    \includegraphics[height=5cm]{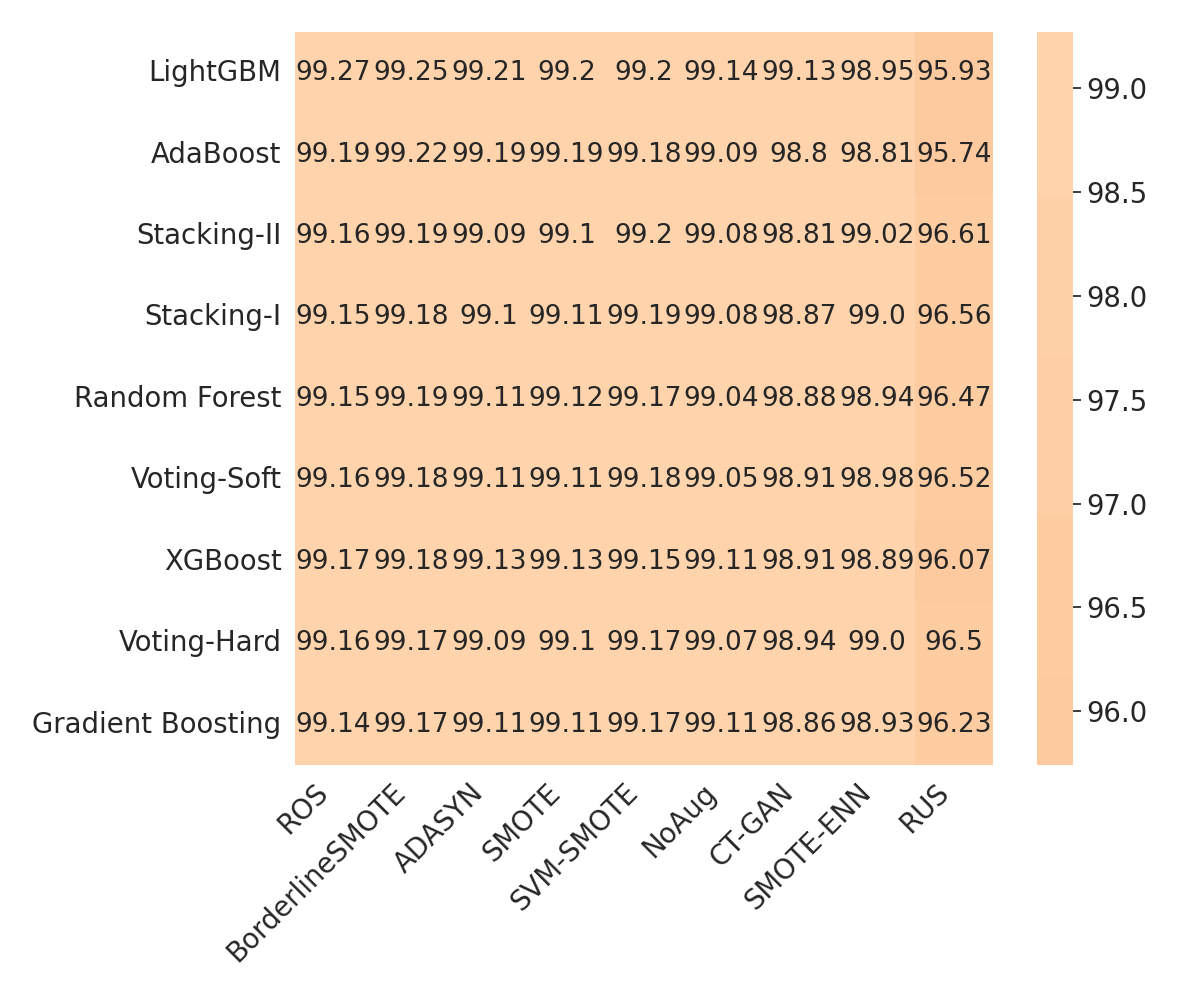}%
    \label{y5}
}
\subfigure[Heatmap of F1 score of the Yeast-6 dataset]{
    \includegraphics[height=5cm]{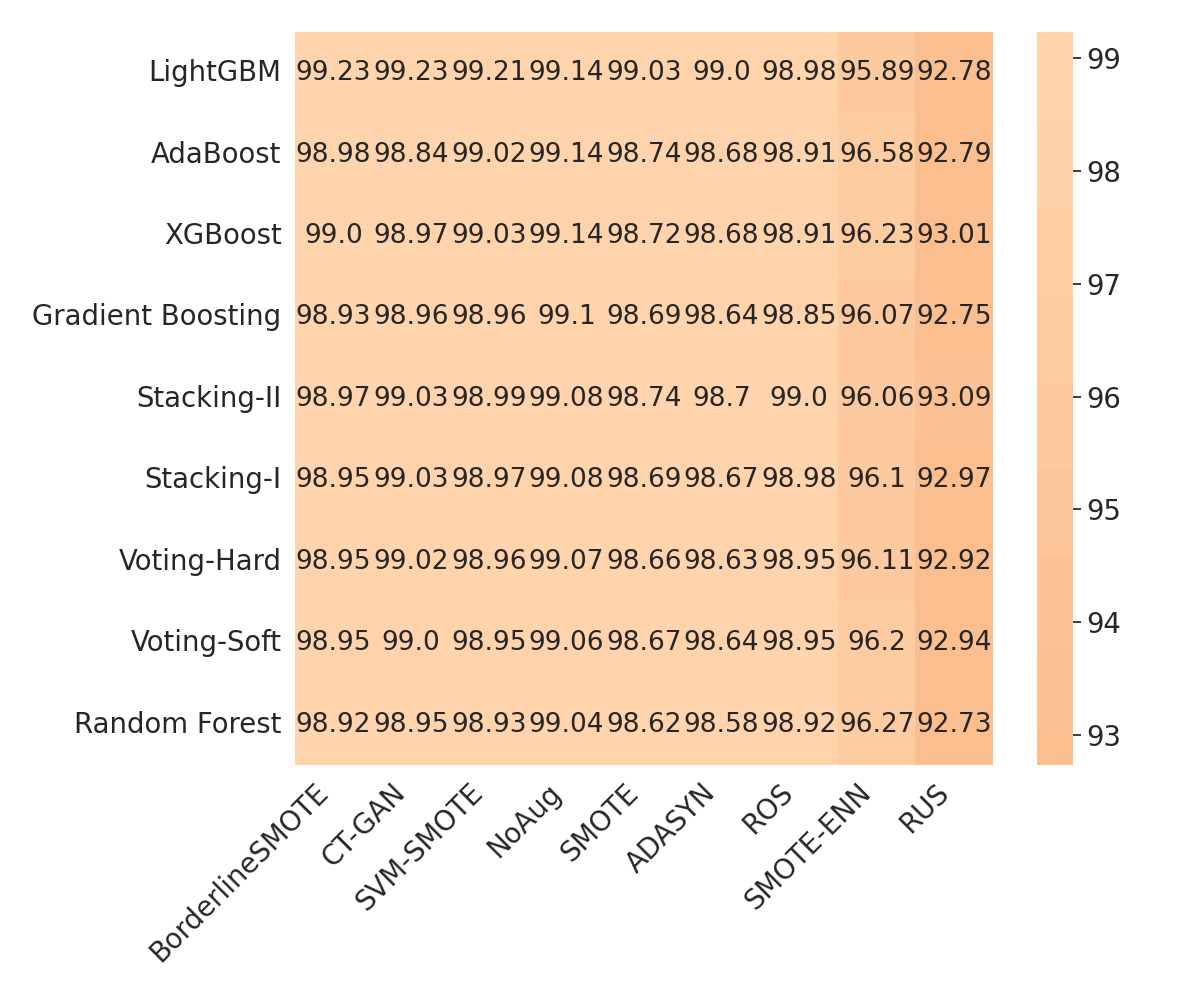}%
    \label{y6}
}
\subfigure[Heatmap of F1 score of the Yeast-1289vs7 dataset]{
    \includegraphics[height=5cm]{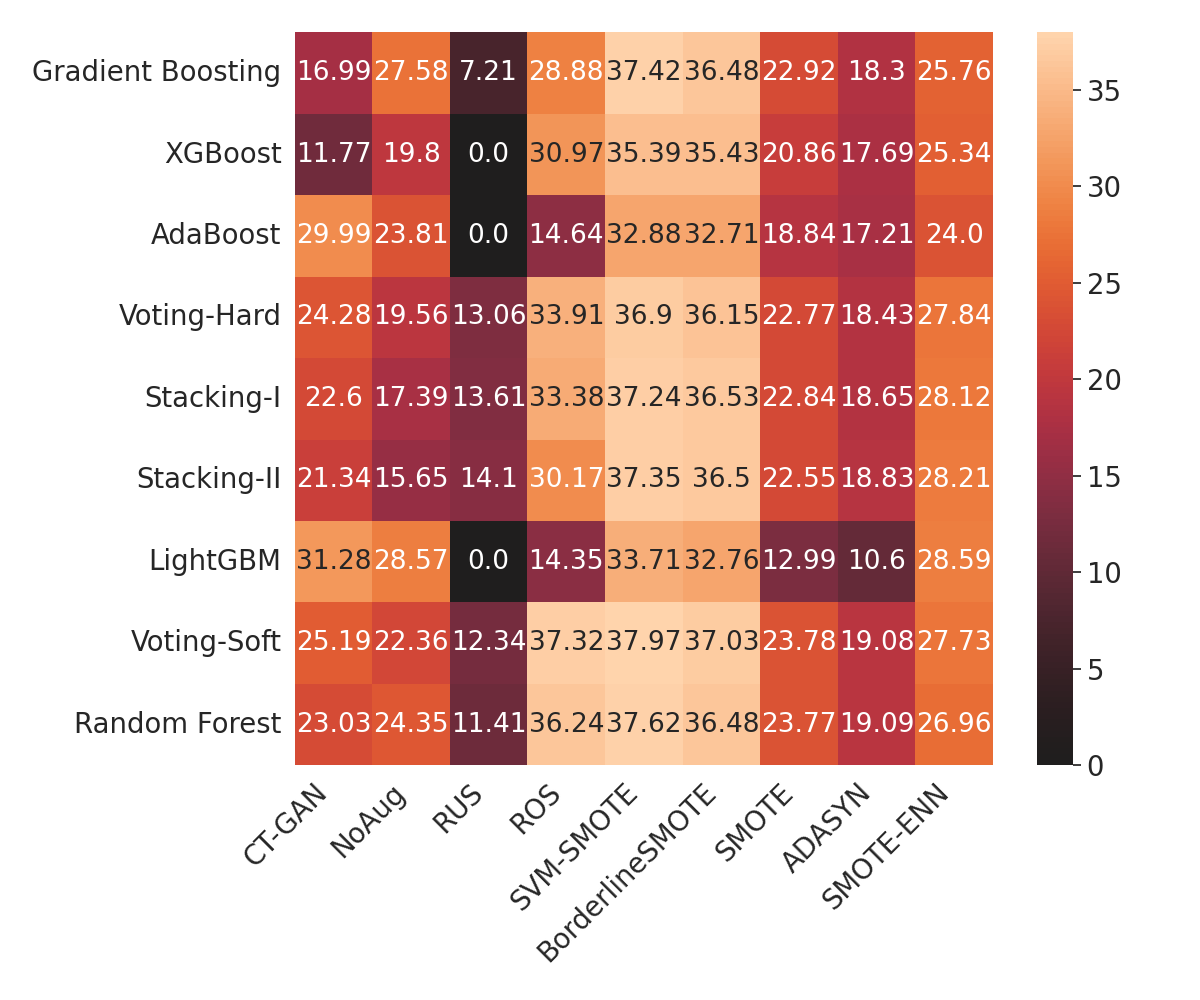}%
    \label{y1289vs7}
}
 \caption{Heatmap of F1 score of the Yeast-derived datasets }
\label{YeastHeat2}
\end{figure}

 Figures \ref{g1}, \ref{g6}, \ref{y1}, \ref{y5}, \ref{e3}, and \ref{e4} are heatmaps of AUC scores of Glass-1, Glass-6, Yeast-1, Yeast-5, Ecoli-3, and Ecoli-4 datasets, respectively.  The results show that Stacking-II   (based on Random Forests and XGBoost) outperformed Stacking-I (based on Decision Trees, Random Forests, K-Nearest Neighbour, and XGBoost). We can observe that Voting-Hard is generally better than Voting-Soft. The combination of SMOTE and LightGBM has outperformed all the others in the Glass datasets. Similarly, ROS and LightGBM in Yeast datasets. and CT-GAN and LightGBM  in Ecoli datasets have outperformed the other combinations. The performance of CT-GAN with Stacking I and II  and LightGBM  has been noteworthy in Ecoli-0, Ecoli-1, and other Ecoli datasets with low imbalance ratios that imply more imbalance.

\begin{figure}
\centering
\subfigure[Heatmap of F1 score of the Ecoli-0 dataset]{
    \includegraphics[height=6cm]{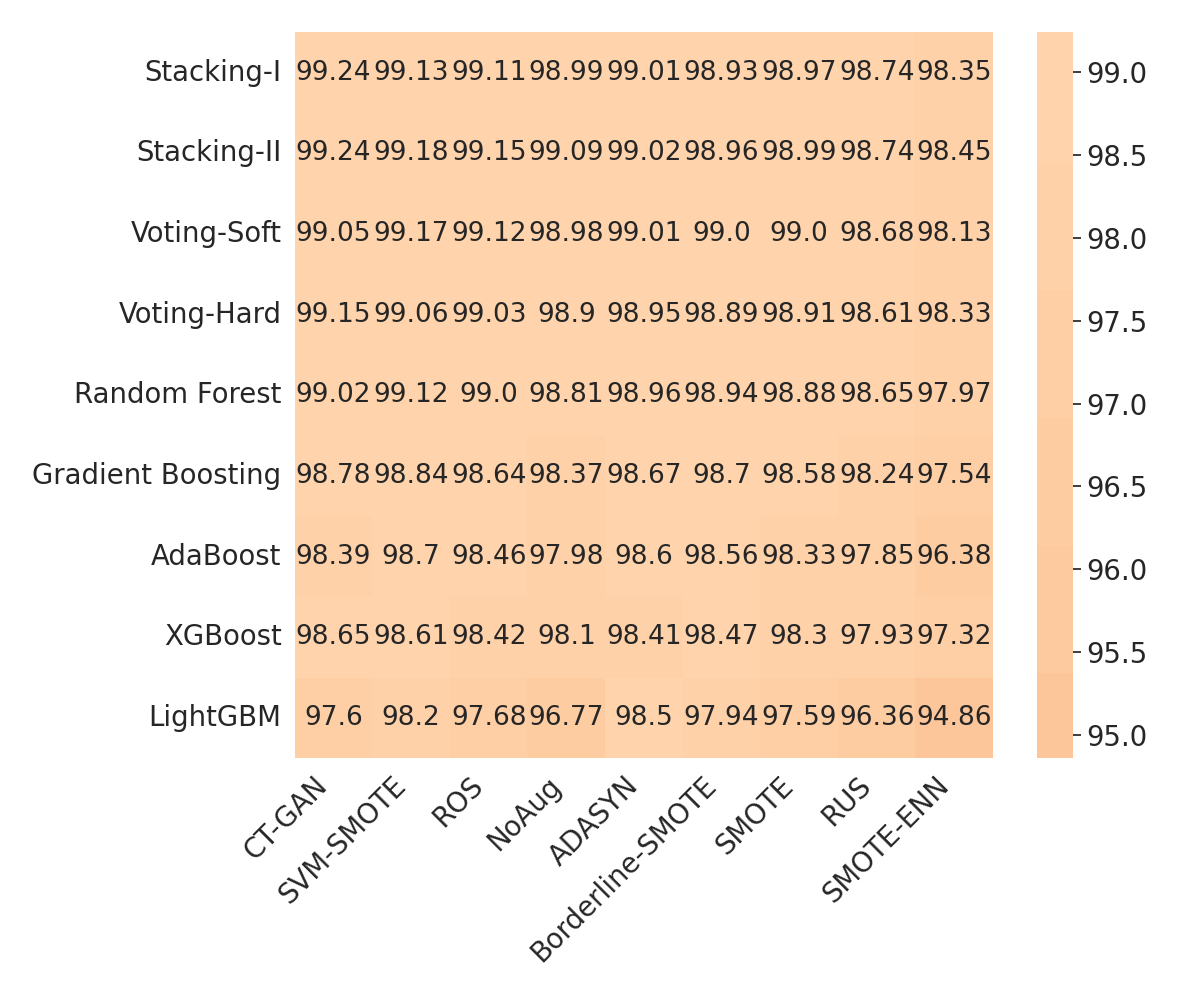}%
    \label{e0vs1}
}
\subfigure[Heatmap of F1 score of the Ecoli-1 dataset]{
    \includegraphics[height=6cm]{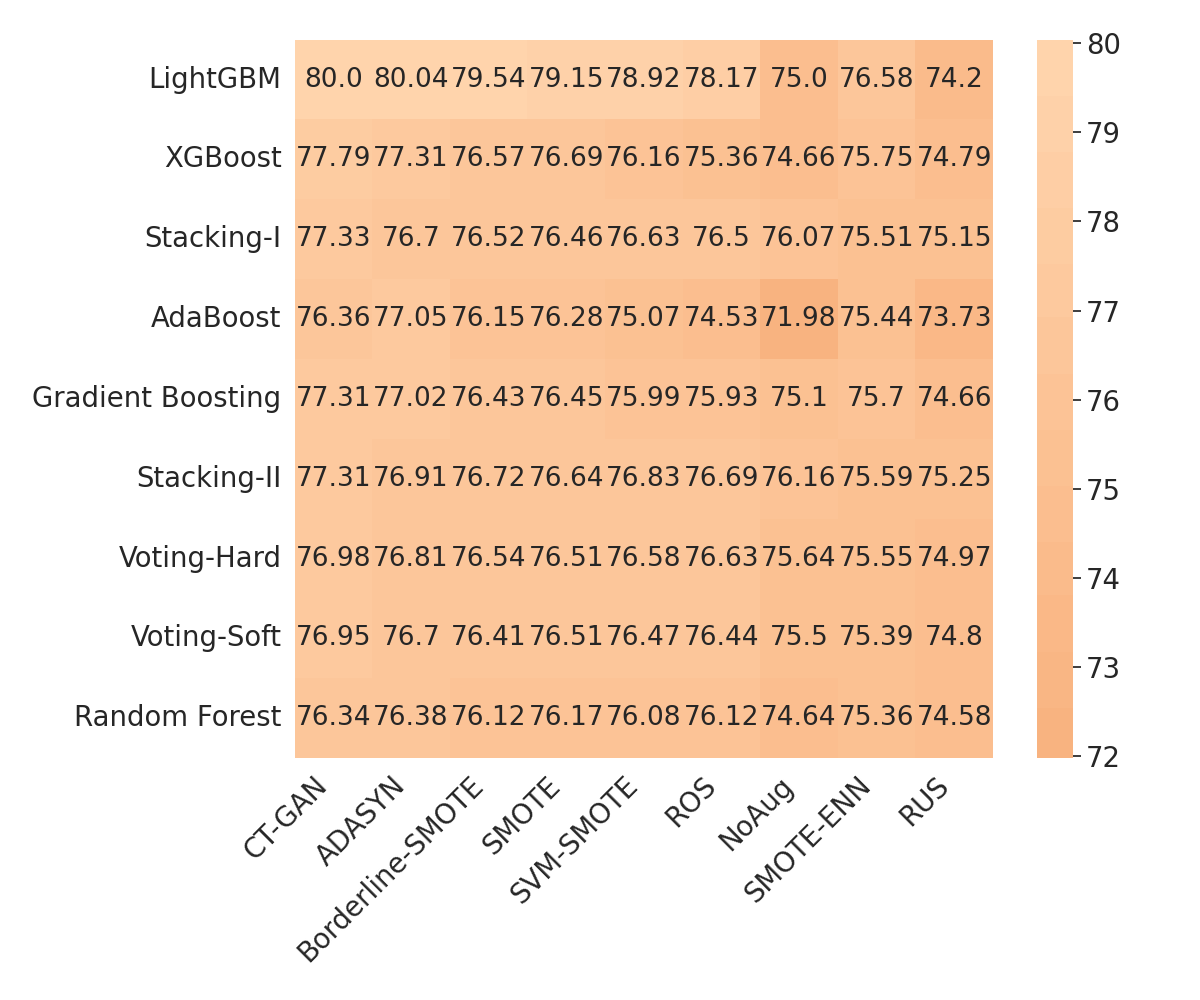}%
    \label{e1}
}
\subfigure[Heatmap of F1 score of the Ecoli-2 dataset]{
    \includegraphics[height=6cm]{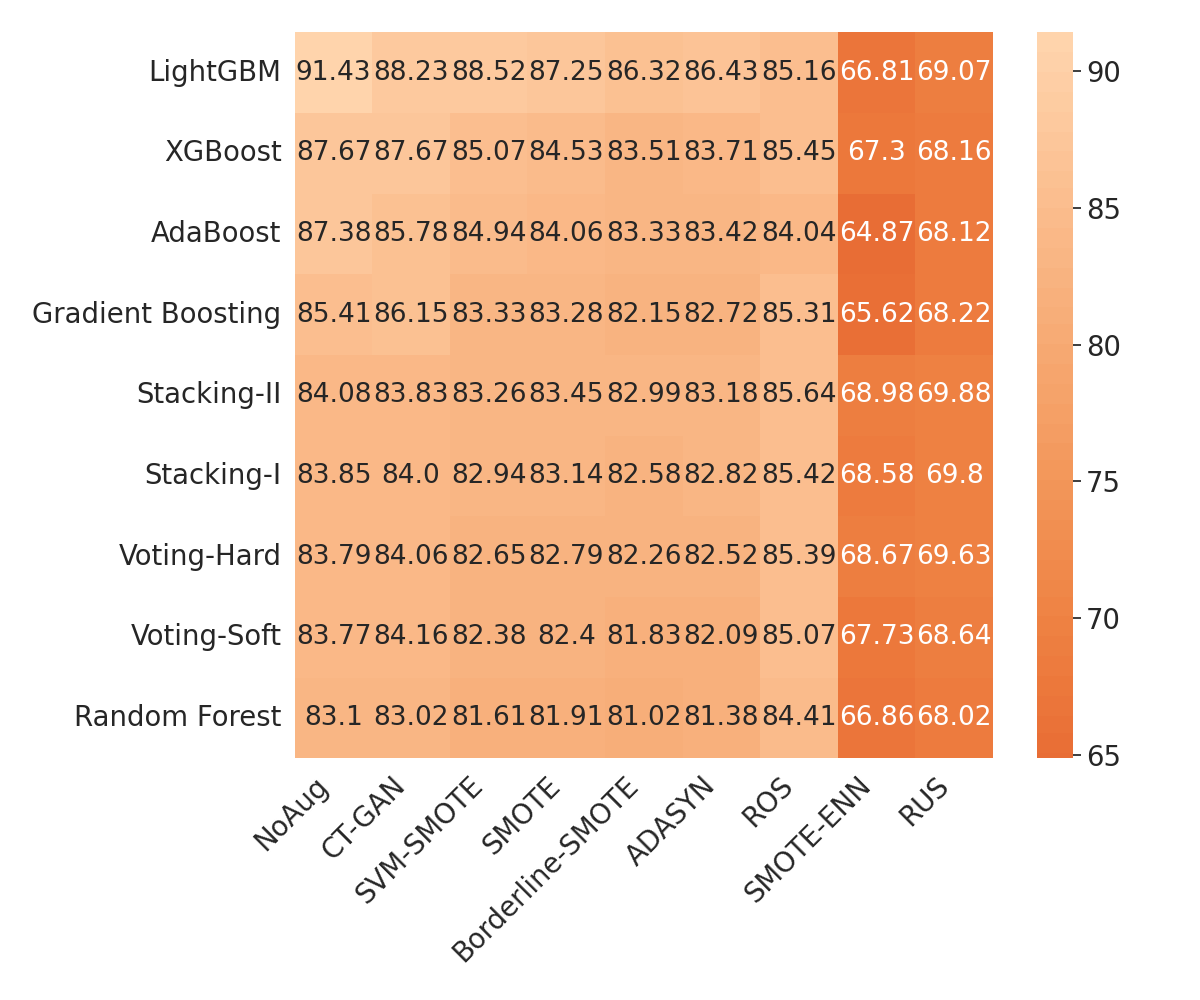}%
    \label{e2}
}

 \caption{Heatmap of F1 score of the Ecoli-derived datasets}
\label{EcoliHeat1}
\end{figure}

 In most datasets (Table \ref{tab:summary}), we find that combinations of LightGBM  with different data augmentation methods demonstrated to be superior to other combinations. We can observe results in Figures \ref{e1}, \ref{g1}, \ref{y1}, and \ref{y4} for LightGBM   showing  better performance than other classifiers. SMOTE-ENN and RUS performed worse than others  in most datasets.

\subsection{Results for multi-class classification}

\textcolor{black}{In order to demonstrate our framework for multi-class datasets, we took the Abalone dataset \cite{abalone} and created a multi-class problem.   The exact age of Abalone is determined by cutting the shell and counting the number of rings through a microscope. The age of Abalone can be determined using physical measurements and the dataset has 4177 instances and 8 features such as sex, length, width, diameter, shell weight, whole weight, etc. The Abalone dataset is a regression problem where the response variable is the shell ring age. We transformed the dataset into a multi-class problem of 5 major classes that are based on ring age as follows:}

\begin{enumerate}

  \item  

\textcolor{black}{Class 1: 0 - 5 years}

  \item  

\textcolor{black}{Class 2: 6- 10 years}

  \item  

\textcolor{black}{Class 3: 11 - 15 years}

  \item 

\textcolor{black}{Class 4: 16 - 20 years}


  
\end{enumerate}

 Furthermore, we created other versions of the Abalone dataset to have both 4 class and  3 class problems as shown in Table \ref{multiDatasets}, which highlights the number of instances for each class. Abalone-v1 has four classes while other datasets used have three classes in multiclass classification scenario. We follow the previous nomenclature (Table 2) for the dataset described as Abalone-12vs3vs4 (Abalone-v1) in Table \ref{multiDatasets}, which implies that the dataset was transformed into a three-class problem  by merging  label of class 2 with class 1, while classes 3 and 4 remained unchanged. The rest of the cases (Abalone-v2, Abalone-v3, and Abalone-v4), follow the same structure.

\textcolor{black}{The results of multiclass classification are shown in  Tables \ref{tab:F1macro}, \ref{tab:F1weighted}, \ref{tab:AUCmacroovo}, \ref{tab:AUCmacroovr}, \ref{tab:AUCweightedovo}, and \ref{tab:AUCweightedovr}.  
The results reveal several key insights into the performance of various classifier models and augmentation techniques across multiple metrics. In terms of F1-macro scores shown in Table \ref{tab:F1macro}, it is evident that both SMOTE-Stacking-I  and SMOTE-LightGBM, outperform other methods.  
The worst-performing model is SMOTE-ENN-AdaBoost. Moving to F1-weighted scores shown in Table \ref{tab:F1weighted}, we find that standalone models with no data augmentation emerge as the top performers, with Gradient Boosting leading and Random Forests showing the weakest performance.}

\textcolor{black}{Our results are consistent with the observations for binary classification problems in terms of the AUC score. Data augmentation has a greater impact than ensemble learning techniques on all four scores when using both the One-vs-One (OvO) and One-vs-Rest (OvR) approaches, with an average calculated using both the macro and weighted strategies.}

\textcolor{black}{We also report the AUC scores calculated using   OvO as shown in Table \ref{tab:AUCmacroovo} macro strategy, where SMOTE-Gradient Boosting emerges as the top performer. Meanwhile,  Random Forests consistently exhibit poor performance in the worst combinations reported. In AUC scores with the OvR macro strategy shown in Table \ref{tab:AUCmacroovr}, Stacking-II without any data augmentation gives the best performance in the best combinations while SMOTE-AdaBoost gives the worst performance, respectively.  
Finally, we find that the AUC-weighted scores shown in Tables \ref{tab:AUCweightedovo} and \ref{tab:AUCweightedovr},  show SMOTE-Gradient Boosting and Stacking-II as the top performers in the best combinations. Random Forests and AdaBoost without any data augmentation give the worst performances in the worst combinations, respectively.  In general, we find that combinations with  SMOTE-based data augmentation with ensemble learning perform better than others.}
 
 \begin{table*}[htbp]
\footnotesize
\centering
\caption{Best and worst performing combinations for all datasets.}
\label{tab:summary}
\begin{tabular}{lll}
\hline
& Best Combination & Worst Combination \\
\hline
Ecoli-0vs1 & Stacking 1 + CTGAN & LightGBM  + SMOTE-ENN\\
Ecoli-1 & LightGBM  + CTGAN & AdaBoost   + RUS\\
Ecoli-2 & LightGBM + None & AdaBoost  + SMOTE-ENN\\
Ecoli-3 & Stacking-1 + CTGAN & LightGBM + RUS\\ 
Glass-0 & Random Forests  + SMOTE-ENN & AdaBoost   + CT-GAN\\
Glass-1 &LightGBM + Borderline-SMOTE & Voting-Hard + None\\
Glass-2 & LightGBM + SMOTE & LightGBM + RUS \\
Glass-5vs12 & Stacking-1 + SVM-SMOTE & LightGBM + CT-GAN\\
Glass-6 & AdaBoost Classifier + ROS & LightGBM + RUS\\
Glass-0123vs567 & LightGBM +   No Augmentation & Stacking-2 + SMOTE-ENN\\ 
Yeast-1vs7 & LightGBM + No Augmentation & LightGBM + RUS\\
Yeast-2vs8 & Stacking-1 + CT-GAN & LightGBM + RUS\\
Yeast-3 & LightGBM + ROS & AdaBoost  + RUS\\ 
Yeast-5 & LightGBM  + ROS & AdaBoost   + RUS\\ 
Yeast-1289vs7 & Gradient Boost   + CT-GAN & Stacking-2 + RUS\\
Yeast-1458vs7 & LightGBM + ROS & Stacking-2 + RUS \\

\hline
\end{tabular}
\end{table*}

\begin{table*}[htbp]
\footnotesize
\centering
\caption{Best and worst performing combinations of the Abalone datasets according to F1 macro-average score}
\label{tab:F1macro}
\begin{tabular}{lllll}
\hline
& Best Combination & F1-macro & Worst Combination & F1-macro\\
\hline
Abalone-v1 & SMOTE + Stacking-1 & 56.559(57.507, 0.385) & SMOTE-ENN + AdaBoost & 16.463(16.595, 0.412)\\
Abalone-v2 & SMOTE + LightGBM& 55.870(56.894, 0.505) & SMOTE + AdaBoost & 43.336(43.336, 0.000)\\
Abalone-v3 & SMOTE + LightGBM& 51.755(52.597, 0.325) & SMOTE + Random Forest & 44.909(45.939, 0.759) \\
Abalone-v4 &  AdaBoost & 51.718(51.718, 0.000) & ROS + AdaBoost & 34.430(34.430, 0.000)\\
\hline
\end{tabular}
\end{table*}

\begin{table*}[htbp]
\footnotesize
\centering
\caption{\textcolor{black}{Best and worst performing combinations of the  Abalone dataset according to F1 weighted-average score.}}
\label{tab:F1weighted}
\begin{tabular}{lllll}
\hline
& Best Combination & F1-weighted & Worst Combination & F1-weighted\\
\hline

Abalone-v1 &   Voting-1 & 68.442(68.924, 0.334) & SMOTE + AdaBoost & 29.637(29.637, 0.000) \\
Abalone-v2 &   Stacking-2 & 72.080(72.818, 0.358) & SMOTE + AdaBoost & 51.415(51.415, 0.000)\\
Abalone-v3 &   Gradient Boost & 76.824(77.116, 0.144) & SMOTE-ENN + AdaBoost & 65.937(69.852, 5.378) \\
Abalone-v4 &   Voting-1 & 73.188(74.216, 0.516) & ROS + AdaBoost & 47.661(47.661, 0.000)\\

\hline
\end{tabular}
\end{table*}

\begin{table*}[htbp]
\footnotesize
\centering
\caption{\textcolor{black}{Best and worst performing combinations of the Abalone dataset according to AUC macro-average score (One vs. One approach).}}
\label{tab:AUCmacroovo}
\begin{tabular}{lllll}
\hline
& Best Combination & AUC-macro & Worst Combination & AUC-macro\\
\hline

Abalone-v1 & SMOTE + Gradient Boost & 86.627(86.728, 0.088) & SMOTE + AdaBoost & 64.763(64.763, 0.000) \\
Abalone-v2 & SMOTE + Gradient Boost & 72.08079.827(79.863, 0.017) &   Random Forests & 63.419(64.714, 0.662)\\
Abalone-v3 & ROS + Voting-II & 81.977(82.269, 0.169) &  Random Forests & 59.168(59.659, 0.238) \\
Abalone-v4 & SMOTE-ENN + Voting-II & 80.515(80.784, 0.152) &  Random Forests & 57.381(57.780, 0.188)\\

\hline
\end{tabular}
\end{table*}

\begin{table*}[htbp]
\footnotesize
\centering
\caption{\textcolor{black}{Best and worst performing combinations of the Abalone datasets according to AUC macro-average score (One vs. Rest approach).}}
\label{tab:AUCmacroovr}
\begin{tabular}{lllll}
\hline
& Best Combination & AUC-macro & Worst Combination & AUC-macro\\
\hline

Abalone-v1 &  LightGBM & 80.660(81.009, 0.151) & SMOTE + AdaBoost & 55.466(55.466, 0.000) \\
Abalone-v2 &   Stacking-II & 83.745(84.086, 0.144) &  Random Forests & 66.681(67.386, 0.491)\\
Abalone-v3 &   Stacking-II & 84.826(85.275, 0.166) & SMOTE + AdaBoost & 63.211(63.211, 0.000) \\
Abalone-v4 & Stacking-II & 80.654(80.900, 0.162) & SMOTE + AdaBoost & 56.825(56.825, 0.000)\\

\hline
\end{tabular}
\end{table*}

\begin{table*}[htbp]
\footnotesize
\centering
\caption{\textcolor{black}{Best and worst performing combinations of the Abalone dataset according to AUC weighted-average score (One vs. One approach).}}
\label{tab:AUCweightedovo}
\begin{tabular}{lllll}
\hline
& Best Combination & AUC-weighted & Worst Combination & AUC-weighted\\
\hline

Abalone-v1 & SMOTE + Gradient Boost & 86.627(86.728, 0.088) & SMOTE + AdaBoost & 64.763(64.763, 0.000) \\
Abalone-v2 & SMOTE + Gradient Boost & 79.827(79.863, 0.017) &  Random Forests & 63.419(64.714, 0.662)\\
Abalone-v3 & ROS + Voting-II & 81.977(82.269, 0.169) &   Random Forests & 59.168(59.659, 0.238) \\
Abalone-v4 & SMOTE-ENN + Voting-II & 80.515(80.784, 0.152) &  Random Forests & 57.381(57.780, 0.188)\\

\hline
\end{tabular}
\end{table*}

\begin{table*}[htbp]
\footnotesize
\centering
\caption{\textcolor{black}{Best and worst performing combinations of the Abalone datasets according to AUC weighted-average (One vs. Rest approach).}}
\label{tab:AUCweightedovr}
\begin{tabular}{lllll}
\hline
& Best Combination & AUC-weighted & Worst Combination & AUC-weighted\\
\hline

Abalone-v1 &   LightGBM & 80.660(81.009, 0.151) & SMOTE + AdaBoost & 55.466(55.466, 0.000) \\

Abalone-v2 &  Stacking-II & 83.745(84.086, 0.144) &  Random Forests & 66.681(67.386, 0.491)\\

Abalone-v3 &  Stacking-II & 84.826(85.275, 0.166) & SMOTE + AdaBoost & 63.211(63.211, 0.000) \\
Abalone-v4 &   Stacking-II & 80.654(80.900, 0.162) & SMOTE + AdaBoost & 56.825(56.825, 0.000)\\

\hline
\end{tabular}
\end{table*}

\begin{figure}[h]
\centering
\includegraphics[height=5cm]{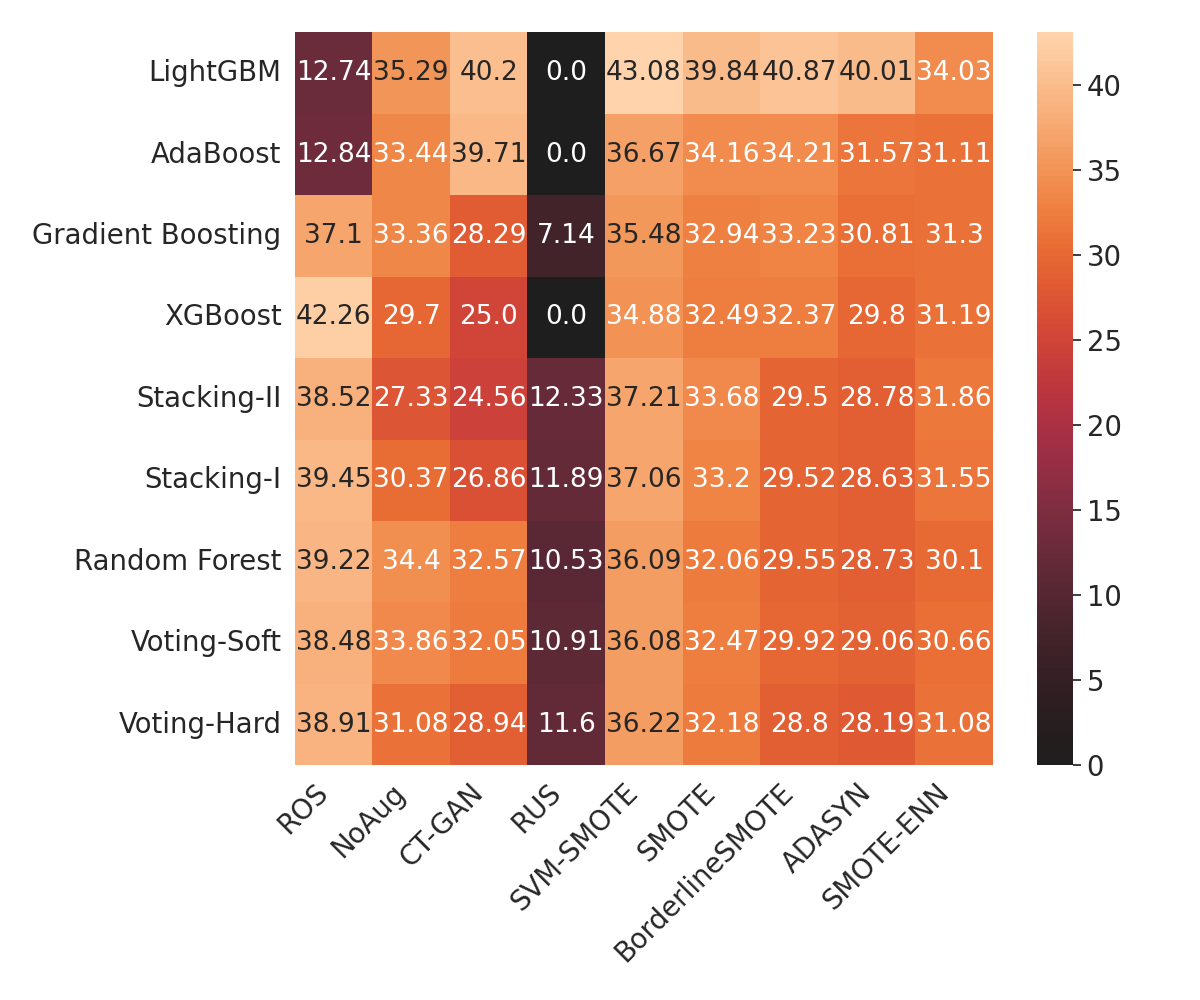}
\caption{Heatmap of F1 score of the Yeast-1458vs7 dataset} 
\label{y1458vs7}
\end{figure}

\section{Discussion}
 
 The problem of class imbalance is a common issue in many applications, such as medical imaging, fraud detection, and natural language processing (NLP). As more and more data is collected, the challenge of emerging machine learning models is to handle class imbalance. In this study, we reviewed selected combinations of prominent established and novel ensemble learning and data augmentation methods. We computationally evaluated the performance of multiple combinations on known datasets, since a similar evaluation was done more than a decade ago \cite{galar2011review}.  We next highlight issues that can enable better progress of the methodologies and problem domains that involve class imbalance problems.

\subsection{Benchmark datasets }

 We used a series of benchmark datasets, which mostly catered to binary class imbalance problems. We modified selected datasets to handle multi-class class imbalance scenarios. We note that the datasets utilised are relatively small tabular datasets with a few dozen features at most, with less than a few thousand instances. In terms of existing benchmark datasets, there exist major limitations in developing robust machine learning models to deal with class imbalance. This is because the benchmark datasets need to take into consideration multiple applications such as image and vision-related problems, natural language processing, and signal processing.  We need to develop better benchmark datasets to evaluate class imbalance problems to suit emerging models and big data problems \cite{leevy2018survey}.

 Big data is still a challenging problem, and one can explore the benefits of combining deep ensemble models with data augmentation \cite{zhou2014big}. Deep learning models have incorporated various architectural properties for automated feature extraction and modelling temporal data, such as CNNs and RNNs.  In future work, we can investigate the combination of deep learned-based ensemble models \cite{ganaie2022ensemble} with data augmentation for spatial and temporal data for small-scale and big data related problems.


\subsection{Natural language processing}

 In  NLP \cite{cambria2014jumping}, class imbalance problems are apparent in classification tasks such as sentiment classification, spam detection, detection of fake news, bullying and abuse on social media, and digital marketing. Some effort has been made by data augmentation methods (SMOTE and GAN) for class imbalance in text classification \cite{padurariu2019dealing}; however, data argumentation methods alone cannot address the issue for more complex NLP problems. Pre-trained models, such as BERT \cite{devlin2018bert}   have the capability to provide robust word embedding, and hence be used for NLP tasks such as text completion  \cite{zhou2020pre}.  Recently, there has been much attention given to dialogue and chat models in NLP powered by large language models such as \textit{generalised pre-trained  Transformers} (Chat-GPT) \cite{floridi2020gpt}. GPT models are trained on large text corpus of data (such as Wikipedia) and have been effective in generative summaries of texts, answering questions, and even undertaking exams \cite{floridi2020gpt}. They have immense potential to address class imbalance in NLP problems. Some work has been used to incorporate BERT and GPT models for data augmentation \cite{kumar2020data}; they can provide an effective way to
augment training data which can be used for low resource language setting; i.e. languages native to a significant population of speakers but do not have substantial amounts of data for training a language model  \cite{ragni2014data}. Furthermore,   transfer learning methods have a huge potential for NLP tasks such as text summarizations and completion \cite{raffel2020exploring};  these can add further resources for data augmentation for addressing class imbalance problems.

\subsection{Computer vision}

Data augmentation methods have been prominently used for image-based data and deep learning models \cite{shorten2019survey}. Data augmentation techniques have been widely used in computer vision and have proven effective in improving model performance and reducing overfitting.  In the future, we expect to see more advanced and sophisticated data augmentation techniques that can better simulate real-world variations in the data. One area that could see significant progress is using generative models for data augmentation. These models can be used to synthesise new, realistic images that can be used to expand the dataset. This can be particularly useful in cases where obtaining additional real-world data is difficult or expensive. Another area that could see progress is using domain adaptation methods for data augmentation. This will allow the model to adapt to different domains and generalise to new unseen data. The scope for the Transformer model \cite{parmar2018image} has not been limited to NLP, pre-trained Transformer models have been also used for image-based problems \cite{chen2021pre}. Hence, data augmentation combined with pre-trained Transformer models can enable the generation of synthetic image and video data that can address class imbalance problems. 

\subsection{Explainable artificial intelligence}

Ensemble learning, particularly tree-based models such as Random Forests and XGBoost are considered white-box models \cite{loyola2019black} since we can extract information about the decision making process in the form of \textit{if-then} rules. They also give information about input feature contributions to the decision-making process. A number of applications, particularly in the medical diagnosis domain, require \textit{explainable artificial intelligence} (XAI) \cite{saeed2023explainable}, where information about how the decision has been made is important.   Hence, it is important to use tree-based ensemble models (XGBoost and Random Forests) with data augmentation methods in such problem domains. However, medical practitioners may question the validity of synthetic data for class imbalance problems. Moreover, GANs \cite{zhang2019detecting,wang2021generative} and related data generation methods such as \textit{deepfake}  \cite{yu2021survey,westerlund2019emergence,li2020celeb} have negative connotations since they can be used to impersonate individuals, by generating images, speech and related biometrics \cite{agarwal2019protecting}. Therefore, XAI needs to take into account class imbalance problems and related problems emerging from limited data, irregular features in data, missing data,  and noisy data. So-called, 'grey box' models can also help in addressing the problem where some information is available \cite{pintelas2020grey}. It would be difficult to convince medical practitioners that a medical diagnosis system uses data generated by deepfake technology to address class imbalance problems.

 \subsection{Extreme forecasting}

 In climate science, climate extremes such as storms, droughts, and floods pose a major threat to our civilisation and habitats \cite{rolnick2022tackling}. Ensemble learning models have been prominent in climate extreme problems such as flood forecasting \cite{cloke2009ensemble,wu2020ensemble}. It is common to find climate extreme problems with class imbalance, which may not be only in classification but also in regression (prediction) problems. The issue is known as extreme forecasting, and this is similar to class imbalance and the same methods can be used to address them. The major challenge in class imbalance is the lack of data instances for a particular class while in extreme forecasting, the challenge is the lack of occurrence of extreme events; hence they both face a lack of data. The combination of data augmentation and ensemble learning can address some of the challenges in climate sciences that deal with limited, sparse, and noisy data for extreme events.

 Such problems not only occur in the area of climate sciences, but they also occur in other fields that have complex relationships between various interests, such as econometrics and financial forecasting. In finance, the market faced drastic changes and hence forecasting volatility has been of interest \cite{poon2003forecasting,bee2018estimating}. Recently, with the COVID-19 pandemic, the financial markets faced extreme conditions \cite{chandra2021bayesian} which resulted in financial crisis in certain countries and the closure of many businesses. Hence, ensemble learning and data augmentation also have the potential to address issues in these areas. 

\textcolor{black}{\subsection{Pre-trained deep learning models  for CI problems}
The utilisation of pre-trained Transformer models has emerged as a powerful and effective strategy for emerging problems in machine learning \cite{demirkiran2022ensemble, arshed2023multi} and NLP \cite{agrawal2022lastresort, shao2021cpt}. Pre-trained models such as BERT and Chat-GPT   have demonstrated remarkable capabilities in understanding complex linguistic patterns and semantics \cite{han2021pre, wang2022pre}. In the case of CI  datasets, these models offer several advantages \cite{acheampong2021transformer} since they can extract nuanced features and contextual information from the data, enabling better discrimination between minority and majority classes. Fine-tuning pre-trained models on imbalanced datasets can lead to improved classification performance since these models can learn to assign more significance to the underrepresented classes \cite{anaby2020not}. Furthermore, techniques such as oversampling or re-weighting can be combined with pre-trained models to further enhance their effectiveness in addressing CI problems \cite{ullah2022explainable}.}

 \subsection{Uncertainty quantification}

The need for uncertainty quantification in model predictions becomes important when we are using synthetic data to improve predictions, such as in the case of class imbalance of forecasting extremes. Bayesian inference provides a rigorous approach for quantifying the uncertainty of model parameters that can be projected in model predictions. Bayesian methods such as variational inferences and MCMC (Markov Chain Monte Carlo)  have been prominent in statistical models and machine learning. In the last decade, there have been major advances with MCMC methods in  Bayesian deep learning \cite{wang2020survey,kendall2017uncertainties,chandra2022revisiting,chandra2021bayesian}. However, these methods have not been used much for handling class imbalance problems.  Uncertainty quantification using Bayesian inference can be used to predict the uncertainty associated with model parameters and data. MCMC methods can also be used for data augmentation, where data can be generated from models or distributions from the data. Furthermore, Bayesian optimisation methods \cite{snoek2012practical,chandra2022distributed}  can be used for hyperparameter tuning in models that address the class imbalance. Bayesian inference with MCMC and variational inference \cite{chandra2021bayesian,kapoor2023cyclone} can also be used for ensemble learning as demonstrated in a Bayesian gradient boosting framework \cite{Bai2023}. Apart from these, other novel ensemble learning approaches such as evolutionary bagging~\cite{ngo2022evolutionary} can be combined with data augmentation methods to address CI problems. There is great potential for Bayesian deep learning models to address class imbalance and extreme forecasting where uncertainty quantification in model predictions is greatly needed, such as in climate sciences and medical diagnosis.

 \subsection{Outlook}

 We note that our results show that CT-GAN combination with ensemble learning methods does not outperform most of the established methods (SMOTE, SMOTE-ENN, RUS, and ADASYN etc.) which are simpler and computationally less expensive to implement.  The data augmentation methods such as CT-GANs are relatively new and hence not included for evaluation in the review paper that came out more than a decade earlier (2011) \cite{galar2011review}. We note that one of the reasons that CT-GANs have not outperformed the established methods in the literature could be because  CT-GANs \cite{xu2019modeling} have been designed for tabular data that feature both discrete and continuous values.  It depends on what level of discrete and continuous value columns are present in the data for them to outperform other methods. Moreover, they also require extensive hyperparameter tuning, which can be done in future work to improve the results further. We recognise the importance of exploring newer techniques but also note that the past methods need to be evaluated for multi-class datasets using a wide range of metrics, which we have implemented in this study. We also release open-source code with the framework so that further extensions can be made. 

\section{Conclusion}

 In this study, we implemented a computational review of combinations of existing and novel data augmentation and ensemble learning models for selected CI problems. Our review shows that the research community has immensely used these combinations to counter the CI problem in the past decade. We implemented combinations of ensemble learning and data augmentation models on selected binary and multiclass imbalanced datasets. We also reported the individual performance ensemble learning models without any data augmentation.   We conclude that certain combinations presented in this study are more effective than others in CI problems; our study found that the best combinations are SMOTE-LightGBM and ROS-LightGBM. We also found that CT-GANs, which are newer additions to the family of data augmentation methods for CI, have not been the top performers in our computational study. We note that it is challenging to design GANs for CI problems, due  to hyperparameter tuning and computational complexity. We found that traditional methods such as SMOTE and ROS are not only better in performance, but also computationally less expensive than GANs.

Ensemble learning provides a powerful approach for improving the performance of conventional machine learning models that face challenges with CI problems. Ensemble learning can effectively leverage the strengths of different machine learning models, training algorithms, and loss functions.  Hence, due to flexibility in designing novel architectures, there is potential for ensemble learning to handle imbalanced datasets. In future work, there is potential for the design of novel ensemble learning methods that naturally combine data augmentation methods for addressing the class imbalance.  
 
\section*{Code and Data }

Github repository for code and data \footnote{\url{https://github.com/sydney-machine-learning/review-classimbalanced-ensemblelearning}}

  \bibliographystyle{elsarticle-num} 
  \bibliography{bibliography.bib}

\end{document}